\documentclass{article}
\usepackage{arxiv}

\usepackage[numbers, sort]{natbib}
\usepackage[utf8]{inputenc}
\usepackage[T1]{fontenc}  
\usepackage{hyperref} 
\usepackage{url} 
\usepackage{amsfonts}   
\usepackage{nicefrac} 
\usepackage{microtype}  
\usepackage{lipsum}
\usepackage{graphicx}
\usepackage{multirow} 
\usepackage{array} 
\usepackage{caption}
\usepackage{tabularx}
\usepackage{booktabs}
\usepackage{footnote}
\usepackage{makecell}
\makesavenoteenv{tabular}
\makesavenoteenv{table}
\usepackage[inline]{enumitem}
\usepackage{multicol}
\usepackage{array}
\usepackage{siunitx}
\usepackage{float}
\usepackage{pdflscape}
\usepackage{enumitem}
\usepackage{pdfpages}
\usepackage{rotating}
\usepackage{soul,color}
\definecolor{aliceblue}{HTML}{4682B4}
\definecolor{greenish}{HTML}{2a9d8f}
\hypersetup{
  colorlinks   = true, 
  urlcolor     = aliceblue, 
  linkcolor    = aliceblue, 
  citecolor   = aliceblue 
}
\usepackage{amsmath}
\definecolor{lstback}{rgb}{0.95,0.95,0.92}
\definecolor{lststring}{RGB}{0,0,0}
\definecolor{lstkeyw}{RGB}{106,90,205}
\usepackage{listings}
\lstdefinelanguage{searchstrategy}%
{
  morekeywords={AND,OR,TITLE,ABS,KEY},
   sensitive=true,
   alsoother={\$,\,}, 
   morecomment=[l]\#,
   morecomment=[n]{\#=}{=\#},
   morestring=[s]{"}{"},
   morestring=[m]{'}{'},
}[keywords,comments,strings]
\lstdefinestyle{mystyle}{
    backgroundcolor=\color{lstback}, 
    stringstyle=\color{lststring},
    keywordstyle=\color{lstkeyw},
    basicstyle=\ttfamily\footnotesize,
    breakatwhitespace=false,         
    breaklines=true,                 
    captionpos=t,                    
    keepspaces=true,                                  
    showspaces=false,                
    showstringspaces=false,
    showtabs=false,                  
    tabsize=2,
    literate={,}{{\textcolor{lstkeyw}{,}}}1, 
}
\usepackage{cleveref}
\usepackage{xspace}

\soulregister\cite7
\soulregister\ref7
\soulregister\pageref7
\setstcolor{red}
\setul{}{.2ex}

\newcolumntype{C}[1]{>{\centering\let\newline\\\arraybackslash\hspace{0pt}}m{#1}}

\definecolor{olivegreen}{RGB}{85, 107, 47}
\definecolor{cerisepink}{rgb}{0.8706, 0.1922, 0.3882}

\newcommand{\new}[1]{{\leavevmode \color{black}{#1}}}

\usepackage{glossaries} 
\glsdisablehyper 
\newacronym{ai}{AI}{Artificial Intelligence}
\newacronym{nn}{NN}{Neural Network}
\newacronym{rl}{RL}{Reinforcement Learning}
\newacronym{pbo}{PbO}{Programming by Optimization}
\newacronym{stn}{STN}{Search Trajectory Network}
\newacronym{isa}{ISA}{Instance Space Analysis}
\newacronym{gp}{GP}{Genetic Programming}
\newacronym{cp}{CP}{Constraint Programming}
\newacronym{lp}{LP}{Linear Programming}
\newacronym{ilp}{ILP}{Integer Linear Programming}
\newacronym{mip}{MIP}{Mixed Integer Programming}
\newacronym{mipp}{MIPP}{Mixed Integer Programming Problem}
\newacronym{milp}{MILP}{Mixed Integer Linear Programming}
\newacronym{mh}{MH}{Metaheuristic}
\newacronym{ts}{TS}{Tabu Search}
\newacronym{hc}{HC}{Hill Climbing}
\newacronym{sa}{SA}{Simulated Annealing}
\newacronym{ga}{GA}{Genetic Algorithm}
\newacronym{brkga}{BRKGA}{Biased Random-key Genetic Algorithm}
\newacronym{grasp}{GRASP}{Greedy Randomized Adaptive Search Procedure}
\newacronym{tsp}{TSP}{Traveling Salesperson Problem}
\newacronym{aco}{ACO}{Ant Colony Optimization}
\newacronym{lns}{LNS}{Large Neighborhood Search}
\newacronym{ec}{EC}{Evolutionary Computation}
\newacronym{ea}{EA}{Evolutionary Algorithm}
\newacronym{co}{CO}{Combinatorial Optimization}
\newacronym{col}{CO}{Continuous Optimization}
\newacronym{ccop}{CCOP}{Cardinality-constrained Optimization Problem}
\newacronym{or}{OR}{Operations Research}
\newacronym{cop}{COP}{Combinatorial Optimization Problem}
\newacronym{colp}{COLP}{Continuous Optimization Problem}
\newacronym{csp}{CSP}{Constraint Satisfaction Problem}
\newacronym{cvpr}{CVRP}{Capacited Vehicle Routing Problem}
\newacronym{vrp}{VRP}{Vehicle Routing Problem}
\newacronym{sat}{SAT}{Boolean Satisfiability Problem}
\newacronym{llm}{LLM}{Large Language Model}
\newacronym{gpt}{GPT}{Generative Pre-trained Transformer}
\newacronym{dgm}{DGM}{Deep Generative Model}
\newacronym{prisma}{PRISMA}{Preferred Reporting Items for Systematic Reviews and Meta-Analyses}
\newacronym{bpmn}{BPMN}{Business Process Model and Notation}
\newacronym{nlp}{NLP}{Natural Language Processing}
\newacronym{nl}{NL}{Natural Language}
\newacronym{quorom}{QUOROM}{Quality of Reporting of Meta-analyses}
\newacronym{aec}{AEC}{Architecture, Engineering, and Construction Industry}
\newacronym{fmcvrp}{FM-MCVRP}{Montreal Capacitated Vehicle Routing Problem}
\newacronym{nl4opt}{NL4Opt}{Natural Language for Optimization Modeling}
\newacronym{pfsp}{PFSP}{Permutation Flowshop Scheduling Problem}
\newacronym{bpp}{BPP}{Bin Packing Problem}
\newacronym{op}{OP}{Orienteering Problem}
\newacronym{dpp}{DPP}{Decap Placement Problem}
\newacronym{mkp}{MKP}{Multiple Knapsack Problem}
\newacronym{kp}{KP}{Knapsack Problem}
\newacronym{gcp}{GCP}{Graph Coloring Problem}
\newacronym{rag}{RAG}{Retrieval-Augmented Generation}
\newacronym{ml}{ML}{Machine Learning}
\newacronym{qp}{QP}{Quadratic Programming}

\newcommand{\arxiv}{arXiv\xspace}

\newcommand{\numqueries}{14\xspace}

\newcommand{\numrecordsretrievedscholar}{\new{648}\xspace}
\newcommand{\numrecordsretrievedscopus}{\new{1,396}\xspace}
\newcommand{\numrecordsretrieved}{\new{2,044}\xspace}

\newcommand{\numrecordsnoauthorscholar}{\new{4}\xspace}
\newcommand{\numrecordsnoauthorscopus}{\new{84}\xspace}
\newcommand{\numrecordsnoauthordistinct}{\new{84}\xspace}
\newcommand{\numrecordsduplicates}{\new{488}\xspace}
\newcommand{\numrecordsduplicatescholar}{\new{269}\xspace}
\newcommand{\numrecordsduplicatescopus}{\new{196}\xspace}
\newcommand{\numrecordsduplicatesshared}{\new{23}\xspace}

\newcommand{\numrecordsretrievednodupsnoauthor}{\new{1,472}\xspace}
\newcommand{\numrecordsremovedautomatically}{68\xspace}

\newcommand{\numrecordsretrievedwrongpapertype}{\new{7}\xspace}
\newcommand{\numrecordsretrievedoptforllm}{\new{47}\xspace}
\newcommand{\numrecordsretrievednollmornoco}{\new{1227}\xspace}

\newcommand{\numrecordsfinalbeforecitationtracking}{\new{123}\xspace}
\newcommand{\numrecordsfoundthroughcitationtracking}{\new{420}\xspace}
\newcommand{\numrecordsduplicatescitationtracking}{\new{101}\xspace}
\newcommand{\numrecordsduplicatescitationtrackingdistinct}{\new{319}\xspace}
\newcommand{\numrecordscitationtrackingremovedmanually}{\new{268}\xspace}
\newcommand{\numrecordscitationtrackingfinal}{\new{51}\xspace}

\newcommand{\numrecordsfoundthroughcitationtrackingwrongpapertype}{\new{5}\xspace}
\newcommand{\numrecordsfoundthroughcitationtrackingoptforllm}{\new{11}\xspace}
\newcommand{\numrecordsfoundthroughcitationtrackingnollmornoco}{\new{251}\xspace}
\newcommand{\numrecordsfoundthroughcitationtrackingwronglanguage}{\new{1}\xspace}

\newcommand{\numstudies}{\new{174}\xspace}

\newcommand{\numstudiesincluded}{\new{103}\xspace}

\newcommand{\numcriteria}{8\xspace}
\newcommand{\numinccriteria}{4\xspace}
\newcommand{\numexccriteria}{4\xspace}

\newcommand{\moreTasks}{\new{24}\xspace}

\newcommand{\problemModeling}{\new{38}\xspace}
\newcommand{\domainKnowledge}{\new{11}\xspace}
\newcommand{\entityRecognition}{\new{25}\xspace}
\newcommand{\modelCreation}{\new{25}\xspace}

\newcommand{\solutionMethods}{\new{64}\xspace}
\newcommand{\codeGeneration}{\new{36}\xspace}
\newcommand{\solutionGeneration}{\new{28}\xspace}
\newcommand{\parameterTuning}{\new{4}\xspace}
\newcommand{\algorithmSelection}{\new{1}\xspace}

\newcommand{\benchmarking}{\new{7}\xspace}

\newcommand{\validation}{\new{9}\xspace}
\newcommand{\solutionValidation}{\new{5}\xspace}
\newcommand{\modelValidation}{\new{6}\xspace}

\newcommand{\counterCP}{\new{7}\xspace}
\newcommand{\counterLP}{\new{19}\xspace}
\newcommand{\counterILP}{\new{2}\xspace}
\newcommand{\counterMILP}{\new{18}\xspace}
\newcommand{\counterHeuristic}{\new{9}\xspace}
\newcommand{\counterHyperHeuristic}{\new{3}\xspace}
\newcommand{\counterEvolutionaryAlgorithm}{\new{3}\xspace}
\newcommand{\counterGeneticAlgorithm}{\new{3}\xspace}
\newcommand{\counterMultiObjective}{\new{3}\xspace}

\newcommand{\counterPythonTotal}{\new{35}\xspace}
\newcommand{\counterPythonCodeGeneration}{\new{30}\xspace}

\newcommand{\llm}[1]{\footnotesize{\texttt{\hyphenchar\font=`\-\relax #1}\xspace}}

\newcommand{\numarchitectures}{\new{22}\xspace}
\newcommand{\numarchgenerative}{\new{13}\xspace}
\newcommand{\numarchtextual}{\new{5}\xspace}
\newcommand{\numarchmultimodal}{\new{4}\xspace}
\newcommand{\numllm}{\new{70}\xspace}

\newcommand{\numllmproblemmodeling}{\new{40}\xspace}
\newcommand{\numllmproblemmodelingmodelcreation}{\new{19}\xspace}
\newcommand{\numllmproblemmodelingentityrecognition}{\new{22}\xspace}
\newcommand{\numllmproblemmodelingdomainknowledge}{\new{12}\xspace}

\newcommand{\numllmsolutionmethod}{\new{60}\xspace}
\newcommand{\numllmsolutionmethodsolutiongeneration}{\new{23}\xspace}
\newcommand{\numllmsolutionmethodcodegeneration}{\new{35}\xspace}
\newcommand{\numllmsolutionmethodparametertuning}{\new{5}\xspace}

\newcommand{\numllmtbenchmarking}{\new{9}\xspace}
\newcommand{\numllmtbenchmarkingvisualanalysis}{\new{8}\xspace}
\newcommand{\numllmtbenchmarkingexplainability}{\new{5}\xspace}

\newcommand{\numllmtvalidation}{\new{10}\xspace}
\newcommand{\numllmtvalidationsolutionvalidation}{\new{7}\xspace}
\newcommand{\numllmtvalidationmodelvalidation}{\new{9}\xspace}

\title{Large Language Models for Combinatorial Optimization: A Systematic Review}

\author{
  \begin{tabular}{@{}l@{\hskip 2cm}l@{}}
    \begin{tabular}[t]{@{}l@{}}
      Francesca Da Ros \\
      University of Udine, Italy \\
      \texttt{francesca.daros@uniud.it}
    \end{tabular}
    &
    \begin{tabular}[t]{@{}l@{}}
      Michael Soprano \\
      University of Udine, Italy \\
      \texttt{michael.soprano@uniud.it}
    \end{tabular}
    \\[4em] 
    \begin{tabular}[t]{@{}l@{}}
      Luca Di Gaspero \\
      University of Udine, Italy \\
      \texttt{luca.digaspero@uniud.it}
    \end{tabular}
    &
    \begin{tabular}[t]{@{}l@{}}
      Kevin Roitero \\
      University of Udine, Italy \\
      \texttt{kevin.roitero@uniud.it}
    \end{tabular}
  \end{tabular}
}

\begin{document}
\maketitle
\begin{abstract}
This systematic review explores the application of \glspl{llm} in \gls{co}. We report our findings using the \gls{prisma} guidelines. We conduct a literature search via Scopus and Google Scholar, examining over \new{2,000} publications. We assess publications against four inclusion and \new{four} exclusion criteria related to their language, research focus, publication year, and type. Eventually, we select \numstudiesincluded studies. 
We classify these studies into semantic categories and topics to provide a comprehensive overview of the field, including the tasks performed by \glspl{llm}, the architectures of \glspl{llm}, the existing datasets specifically designed for evaluating \glspl{llm} in \gls{co}, and the field of application. Finally, we identify future directions for leveraging \glspl{llm} in this field.
\glsresetall
\end{abstract}


\section{Introduction}
\label{sec:introduction}

\glspl{cop} are a class of optimization problems characterized by discrete variable domains and finite search space. \gls{co} plays a crucial role in identifying promising solutions in many areas requiring complex decision-making capabilities, such as industrial \cite{Vass2022} \new{and} employee scheduling \cite{KLETZANDER2024104172,CESCHIA2023100379}, facility location \cite{10.1007/978-3-031-62922-8_11,CESCHIA2024109858}, and timetabling \cite{Steiner2024,CALIK2024106438}. Traditionally, such problems are modeled with techniques like \gls{lp}, \gls{ilp}, \gls{milp}, and \gls{cp}, further solved through commercial solvers such as \new{IBM ILOG} CPLEX \cite{IBMsched2017} or Gurobi \cite{gurobi} and through heuristic and \gls{mh} algorithms \cite{Sörensen2018}.

While many successful \gls{co} applications have been developed, the design and engineering of optimization tasks remain primarily human-driven. Users must convert the problem into an optimization model by defining a set of variables, constraints, and one or more objective functions, then coding and running a software solver or algorithm to find solutions. These activities are not trivial and require a certain extent of expertise.

Inspired by the recent advancements in the usage of \glspl{llm} to perform a wide array of complex tasks, there is growing interest in integrating \glspl{llm} into \gls{co} to mitigate the human-intensive aspects of optimization \cite{meadows2024survey,fan2024artificial,huang2024large,wu2024evolutionary}.
The abilities of \glspl{llm} to process, interpret, and generate human language make them particularly suited for tackling activities within \gls{co}, including the translation of natural language descriptions to formalisms such as mathematical models \cite{jang2022tag,he2022linear} and code generation \cite{tsouros2023holy,lawless2024i}. 

The rapid advancement in \gls{ai} and in particular in \gls{nlp} has led to a rapid increase in the capabilities and applications of \glspl{llm}, resulting in a proliferation of scholarly works and models being developed. 
While highlighting the increasing activity in the field, this multitude of studies has created a complex body of knowledge that is challenging to navigate. Looking specifically at the application of \glspl{llm} to \gls{co}, the academic literature is limited and fragmented, with existing works characterized by diverse methodologies, areas of applications, and findings.

Therefore, this systematic review aims to consolidate the current state-of-the-art in \glspl{llm} applied to \gls{co}. 
We identify, screen, analyze, and systematically organize the literature to clarify the topic and identify strategic directions for ongoing and future research efforts. The process is reported following the \gls{prisma} guidelines.
Through this examination, we seek to understand the capabilities of \glspl{llm} in addressing complex optimization tasks and to explore the evolving trends and directions in this field.
By systematically synthesizing and analyzing existing research, this review aims to provide a structured understanding of how \glspl{llm} are employed in \gls{co} and offer insights that can inform future research in the field.

The remainder of this review is structured as follows. In \Cref{sec:aims}, we discuss the aims and motivations that drove our work. We then explore the relationships and differences with related work in \Cref{sec:related-work}. In \Cref{sec:background}, we present the background necessary to understand the interconnections between \glspl{llm} and \gls{co}. We provide a detailed account of the methodology we followed in \Cref{sec:methodology}. In \Cref{sec:analysis}, we classify and discuss the studies included in our review. Next, we outline future research directions in \Cref{sec:future-research-directions} and discuss the limitations of our approach in \Cref{sec:limitations}. Finally, we draw some conclusions and propose future work in \Cref{sec:conclusions}.

\section{Aims and Motivations} 
\label{sec:aims}

The main objective of this systematic review is to critically evaluate the current state of \glspl{llm} application in \gls{co}.
To this \new{end}, \new{we aim to answer the following questions}: 
\begin{enumerate}[label=\emph{(\roman*)}] 
    \item How are \glspl{llm} currently being applied to \glspl{cop}? 
    \item Which tasks of the optimization process are aided with \glspl{llm}? 
    \item Which \gls{llm} architectures and training paradigms are most effective for \gls{co}?
    \item Which application fields are employing \glspl{llm} within \gls{co}?
    \item What are the main trends in the application of \glspl{llm} within \gls{co}? 
    \item What are possible research directions in the application of \glspl{llm} to \gls{co}? 
\end{enumerate} 

Two primary motivations drive this systematic review.
Firstly, traditional optimization approaches primarily rely on human expertise (e.g., application domain knowledge, coding capabilities, etc.), thus limiting their applicability, scalability, and efficiency. \new{Leveraging \glspl{llm} offers a promising direction, as their success in tasks like entity recognition, domain knowledge, and coding suggests their potential for \gls{co}.} 
For instance, \glspl{llm} could be used for translating high-level descriptions of solution algorithms to executable code \cite{liu2024evolution}.

Secondly, the rapid evolution of \gls{nlp} \new{(in particular of \glspl{llm}) and their application in} optimization techniques \new{(specifically \gls{co})}, highlights the urgency \new{of} a comprehensive systematic review. 
Existing literature and reviews are scattered, lacking a cohesive framework that provides researchers and practitioners with clear guidance in the domain of \glspl{llm} applied to \gls{co}.

While exploring the reverse relationship between \glspl{llm} and \gls{co} (i.e., applying \gls{co} techniques to enhance \glspl{llm}) is also a valid research direction, we deliberately chose not to include it in this survey. This is primarily because such applications essentially use well-established \gls{co} methods in a new, albeit challenging, domain. Instead, we focus on the novel and emerging contributions of \glspl{llm} to the practice of \gls{co}, as this also aligns more closely with the primary research interests of some of the authors, particularly in exploring the potential for \glspl{llm} to enhance \gls{co} techniques.
For the same reason, our review emphasizes \gls{co} rather than other paradigms, such as continuous optimization. This decision is also motivated by the inherent differences between the two paradigms: continuous optimization typically relies on mathematical models or black-box ones, whereas \gls{co} encompasses a broader and more diverse set of problem domains (e.g., scheduling, assignment, routing, permutation-based problems) characterized by intricate substructures and complex constraints. This diversity makes the application of \glspl{llm} to \gls{co} potentially more impactful, given the richness and complexity of the \gls{co} landscape.

\section{Related Work} 
\label{sec:related-work}

A limited number of surveys have addressed the intersection of \gls{llm} and \gls{co} along with related topics.
It is important to remark that none of these works are systematic reviews.
In this section, we overview their main characteristics and highlight the differences with respect to our work.

\citet{fan2024artificial} \new{presented} a review of \gls{ai} applications in \gls{or}, with a focus on three specific aspects of the optimization process: conversion of data into modeling parameters, model formulations, and model optimization.
Although they extensively \new{covered} various \gls{ai} techniques, such as \gls{rl} and \glspl{nn}, the usage of \glspl{llm} \new{was} investigated only in the context of model formulation \cite[Section 4]{fan2024artificial}.
\citet{huang2024large} \new{explored} the topic by including not only \gls{co}, but also continuous optimization and the use of optimization for \glspl{llm}.
However, when discussing \glspl{llm} for optimization, they \new{examined} a set of 20 papers and \new{focused} on a narrow range of optimization aspects, specifically algorithm generation and the usage of \glspl{llm} as search operators.
\new{\citet{han-2024-syrvey} \new{reviewed} the literature on \glspl{llm} concerning their mathematical capabilities and briefly \new{discussed} their applications in \gls{co} along with math word problems, geometry problems, and theorem proving.}
\new{\citet{liu2024systematicsurveylargelanguage} \new{described} the application of \glspl{llm} to algorithm design in various fields, like optimization, \gls{ml}, mathematical reasoning, and scientific discovery.}
\citet{wu2024evolutionary} \new{provided} a review on the intersection of \gls{ec} and \glspl{llm}.
As \new{done} by \citet{huang2024large}, they also \new{covered} aspects like continuous optimization and the use of optimization for \glspl{llm} enhancement.
However, their focus on \gls{ec} \new{represented} only a partial view of the \gls{co} landscape and, more generally, of optimization techniques.
\new{Similarly, \citet{yu2024deepinsightsautomatedoptimization} and \citet{10.1145/3638530.3664086} \new{described} the evolution of \glspl{ea} to solve optimization problems from heuristic approaches to \glspl{llm}.}

Our systematic review differ from the aforementioned surveys in the following ways:
\begin{enumerate}[label=\emph{(\roman*)}] 
\item We systematically \new{reviewed} the topic using a rigorous methodology to identify, screen, include, and analyze relevant literature works. \new{We reported the process} following the \gls{prisma} \new{2020} guidelines.
\item We \new{limited} our focus to the usage of \glspl{llm} within \gls{co}, thus \new{excluding optimization areas outside} discrete combinatorial optimization, \new{as well as the use of optimization within} \glspl{llm}, \new{nor, more generally, the use of} \glspl{llm} \new{for mathematical problem-solving}.
\item We \new{addressed} the entire optimization process, \new{not focusing solely on specific parts}.
\item We \new{considered} a broader set of optimization techniques and algorithms, \new{not just} evolutionary techniques.
\item We \new{described} datasets, frameworks, tools, and metrics that \new{could} enhance applications of \glspl{llm} within \gls{co}.
\end{enumerate}

\section{Background} 
\label{sec:background}

\subsection{Large Language Models} \label{sec:large-language-models}
\glspl{llm} have fundamentally transformed the landscape of \gls{nlp} and related fields. At the core of this transformation is the Transformer architecture introduced by \citet{vaswani2017attention}, which revolutionized the \gls{nlp} field with the introduction of an improved self-attention mechanism built on top of the one proposed by \citet{bahdanau2014neural}. This mechanism allows models to weigh the importance of different words in a sentence, irrespective of their distance, leading to more contextually aware representations.

The Transformer architecture operates with an arrangement of two specified modules, an encoder and a decoder, designed to transform the input data into more abstract and general-purpose representations. Each encoder and decoder in the architecture is built from layers that perform specific functions. First, positional encoding is introduced to inject information about the order of input words into the model, which is crucial for processing sequences where the arrangement of elements carries meaning, as is common in \gls{nlp}. Following positional encoding, the embedding layer converts the input tokens (i.e., individual pieces of text, typically words or sub-words) into vectors of continuous numbers. This transformation turns linguistic information into a mathematical form that neural networks can process.
Once the input has been encoded and embedded, it passes through the Transformer's core mechanisms, i.e., the attention layers. The encoder relies on a self-attention mechanism to independently assess and emphasize different parts of the input data. This allows the model to understand each input segment in relation to the rest of the input, thus enhancing the context awareness of the system. In parallel, the decoder also employs a self-attention layer but focuses on generating the next output token in the sequence conditioned on the tokens generated so far. This ensures that each generated element is contextually aligned with the previous (i.e., already generated) text.
Additionally, cross-attention layers in the decoder access the encoder's output to guide the generation process. These layers allow the decoder to reference the full context provided by the encoder, ensuring that each output token is a contextually appropriate continuation or response to the specific provided input sequence. This comprehensive mechanism of encoding, self-attention, and cross-attention within the Transformer allows for highly effective processing and generation of text, making it a robust model for various complex language understanding and generation tasks.

Building on the innovative self-attention mechanism of the Transformer, subsequent models have been developed with specialized architectures tailored for different tasks. 
\texttt{ELMo} \cite{peters-etal-2018-deep} defined the usage on contextual embeddings. Encoder-only models, like \texttt{BERT} \cite{devlin-etal-2019-bert}, specialize in tasks that require a deep understanding of language context, making them ideal for applications like sentiment analysis and named entity recognition. On the other hand, decoder-only models, such as \texttt{GPT-3} \cite{NEURIPS2020_1457c0d6} and \texttt{GPT-4} \cite{bubeck2023sparks, openai2024gpt4}, excel in generating coherent and contextually appropriate text, powering applications in creative writing and dialogue systems.

After encoder and decoder-only models, the introduction of instruction-based models marks a significant evolution in \glspl{llm}. These models are trained to follow user-provided instructions \cite{ouyang2022training,JMLR:v25:23-0870}, making them versatile tools across various tasks without needing task-specific fine-tuning. Instruction-based training involves exposing the model to various tasks during training, along with corresponding instructions, thereby enabling the model to generalize from instructions at inference time. This approach has started the development of emerging abilities in \glspl{llm}, such as zero-shot learning capabilities \cite{kojima2022large}, complex reasoning strategies \cite{NEURIPS2022_9d560961,Besta_Blach_Kubicek_Gerstenberger_Podstawski_Gianinazzi_Gajda_Lehmann_Niewiadomski_Nyczyk_Hoefler_2024},  and the discovery of latent abilities that emerge as models sled \cite{wei2022emergent}.

Following the introduction of instruction-based models in \glspl{llm}, several state-of-the-art models have epitomized this approach, showcasing remarkable capabilities in handling a wide array of tasks directly based on user instructions. 
Some notable models in this field include
\texttt{PaLM} \cite{10.5555/3648699.3648939}, which represents a significant advancement in scaling up transformer-based architectures designed to perform well across diverse linguistic tasks. Its successor, \texttt{PaLM-2} \cite{anil2023palm}, builds upon this foundation with improved training techniques and larger model capacities, further enhancing its ability to understand and generate nuanced text based on instructions. \texttt{PaLM-E} \cite{10.5555/3618408.3618748} further extends the \texttt{PaLM} series by emphasizing efficiency in energy usage and processing speed, making it a more sustainable option for deploying sophisticated \gls{nlp} tasks at scale.

Another architecture is \texttt{LLaMA} \cite{touvron2023llama} and its successors, \texttt{LLaMa 2} \cite{touvron2023llama2} and \texttt{LLaMa 3} \cite{dubey2024llama3herdmodels}, which focus on achieving high cross-task effectiveness with relatively smaller model sizes, facilitating easier deployment and lower operational costs without compromising on capability. These architectures have demonstrated significant prowess in tasks requiring deep contextual understanding. 
\texttt{Mistral 7B} \cite{jiang2023mistral} and \texttt{Mixtral 8x7B} \cite{jiang2024mixtral} introduce a unique approach to model training called Mixture of Experts (MoE) that involves dynamic adjustments of model parameters based on task complexity, which enhances the model's adaptability and performance across different \gls{nlp} tasks.
\texttt{Gemini 1.5} \cite{geminiteam2024gemini} is another innovative model that integrates dual mechanisms of understanding and generation to improve interaction dynamics in conversational AI applications.
\texttt{Bard} \cite{manyika2023overview} focuses on incorporating broad, encyclopedic knowledge and the ability to update its understanding in real-time, making it exceptionally useful for applications that require up-to-date information.
Finally, \texttt{Claude} \cite{anthropic2024claude} distinguishes itself by its ethical training framework, prioritizing safety and fairness, setting a new standard in responsible AI development. 

These models exemplify the most advanced and state-of-the-art instruction-based learning, where each has been tailored to excel in standard benchmarks and improve specific aspects such as versatility, efficiency, and ethical considerations.

\subsection{Combinatorial Optimization} 
\label{sec:combinatorial-optimization}

\glspl{cop} are a class of optimization problems defined by discrete decision variables and the objective of finding one or more optimal solutions within a finite search space of solutions \cite{KARIMIMAMAGHAN2022393}.
Many of these problems are classified as NP-hard, meaning that, based on current knowledge, they require exponential time to be optimally solved.
Besides the prominent \gls{sat}, prototypical \glspl{cop} include the \gls{pfsp} \cite{PAGNOZZI2021100180}, the \gls{kp}, the \gls{gcp}, \new{and} the \gls{tsp} \cite{POP2024819}.
\gls{co} is also widely applied to tackle real-world problems across various domains. Examples include employee scheduling \cite{KLETZANDER2020104794,10.1007/978-3-031-62912-9_15}, machine scheduling \cite{moser2022exact,Lackner2023}, educational timetabling \cite{CESCHIA20231}, and automotive production \cite{winter2019solution}, to name a few.

\begin{figure}[ht]
    \centering
    \includegraphics[width=1\linewidth]{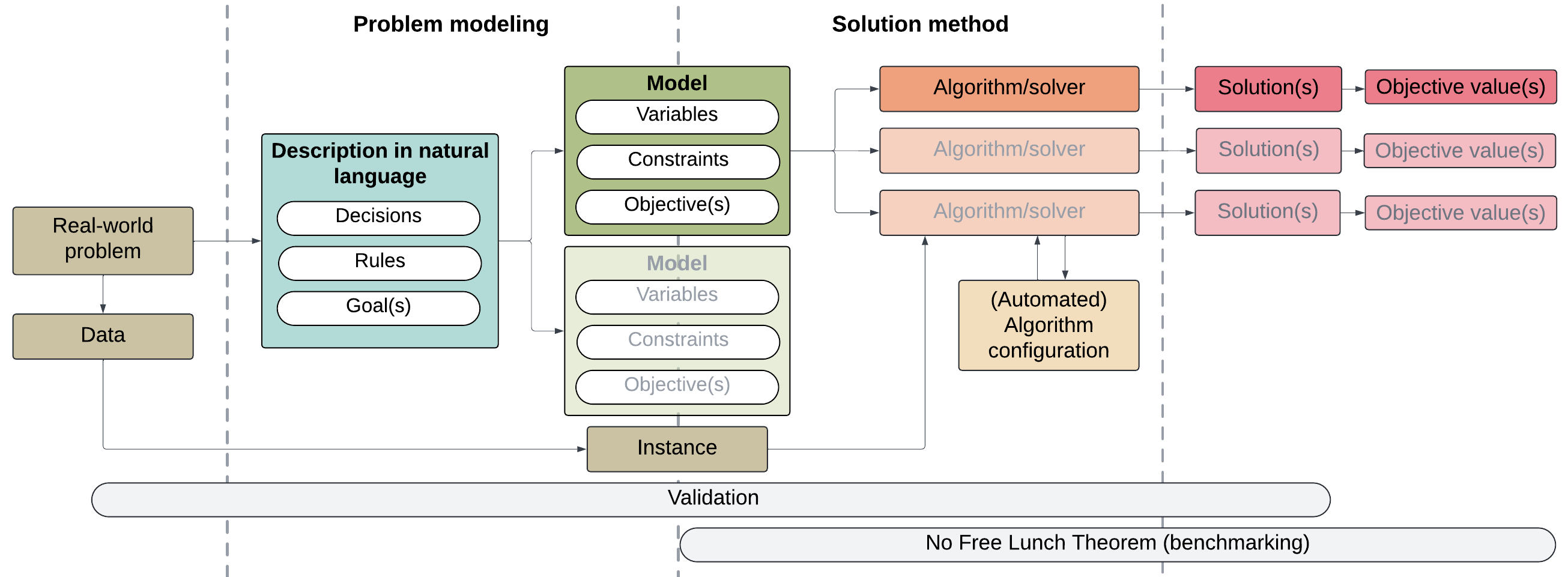}
    \caption{Overview of the steps for addressing an optimization problem.}
    \label{fig:optimization-process} 
\end{figure}

\Cref{fig:optimization-process} outlines the steps practitioners and researchers undertake to address an optimization problem \cite{tsouros2023holy,BENGIO2021405}. 
In the following, we detail the process specifically for the case of \gls{co}.

The first step regards the description in \gls{nl} of the decision-making process or real-world issue to be tackled.
Its specifications must be outlined to identify the decisions to be made, the rules that must be followed, and the goals to be achieved.

After framing the decision-making process, the practitioner must create a formal representation to enable a software solver or a custom algorithm to tackle it.
A given \gls{nl} description can correspond to multiple representations, called \emph{models} or \emph{problem formulations}.
Models are articulated in terms of 
\begin{enumerate*}[label=\emph{(\roman*)}] 
\item \emph{variables} (or decision variables), which represent the decisions to be made; \item \emph{constraints}, which are restrictions on the possible values of the variables and capture problem substructures \new{and/or business rules}; and \item at least one \emph{objective} (or fitness) \emph{function}, which assesses the quality of a solution. 
\end{enumerate*}
When the problem tackles more than one objective function, it is the case of multi-objective optimization. 
The development of an effective problem formulation is known as \emph{modeling}. How we define variables, objectives, and constraints determines the model type. \new{E.g.}, if variables assume integer values and constraints are linear, the model is an \gls{ilp}. Other model types, such as \gls{milp} and \gls{lp}, handle both discrete and continuous variables, while structured constraints can be represented using \gls{cp}.

While a problem formulation provides an abstract formal description of a decision process, an \emph{instance} (or problem instance) represents a specific case of the problem defined by concrete data.
A \emph{solution} to a problem instance is an assignment of values to all the decision variables. \new{As mentioned before, the solution quality is evaluated through an objective function, which provides one or more objective values, also referred to as scores or costs (typically numerical values).}
The search space of an instance encompasses all possible assignments of the variables; the subset of solutions that satisfy all constraints is called the set of \emph{feasible} solutions. \new{A solution which violates at least one constraint is called \emph{infeasible}. When no feasible solution exists for an instance, resulting in a logically inconsistent instantiation of the problem model, the model instance is also called infeasible.}

The search space is explored using \emph{solution methods}, which are algorithms or software solvers.
Solution methods for \glspl{cop} can be classified based on the completeness of their search (i.e., complete or incomplete methods) and how solutions are constructed (i.e., perturbative or constructive methods). 
Note that models and solution methods are strictly coupled together.

Complete methods exhaustively explore the search space, ensuring that 
\begin{enumerate*}[label=\emph{(\roman*)}]
\item If a solution exists, the complete method will eventually find it; 
\item When the search terminates, the best solution found is guaranteed to be optimal 
\end{enumerate*} (for this reason, complete methods are also called exact methods).
Examples of these methods include \gls{lp}, \gls{milp}, and \gls{cp} \cite{rossi-2006-handbook-cp}. For such methods, numerous widely accepted software tools exist, including IBM ILOG CPLEX and IBM CP Optimizer \cite{IBMsched2017}, Gurobi Optimizer \cite{gurobi}, OR-Tools \cite{ortools}, Gecode \cite{gecode}, and the solvers included in the MiniZinc distribution \cite{nethercote_minizinc_2007} together with interfaces and APIs available in common programming languages. 

However, given that most \glspl{cop} are NP-hard \cite{Garey1976117,KUBIAK202126},
 exhaustive exploration can result in an exponential increase in runtime w.r.t.  the size of the problem instance.
Additionally, in many real-world applications, it is not necessary to certify the optimality of a solution. Instead, obtaining a good, feasible solution in a reasonable amount of time is sufficient.
Incomplete methods address such necessities by exploring the search space non-exhaustively, often using stochastic approaches. 
When these methods are tailored to the problem, they are called heuristics; when they employ problem-independent strategies, they are referred to as \glspl{mh} \cite{marti-2024-50-history}.
Examples of \glspl{mh} include \gls{ts} \cite{glover-1997-ts-book}, \gls{sa} \cite{kirkpatrick-1983-sa,franzin-2019-sa-component}, \gls{grasp} \cite{laguna202320}, \gls{lns} \cite{shaw-1998-lns} and its adaptive version \cite{WINDRASMARA2022105903}, \gls{ga} \cite{holland-1992-genetic-algorithm}, \gls{brkga} \cite{LONDE2024}, and \gls{aco} \cite{dorigo-2004-aco}.
Differently from complete methods, widely accepted tools have not been available for incomplete methods, where researchers still rely on customized coding solutions \cite{swan-2022-mil, swan-2019-open-closed-princ, swan-2015-research-agenda} and on a sprout of possible frameworks and libraries \cite{parejo-2012-survey,lopes-silva-2018-survey}.

Constructive methods build solutions from scratch and iteratively set all variables according to specific policies.
Examples include \gls{cp} and \gls{aco}. 
Conversely, perturbative methods start from an existing complete solution and generate new solutions by modifying some variables.
Examples include methods based on the concept of neighborhood, such as \gls{ts} and \gls{sa}.

Algorithmic choices in solvers and algorithms can be addressed through \emph{automated algorithm configuration}, as advocated by the \gls{pbo} manifesto \cite{hoos-2012-pbo}. 
Such a paradigm calls for a shift of perspective in software development, where the design of components is approached through optimization. This involves a systematic exploration of design alternatives to select a configuration for the different components via automated tools (e.g., irace \cite{lopex-ibanez-2016-irace}). 

The process undergoes various types of \emph{validation}.
First, a practitioner must ensure that the models adhere as closely as possible to the real-world scenario, even though some assumptions are necessary to generalize the concepts (i.e., validation of specifications). 
Secondly, \new{model and technical validation} should be applied to check the correctness of the code, ensure it adheres to the model, and evaluate its performance efficiency.
Furthermore, to efficiently solve a problem, evaluating and comparing different algorithms, solvers, and models is essential. This comparison is necessary because of the \emph{No Free Lunch Theorem} \cite{wolpert-1997-no-free-lunch}, which states that no single algorithm is best suited for all instances of an optimization problem. Therefore, rigorous tests are conducted to determine the most effective approach for the specific problem, ensuring that the chosen solution method is robust and efficient (in other fields, this type of validation is addressed as a \emph{benchmarking} process).

\section{Methodology}
\label{sec:methodology}

\subsection{PRISMA}
\label{sec:prisma-methodology}

The \acrfull{prisma} guidelines are a well-known framework for reporting systematic reviews. The current version was proposed in 2020 \cite{Pagen71} and builds upon the \gls{prisma} 2009 formulation \cite{moher2009preferred}. This methodology evolved from the earlier \gls{quorom} guidelines \cite{MOHER19991896}, \new{which were firstly} introduced in 1999. 
\new{This evolution over time} has broadened \new{the} scope \new{of the guidelines}. While \gls{quorom} primarily focused on improving reports of meta-analyses in clinical trials, \gls{prisma} addresses systematic reviews that evaluate the effects of health interventions more broadly. 
These guidelines are sufficiently general to apply to reports of systematic reviews that evaluate other types of interventions \cite{00005792-202101290-00042,00005792-201711270-00083}, systematic reviews with objectives beyond evaluating interventions \cite{Heinrich2022,Ross2016}, and systematic reviews not related with the medical field \cite{SOPRANO2024103672,electronics10141611,NASSEN2023101980}.
\gls{prisma} can be used to report systematic reviews that involve result synthesis, such as meta-analyses and other statistical methods. It is also helpful for reviews identifying only a single eligible study and for mixed-methods approaches.

At its core, \gls{prisma} is composed of four main elements: the statement \cite{Pagen71}, the explanation and elaboration document \cite{Pagen160}, the checklist,\footnote{\url{https://www.prisma-statement.org/prisma-2020-checklist}} and the flow diagram.\footnote{\url{https://www.prisma-statement.org/prisma-2020-flow-diagram}} The statement introduces the purpose of the methodology. The checklist consists of 27 items across 7 sections, providing guidelines for writing a systematic review report. It should be used \new{alongside} the explanation and elaboration document, which offers additional reporting guidance for each item. The diagram shows the flow of information through the review phases.

In this \new{work}, we adopt, use, rely on, and refer to the \gls{prisma} \new{2020} guidelines \cite{Pagen71, Pagen160}.

\subsection{Terminology}
\label{sec:methodology_subsec:terminology}

Most of the terminology used within \gls{prisma} resources \cite{Pagen71, Pagen160} derives from its original field of application -- systematic reviews of health-related interventions -- which may be confusing for researchers from other disciplines. We particularly refer to the following definitions \cite{Pagen71}:
\begin{enumerate}[label=\emph{(\roman*)}]
    \item \emph{Study}: an experiment \new{including} a defined group of participants and one or more interventions and outcomes.
    \item \emph{Report}: a paper providing information about a particular study. Multiple reports may refer to the same study.
    \item \emph{Record}: the title or abstract (or both) of a report indexed in a database or website.
    \item \emph{Outcome}: a measurement event for participants in a study.
    \item \emph{Result}: the combination of a point estimate and precision measurement for an outcome.
\end{enumerate}

Since our systematic review covers a broader scope than health-related interventions, where we expect to find only individual eligible studies rather than multiple studies referring to a given experimental design, we will join the definitions of \emph{study} and \emph{report}, using \emph{study} to refer to papers in general. This approach will also apply to the definitions of \emph{result} and \emph{outcome}. We consider the term \emph{record} to encompass titles and abstracts of papers.
For example, the \gls{prisma} flow diagram explicitly distinguishes between records screened and reports sought for retrieval. We will adjust this to refer only to records as we define them. 

\subsection{Literature Collection} 
\label{sec:methodology_subsec:literature-collection}

\subsubsection{Process}
\label{sec:methodology_subsec:literature-collection_subsec:process}

\begin{figure}[ht]
    \centering
    \includegraphics[width=1\linewidth]{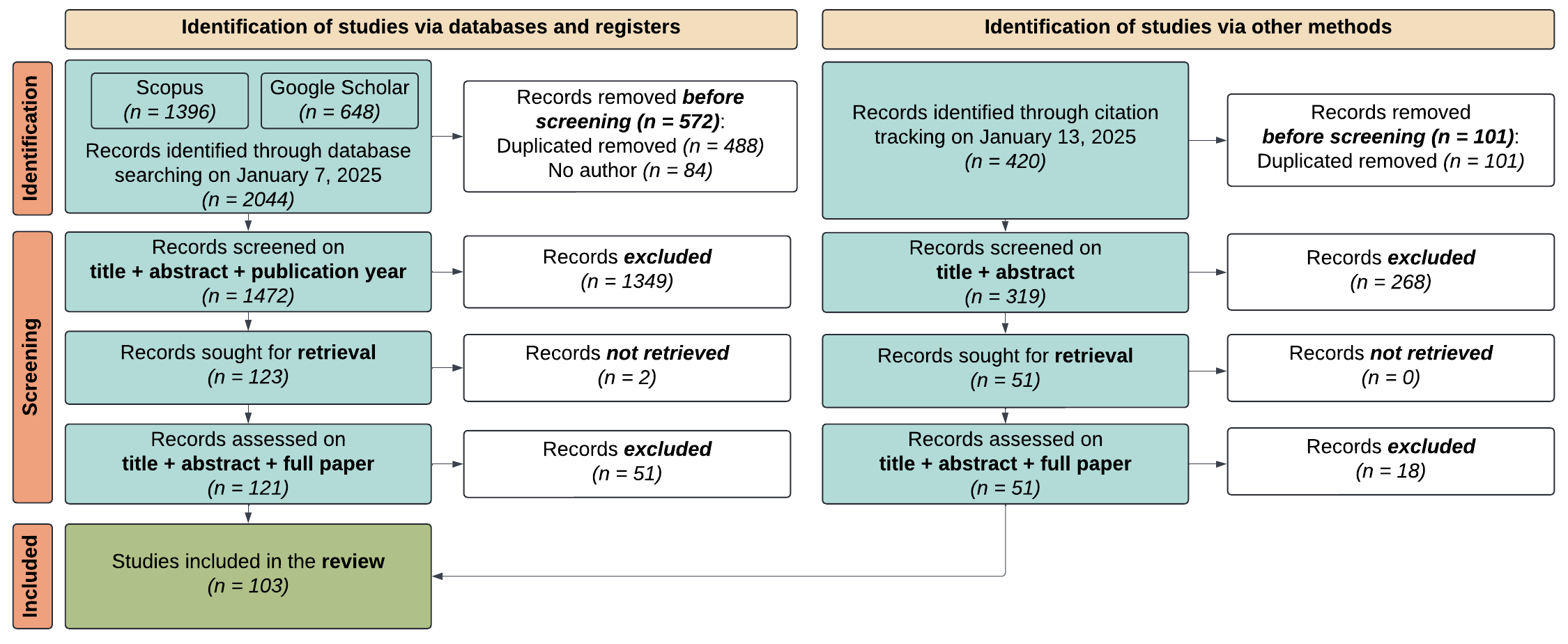}
    \caption{Literature identification, screening, and inclusion activities conducted following the \gls{prisma} guidelines.} 
    \label{sec:methodology_subsec:literature-collection_subsec:process_fig:flow-diagram}
\end{figure}

\new{Our systematic review includes studies up to the end of 2024.} The literature collection process, conducted according to \gls{prisma}, involves three main activities: identification, screening, and inclusion (\Cref{sec:methodology_subsec:literature-collection_subsec:process_fig:flow-diagram}). \new{To provide a comprehensive overview of our application of the \gls{prisma} guidelines, we \new{have included} the checklists in \Cref{sec:prisma-checklist}.}

\begin{figure}[ht]
    \centering
    \includegraphics[width=1\linewidth]{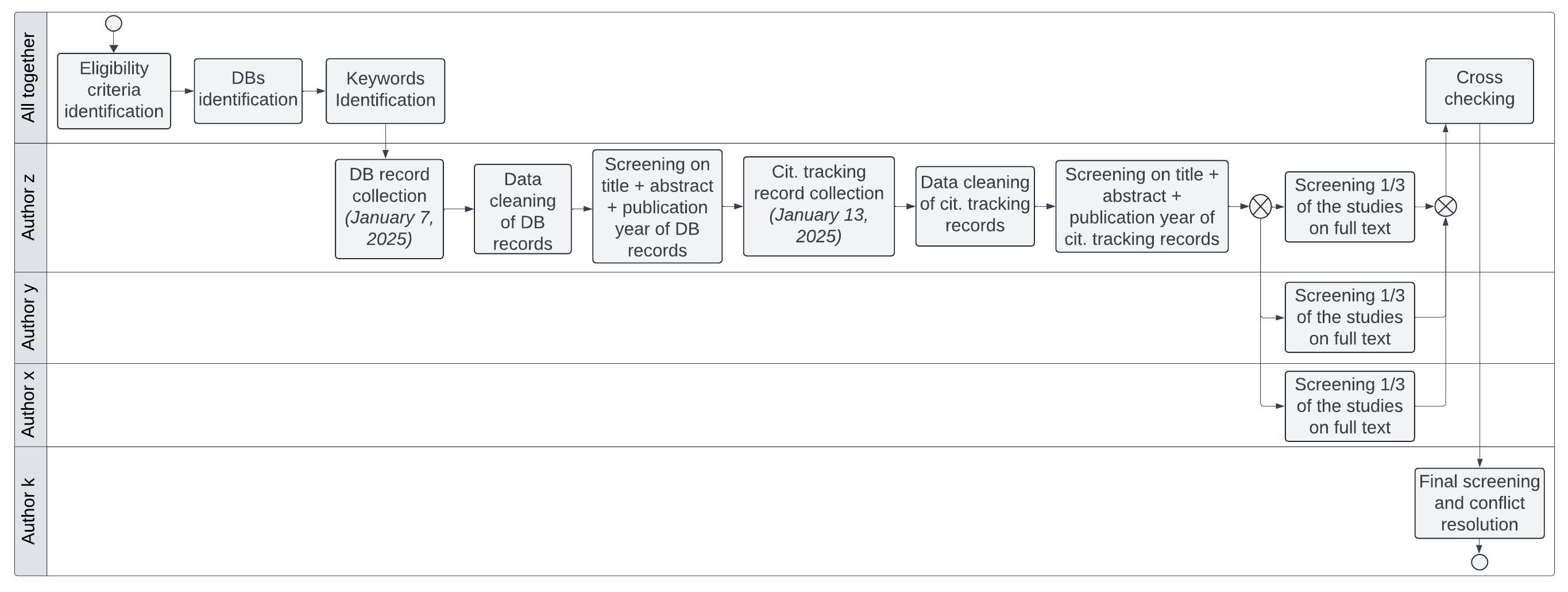}
    \caption{\gls{bpmn} diagram describing the identification and selection activities of records in the literature collection process.} 
    \label{sec:methodology_subsec:literature-collection_subsec:process_fig:bpmn}
\end{figure}

While the inclusion activity simply involves reporting the number of studies included in the systematic review, the task of identifying and screening records is more complex. \Cref{sec:methodology_subsec:literature-collection_subsec:process_fig:bpmn} reports a \gls{bpmn} diagram \cite{omg2011bpmn} \new{describing how we conducted these activities.}
\new{Initially, all authors worked together to define the keywords for searching records, establish eligibility criteria for study inclusion/exclusion, and select the databases for record identification.} Subsequently, one author sent queries to the selected databases to retrieve records and performed data cleaning to remove duplicates and records without authors. \new{Then}, the author screened the remaining records by reviewing titles, abstracts, and publication years in order to apply an initial filter. The resulting list was then augmented through citation tracking \cite{Cooper2017}.
After this phase, three authors independently read the full text of one-third of the studies each. They then decided whether to include each study based on the eligibility criteria. Subsequently, we collectively cross-checked the studies read by the other authors. The author not initially involved in full-text reading performed the final screening and resolved conflicts.

\subsubsection{Identification}
\label{sec:methodology_subsec:literature-collection_subsec:identification}

On \new{January 7, 2025}, we searched two popular databases to identify records: Scopus and Google Scholar. 
Scopus is an extensive, multidisciplinary database of peer-reviewed literature, while Google Scholar allows for broader searches, including pre\new{-}prints and other types of studies. 

To identify studies addressing the usage of \glspl{llm} \new{within} \gls{co}, we used \new{the following set of keywords: ``large language models'', ``generative artificial intelligence'', ``GPT'', ``optimization'', ``combinatorial optimization'', ``mathematical formulations'', ``metaheuristics'', ``constraint programming'', ``integer programming'', ``integer linear programming'', ``NL4Opt'', ``Ner4Opt''
}. 
These keywords were combined into \numqueries search queries, \new reported in \Cref{sec:methodology_subsec:literature-collection_subsec:identification_tab:queries} \new{along with} the number of matches found for each query. Besides the self-explanatory queries \new{(i.e., those including the relevant \gls{llm} and \gls{co} terms like “combinatorial optimization”)}, we also include the keywords “NL4Opt” and “Ner4Opt” in our Google Scholar search. These terms relate to a recent competition focused on extracting the meaning and formulation of an optimization problem from its textual description.
Note that Google Scholar only provides an estimated number of matches and that we used the advanced query functionality on Scopus to restrict the search to study titles, abstracts, and keywords using the \verb|TITLE-ABS-KEY(...)| construct. 
Using these queries, we retrieved approximately \numrecordsretrievedscholar records from Google Scholar and \numrecordsretrievedscopus from Scopus, for a total of \numrecordsretrieved records. This list required further processing because both databases contain duplicates. We found \numrecordsduplicatescholar duplicate records from Google Scholar, \numrecordsduplicatescopus from Scopus, \new{and \numrecordsduplicatesshared shared records,}  totaling to \numrecordsduplicates duplicates. 
Another case to consider is a small subset of records that lack an author string and should, therefore, be removed from the list of \numrecordsretrieved records. Specifically, we identified \numrecordsnoauthorscholar records without an author string on Google Scholar and \numrecordsnoauthorscopus records on Scopus. When combining the two lists, the number of distinct records without an author string is \numrecordsnoauthordistinct, as all four records found on Google Scholar are also present on Scopus.
The final number of distinct records from Google Scholar and Scopus is $\numrecordsretrievednodupsnoauthor$. 

\begin{table}[ht]
    \centering
    \caption{Queries used to retrieve records from Scopus and Google Scholar.}
    \label{sec:methodology_subsec:literature-collection_subsec:identification_tab:queries}
    \resizebox{1\textwidth}{!}{%
    \begin{tabular}{
    p{7.6cm} 
    C{0.9cm} 
    p{6.5cm} 
    C{0.9cm}} 
    \toprule
    \multicolumn{2}{c}{\footnotesize\textbf{Scopus}} & \multicolumn{2}{c}{\footnotesize\textbf{Google Scholar}} \\
    \cmidrule(lr){1-2} \cmidrule(lr){3-4}
    \footnotesize Query & \footnotesize Count & \footnotesize Query & \footnotesize Count \\
    \midrule
    \footnotesize{"large language models" and "optimization"} & \footnotesize \new{944} &  \footnotesize{"large language models" and "combinatorial optimization"} & \footnotesize \new{334}\\
    \footnotesize{"GPT" and "optimization"} & \footnotesize \new{394} & \footnotesize{"large language models" and "constraint programming"} & \footnotesize \new{104} \\
    \footnotesize{"large language models" and "constraint programming"} & \footnotesize \new{10} & \footnotesize{"large language models" and "mathematical formulations"} & \footnotesize \new{106} \\
    \footnotesize{"large language models" and "integer programming"} & \footnotesize \new{14} & \footnotesize{"large language models" and "metaheuristics"} & \footnotesize \new{88} \\
    \footnotesize{"large language models" and "integer linear programming"} & \footnotesize \new{15} & \footnotesize{"NL4Opt"} & \footnotesize \new{11} \\
    \footnotesize{"generative artificial intelligence" and "combinatorial optimization"} & \footnotesize 0  & \footnotesize{"Ner4Opt"} & \footnotesize 5\\
    \footnotesize{"large language models" and "combinatorial optimization"} & \footnotesize \new{14} \\
    \footnotesize{"GPT" and "combinatorial optimization"} & \footnotesize 5 \\
    \bottomrule
    \end{tabular}
    }
\end{table}

\subsubsection{Screening}
\label{sec:methodology_subsec:literature-collection_subsec:screening}

We define \numcriteria eligibility criteria to screen the list of studies referred by the \numrecordsretrievednodupsnoauthor records retrieved. Specifically, we define \numinccriteria inclusion (INCL) criteria and derive  \numexccriteria exclusion (EXCL) criteria: 
\begin{enumerate}[label = \textsf{INCL\arabic*}, leftmargin=4em]
    \item \label{sec:methodology_subsec:literature-collection_subsec:screening_criteria:inclusion-language} The study is written in English.
    \item \label{sec:methodology_subsec:literature-collection_subsec:screening_criteria:inclusion-optimization} The study focuses primarily on the usage of \glspl{llm} in the field of \gls{co}.
    \item \label{sec:methodology_subsec:literature-collection_subsec:screening_criteria:inclusion-year} The study has been published in 2016 or later. 
    \item \label{sec:methodology_subsec:literature-collection_subsec:screening_criteria:inclusion-type-of-paper} The study is in the form of a research article, a case report, a technical note, a narrative review, a systematic review, a position paper, a pictorial essay, or a PhD Dissertation.
\end{enumerate}
\begin{enumerate}[label = \textsf{EXCL\arabic*}, leftmargin=4em]
    \item \label{sec:methodology_subsec:literature-collection_subsec:screening_criteria:exclusion-language} The study is written in languages other than English.
    \item \label{sec:methodology_subsec:literature-collection_subsec:screening_criteria:exclusion-optimization} The study focuses primarily on optimization in the context of \glspl{llm}, solely on \gls{co}, solely on \glspl{llm}, or solely on optimization different from \gls{co}, such as \acrlong{col}.
    \item \label{sec:methodology_subsec:literature-collection_subsec:screening_criteria:exclusion-year} The study has been published in 2015 or earlier. 
    \item \label{sec:methodology_subsec:literature-collection_subsec:screening_criteria:exclusion-type-of-paper} The study is not in the form of a research article, a case report, a technical report, a narrative review, a systematic review, a position paper, a pictorial essay, or a PhD Dissertation.
\end{enumerate}
In summary, we focus on how \glspl{llm} are used in optimization, specifically targeting studies that address the subfield of \gls{co} (\ref{sec:methodology_subsec:literature-collection_subsec:screening_criteria:inclusion-optimization}); consequently, we exclude those taking the opposite perspective (\ref{sec:methodology_subsec:literature-collection_subsec:screening_criteria:exclusion-optimization}). \new{Note that some of the exclusion criteria may appear redundant w.r.t. the inclusion ones; this is due to a rigorous application of the \gls{prisma} guidelines in reporting on our activity of review.}

The formulation of these criteria is based on the concept of \lq\lq primary focus\rq\rq{} on the use of \glspl{llm} in the field of \gls{co}. According to our criteria, a study focusing on the use of \glspl{llm} in \glspl{cop} should either:
\begin{enumerate}[label=\emph{(\roman*)}] 
    \item Describe the usage of \glspl{llm} within one of the tasks of the overall optimization process.
    \item Address specific aspects of a \gls{cop}.
    \item Propose improvements or new approaches for a given optimization paradigm.
    \item Outline use cases, perspectives, and/or potential future applications.
    \item Involve specific \glspl{llm} and, if applicable, provide metrics and/or reference datasets.
\end{enumerate}
We believe that this notion of \lq\lq primary focus\rq\rq{} allows us to consider not only research papers but also reviews and surveys, which are equally important as they may provide general directions and perspectives (\ref{sec:methodology_subsec:literature-collection_subsec:screening_criteria:inclusion-optimization} and \ref{sec:methodology_subsec:literature-collection_subsec:screening_criteria:inclusion-type-of-paper}). This also helps clarify the rationale behind \ref{sec:methodology_subsec:literature-collection_subsec:screening_criteria:exclusion-optimization} and \ref{sec:methodology_subsec:literature-collection_subsec:screening_criteria:exclusion-type-of-paper}; if any of the components described above is missing, we exclude the study from consideration. 
For example, let us hypothesize a study on how an \gls{llm} could enhance an evolutionary procedure for solving a \gls{cop}; \new{this study would align with our inclusion criteria. Conversely, consider a study proposing the use of a \gls{mh} to generate prompts for \glspl{llm}; such a study would not meet our criteria and would be excluded.}
Concerning the criteria for the publication year (\ref{sec:methodology_subsec:literature-collection_subsec:screening_criteria:inclusion-year}/\ref{sec:methodology_subsec:literature-collection_subsec:screening_criteria:exclusion-year}),  
\new{we selected 2016 as the earliest publication year to ensure comprehensive coverage for our systematic review. This choice aligns with key developments in \glspl{llm}, particularly following the ``Attention Is All You Need'' paper~\cite{vaswani2017attention}, which introduced the Transformer architecture in 2017 (\Cref{sec:large-language-models})}

Regarding the types of publications considered, \new{we have identified a subset that aligns with our objectives} (\ref{sec:methodology_subsec:literature-collection_subsec:screening_criteria:inclusion-type-of-paper}). \new{We therefore exclude all other publication types} not specified therein (\ref{sec:methodology_subsec:literature-collection_subsec:screening_criteria:exclusion-type-of-paper}). 

We screened the identified records (\numrecordsretrievednodupsnoauthor) according to our eligibility criteria. 
\new{We removed \numrecordsremovedautomatically records because they were published earlier than 2016 (\ref{sec:methodology_subsec:literature-collection_subsec:screening_criteria:inclusion-year}/\ref{sec:methodology_subsec:literature-collection_subsec:screening_criteria:exclusion-year}).} 
\new{A total of \numrecordsretrievednollmornoco records tackled solely \glspl{llm}, solely optimization, or a type of optimization different from \gls{co}}. Additionally, we excluded \numrecordsretrievedoptforllm records that addressed optimization in the context of \glspl{llm} (\ref{sec:methodology_subsec:literature-collection_subsec:screening_criteria:inclusion-optimization}/\ref{sec:methodology_subsec:literature-collection_subsec:screening_criteria:exclusion-optimization}). We also removed \numrecordsretrievedwrongpapertype records because they referred to studies of the wrong type (\ref{sec:methodology_subsec:literature-collection_subsec:screening_criteria:exclusion-type-of-paper}). In total, we removed \new{1,349} records, resulting in a final count of \numrecordsfinalbeforecitationtracking records.  

On \new{January 13, 2025}, we performed citation tracking \cite{Cooper2017} on the \numrecordsfinalbeforecitationtracking records that remained after the initial screening.  
Specifically, we traced all the citations received by the collected studies referred to in the records, identifying a total of \numrecordsfoundthroughcitationtracking records. We then checked for duplicates in this set, identifying \numrecordsduplicatescitationtracking duplicates. Thus, we ended up with \numrecordsduplicatescitationtrackingdistinct distinct records from citation tracking. 

We manually addressed each record, evaluating them according to our criteria as we did previously. Among the \numrecordsduplicatescitationtrackingdistinct distinct records, we found that \numrecordsfoundthroughcitationtrackingnollmornoco tackled solely \glspl{llm}, solely optimization, or a type of optimization different from \gls{co} (\ref{sec:methodology_subsec:literature-collection_subsec:screening_criteria:exclusion-optimization}). Additionally, \numrecordsfoundthroughcitationtrackingoptforllm records referred to studies that tackled optimization in the context of \glspl{llm}. We excluded \numrecordsfoundthroughcitationtrackingwrongpapertype records because they referred to studies of the wrong type (\ref{sec:methodology_subsec:literature-collection_subsec:screening_criteria:exclusion-type-of-paper}) \new{and \numrecordsfoundthroughcitationtrackingwronglanguage study because it was not written in English (\ref{sec:methodology_subsec:literature-collection_subsec:screening_criteria:exclusion-language})}. In total, we removed a total of \numrecordscitationtrackingremovedmanually records, leaving us with \numrecordscitationtrackingfinal records. By examining these citations, we uncovered additional highly related or complementary sources \new{w.r.t.} the initial set of records, thereby broadening and enriching our literature review.

In summary, \new{combining the records retrieved through databases (\numrecordsfinalbeforecitationtracking) and those obtained through citation tracking (\numrecordscitationtrackingfinal), we had a total of \numstudies records. } 

\subsubsection{Inclusion}
\label{sec:methodology_subsec:literature-collection_subsec:inclusion}

We read the full text of each of the \numstudies studies referred to by the records obtained after the screening activity to decide which ones to include in our systematic review. This involved a more thorough evaluation of each study concerning our eligibility criteria.
\new{Below are a few examples: }
\begin{itemize}[label={--}, leftmargin=2em]
    \item \citet{amarasinghe2023aicopilot} \new{proposed} to automate the formulation of optimization problems from \gls{nl} descriptions, using fine-tuned \glspl{llm} to generate modular code for complex real-world business optimization scenarios. In this study, \glspl{llm} are used to enhance the formulation of a \gls{cop}, thereby satisfying all our inclusion criteria.
    \item \citet{10.1145/3583133.3596401} used an \gls{llm} to identify and decompose six well-performing swarm algorithms for continuous optimization. Despite the involvement of \glspl{llm}, we exclude this study because it focuses exclusively on continuous optimization problems (\ref{sec:methodology_subsec:literature-collection_subsec:screening_criteria:exclusion-optimization}).
    \item \citet{wang2024pivoting} proposed a survey on the usage of \glspl{dgm} for \glspl{cop}. We exclude this study since \glspl{dgm} are not \glspl{llm} (\ref{sec:methodology_subsec:literature-collection_subsec:screening_criteria:exclusion-optimization}).
\end{itemize}
Thus, the final number of studies included in this systematic review is \numstudiesincluded \xspace\new{(i.e., 2 studies were not retrieved and 69 were not included \new{as per inclusion/exclusion criteria}, thus \numstudies = \numstudiesincluded + 2 + 69)}.

\Cref{tab:count-included-papers} summarizes the characteristics of the included studies. 
\new{Please note that when feasible, the characteristics reported are those of the peer-reviewed version of the studies, i.e., if a study was previously published in a pre-print form and later in a peer-reviewed version, we consider the latter here.}
The majority \new{(82 studies, 80\%)} were published \new{in} 2024.
The remaining studies were published in \new{2023 (17 studies, 16\%) and in 2022 (4 studies, 4\%)}. 
Considering their identification approach, \new{70 (68\%)} were retrieved from databases; additionally, we included \new{33 (32\%)} studies referred to by records found through citation tracking. Regarding the publication venues, approximately \new{50} studies (\new{49}\%) were disseminated through pre-print distribution services such as \arxiv. The remaining studies were peer-reviewed, with \new{25} (\new{24\%}) published in journals and \new{28} (\new{27}\%) in conference proceedings. 
Focusing on the type of each study, as specified in our inclusion criteria (\ref{sec:methodology_subsec:literature-collection_subsec:screening_criteria:inclusion-type-of-paper}), \new{84} of them (approximately \new{82}\%) are research studies addressing specific research questions and/or experimental settings. Additionally, we included \new{11} (\new{10}\%) literature reviews, one (\new{1}\%) technical report, and \new{7} (7\%) position papers, as these provide perspectives, guidelines, or future directions. 

\begin{table}[ht]
\caption{Overview of the \numstudiesincluded studies included in our review.}
\label{tab:count-included-papers}
\centering
\resizebox{\textwidth}{!}{%
\begin{tabular}{
C{1.1cm} 
C{0.9cm}  
C{0.7cm}
C{2cm}
C{1.06cm}
C{1.4cm} 
C{1.25cm} 
C{1.2cm} 
C{1cm} 
C{0.9cm} 
C{1.05cm} 
} 
\toprule
\footnotesize\textbf{Year}& \footnotesize\textbf{Total} & \multicolumn{2}{c}{\footnotesize\textbf{Identification}} & \multicolumn{3}{c}{\footnotesize\textbf{Venue}} & \multicolumn{4}{c}{\footnotesize\textbf{Publication Type}}\\ 
\cmidrule(lr){3-4} \cmidrule(lr){5-7} \cmidrule(lr){8-11}
 & & \footnotesize{DB} & \footnotesize{Citation Tracking} & \footnotesize{Journal} & \footnotesize{Conference} & \footnotesize{Pre-print} & \footnotesize{Research} & \footnotesize{Review} & \footnotesize{Report} & \footnotesize{Position} \\
\midrule
\footnotesize 2022 &\footnotesize 4  &\footnotesize 3  & \footnotesize 1 &\footnotesize 0 & \footnotesize 1 & \footnotesize 3  & \footnotesize 4  & \footnotesize 0 & \footnotesize 0 & \footnotesize 0 \\
\footnotesize 2023 &\footnotesize \new{17} &\footnotesize \new{16} &\footnotesize \new{1} &\footnotesize 3 &\footnotesize 4 & \footnotesize \new{10} & \footnotesize \new{15} & \footnotesize 0 & \footnotesize 1 & \footnotesize 1 \\
\footnotesize 2024 &\footnotesize \new{82} &\footnotesize \new{51} &\footnotesize \new{31} &\footnotesize \new{22} & \footnotesize \new{23} & \footnotesize \new{37} & \footnotesize \new{65} & \footnotesize \new{11} & \footnotesize 0 & \footnotesize \new{6} \\
\midrule
\footnotesize\textbf{Total} &\footnotesize \new{103} &\footnotesize \new{70} &\footnotesize \new{33} &\footnotesize \new{25} &\footnotesize \new{28} &\footnotesize \new{50} &\footnotesize \new{84} &\footnotesize \new{11} &\footnotesize 1 &\footnotesize \new{7} \\
\bottomrule
\end{tabular}}
\end{table}

\section{Analysis}
\label{sec:analysis}

In this section, we classify and discuss the \numstudiesincluded studies. \new{The analysis is conducted in terms of optimization process, \glspl{llm}, benchmark datasets, and application domain. }  
Each selected study is described in \Cref{sec:description-of-selected-studies}.

\subsection{Optimization Process}
\label{sec:optimization-process-analysis}

\new{In this section, we report the analysis of the included studies w.r.t. the optimization process, both considering the general task (e.g., problem modeling) and the related activities (e.g., domain knowledge). A detailed tabular overview is available in \Cref{app:analysis-co}}. 

\new{This analysis} accounts for \new{86} out of the \numstudiesincluded studies \new{(83\%)}. \new{The omitted studies} provide perspective and general direction on the topic (\Cref{sec:position-papers}). 
\new{In the counts related to the number of studies per step or task, a study addressing multiple tasks within the same step is counted only once for that step}. For instance, \citet{li2023synthesizing} addresses both entity recognition and model creation within the problem modeling step. Therefore, it is counted as 1 in the problem modeling step. If a study addresses multiple steps, it is counted as 1 in each step. For instance, \citet{tsouros2023holy} addresses activities both in the problem modeling and solution method steps, thus it is counted as 1 in each of them. 
Most of the studies (\solutionMethods out of \numstudiesincluded) focus on activities related to the solution method, representing \new{63\%} of the total. The problem modeling task follows closely, with \problemModeling studies, accounting for \new{37\%}. A limited number of studies are related to benchmarking (\benchmarking, \new{7\%}) and validation (\validation, \new{9\%}).
In the following sections, we outline how \glspl{llm} are used within each step of the process.

The optimization process could benefit from the application of \glspl{llm} in several ways, as advocated by \citet{wasserkrug2024large}, who argues for their integration throughout the entire process. While such a holistic approach is possible, only \moreTasks studies (\new{23\%}) address multiple steps across the pipeline. 

\subsubsection{Problem Modeling}

Problem modeling aims to translate problems, expressed in \gls{nl}, into mathematical and computational models. This step is foundational, as the quality of the models influence the success of optimization. 

A specialized understanding of the domain in which the \gls{cop} is located is crucial for effective modeling and for ensuring realistic and applicable solutions. Traditionally, human experts are necessary in this phase as they bring deep knowledge of industry-specific rules, regulations, etc. In such context, \gls{llm} can help by extracting implicit knowledge from the real-world data and integrating it into solution processes  \cite{LIU2023100520,ai5010006,zhao2024survey,10818476} \new{and to generate solutions \cite{delarosa2024trippaltravelplanningguarantees,10.1145/3664646.3665084,Da2024}. \citet{li2025graphteamfacilitatinglargelanguage} integrates the domain knowledge, with both entity recognition and the generation of solution code}.

In the given problem context, it is crucial to identify the key entities and their interrelations to properly model variables, constraints, and objectives. \glspl{llm} can be particularly useful for recognizing and labeling these elements within a \gls{nl} description of the problem  \cite{wang2023opdnl4opt,doan2022vtccnlp}. This capability can enhance the accessibility and usability of the models, enabling non-domain experts to solve significant problems across various industries. Such activity is frequently paired with the extraction of domain knowledge \cite{ALIPOURVAEZI2024110574,smartcities7050094}, the creation of the model \cite{li2023synthesizing,ahmed2024lm4opt,ning2023novel,he2022linear,ramamonjison-etal-2022-augmenting,jang2022tag,10711695,10803039}, and toward code generation \cite{huang2024words,li2025graphteamfacilitatinglargelanguage}. \new{Note that \citet{10803039} also addresses solution generation}. As mentioned, \glspl{llm} can produce optimization models \cite{gangwar2023highlighting,regin_et_al:LIPIcs.CP.2024.25} and assist users in doing so, as evidenced, \new{for instance}, by the conversational agents developed by \citet{abdullin-etal-2023-synthetic}. \new{The generation of an optimization model is often accompanied by the production of the related code \cite{huang2025orlmcustomizableframeworktraining,10738100,lawless2024i,xiao2024chainofexperts}. Several studies have jointly addressed entity recognition, model creation, and code generation \cite{tsouros2023holy,ju-etal-2024-globe,Mostajabdaveh04112024,hao2025planningrigorgeneralpurposezeroshot,jiang2025llmopt,yang2024optibenchmeetsresocraticmeasure,ahmaditeshnizi2024optimus03usinglargelanguage,ahmaditeshnizi2024optimus,zhang-etal-2024-solving}, with \citet{michailidis_et_al:LIPIcs.CP.2024.20} also addressing solution generation.}

A total of \problemModeling studies focused on problem modeling, encompassing the identification of domain knowledge (\domainKnowledge studies), entity recognition (\entityRecognition studies), and model creation (\modelCreation studies).

\subsubsection{Solution Methods}

Solution methods refer to strategies and algorithms used to find (near) optimal solutions to \glspl{cop}. The choice of the solution method is linked to the problem model and is deemed critical, as it determines the efficiency, scalability, and accuracy of the results. Thus, optimization researchers' expertise is essential to select the right solving strategy 
and to design algorithms and their components. 
\glspl{llm} have been adopted in the design of solution methods, especially through code generation; \new{there exists one case where \glspl{llm} have been used for algorithm selection \cite{10.24963/ijcai.2024/579}}. Complete algorithms have been designed starting from \gls{nl} description of solution methods for \gls{cp} \cite{amarasinghe2023aicopilot,voboril2025generatingstreamliningconstraintslarge} and \new{\gls{milp} \cite{li2024foundationmodelsmixedinteger,li2023large,almonacid2023automatic}}. \citet{liu2023algorithm} and \citet{vanstein2024intheloophyperparameteroptimizationllmbased} used \glspl{llm} to generate heuristic algorithms within an evolutionary framework, \new{with the latter also including in the loop parameter tuning using an automated tool (SMAC} \cite{JMLR:v23:21-0888}). \new{On the contrary, a few studies implemented a \gls{llm}-based parameter tuning \cite{a17120582,10720437,DBLP:conf/esann/MartinekLG24,lawless2024llmscoldstartcuttingplane}}.
\citet{ye2024large} and \citet{liu2024evolution} developed heuristics for a hyper-heuristic framework. \citet{Romera-Paredes2024} introduced FunSearch, a method for discovering new heuristics for the online bin packing problem, achieving improvements over widely used baselines. Following, many other studies tackled the topic of generating code for heuristics  \cite{liu2023algorithm,10.1007/978-3-031-70068-2_12,yatong2024tseohedgeservertask,chen2024uberuncertaintybasedevolutionlarge,liu2024llm4adplatformalgorithmdesign,yu2024autornetautomaticallyoptimizingheuristics,yao2024multiobjectiveevolutionheuristicusing,sun2024autosatautomaticallyoptimizesat}.
\citet{mao2024identify} integrated \glspl{llm} into mutation and crossover operators for \glspl{ga}.
Most of the generated code is written in Python (\counterPythonCodeGeneration, note that overall \counterPythonTotal studies use Python). 

Generating solutions that satisfy the constraints and are diverse enough is complex and sometimes even unfeasible \cite{Verduin2023}. \glspl{llm} have been asked to directly produce solutions \cite{RePEc:spr:snopef:v:4:y:2023:i:4:d:10.1007_s43069-023-00277-6,SAKA2024100300,chin2024learning,buildings13071772,yang2024large,huang2024multimodal,10628050,Wang_2024,jiang2024largelanguagemodelscombinatorial,bohnet2024exploringbenchmarkingplanningcapabilities,10675146,huang2024exploringtruepotentialevaluating,guo2024optimizinglargelanguagemodels,liu2024large1,liu2024large,teukam2024integrating,a17120582,reinhart-2024,hu2024scalableaccurategraphreasoning,make6030093,10704489}. 
Such an approach is useful, for instance,in \glspl{mh}, where complete solutions are required as a starting point or to be mutated during the evolutionary process, and in population-based optimization algorithms, where a large number of (possibly diverse) solutions are required at each iteration. Finally, code generation and solution generation have been tackled together by 2 works \cite{Khan_2024,borazjanizadeh2024navigatinglabyrinthevaluatingenhancing}.

A total of \solutionMethods studies focused on solution methods, encompassing code generation (\codeGeneration studies), solution generation (\solutionGeneration studies), \new{parameter tuning (\parameterTuning), and algorithm selection (\algorithmSelection)}. 

\subsubsection{Validation}

\new{The validation step ensures that the model, the code, and the produced solutions align with the problem requirements and meet end-users' needs. }
 This process often requires substantial human input, as domain experts are typically involved in an iterative feedback loop to critically review the models and solutions to verify that all specifications have been adequately addressed. As \glspl{llm} have been proven to perform fairly well in such tasks \cite{10329992}, \new{they have also been applied to optimization}.  \citet{huang2024words} employed an \gls{llm} to validate solutions in an end-to-end framework (spanning from problem modeling to technical validation). Conversely, \citet{chen2023diagnosing} acted at the model level, employing a \gls{llm} agent to validate the (MI)LP models. 
\new{Specifically, they} integrated \glspl{llm} in diagnosing \new{inconsistent (i.e., over-constrained) \gls{milp} optimization models, trying to isolate the minimal subset of linear constraints (including variable bounds) that makes the model instance infeasible}. 
\new{Validation has also been addressed by other studies, reported also in other steps of the optimization process \cite{10704489,make6030093,xiao2024chainofexperts,zhang-etal-2024-solving,Mostajabdaveh04112024,hao2025planningrigorgeneralpurposezeroshot}.}
Note that we do not address bug checks in \gls{llm}-generated code as proper technical validation (e.g., during code generation for the solution method step) since we assume such checks are inherently part of that activity (if not state otherwise).
A total of \validation study focused on validation (\solutionValidation on solution validation and \modelValidation on model validation). 

\subsubsection{Models and Algorithm Types} 

Since the optimization process can be clearly outlined in terms of tasks and steps, it is crucial to recognize that modeling choices, solution methods, and validation procedures are closely interconnected. This section presents an overview of the algorithms and solvers considered in the studies. Please refer to the sections related to the single steps.

Considering the underlying model and algorithm type, the majority of studies (\counterLP) focus on \gls{lp} \new{and} \gls{milp} (\counterMILP). Other exact methods involve \gls{cp} (\counterCP) and \gls{ilp} (\counterILP). (Meta-)Heuristic algorithms account for \new{20} studies; among them, \counterHeuristic focus on heuristics, \counterHyperHeuristic focuses on hyper-heuristics, \counterGeneticAlgorithm on \glspl{ga}, and \counterEvolutionaryAlgorithm on \glspl{ea}. Furthermore, \counterMultiObjective study explores the capabilities of \gls{llm} w.r.t. multi-objective \gls{co} \cite{liu2024large}. \new{One study deals with SAT -- even though \glspl{llm} are used to generate heuristics for their solution \cite{sun2024autosatautomaticallyoptimizesat}.}

One reason for the prominence of \gls{lp} in these studies is the \gls{nl4opt} competition. The primary objective of this competition was to assess whether models could generalize to unseen problem domains, with an emphasis on two sub-tasks: named entity recognition for \gls{lp} components and \gls{lp} model generation. For more detailed information on the competition, readers are forwarded to the report by \citet{ramamonjison2023nl4opt}, while a description of the related dataset is provided in \Cref{sec:datasets}.

\subsubsection{Benchmarking}

In \gls{co}, benchmarking involves evaluating and comparing the performance of algorithms and techniques. The objective is to determine how effective a solution method is and, potentially, understand the reasons behind its performance.
Conventional comparison methods involve gathering numerical data from algorithm runs and reporting them in the form of tables and boxplots, alongside considerations of computational times and statistical tests (such as the Friedman test) to account \new{for stochasticity.} 
For decades, researchers have emphasized the need for additional tools to better understand the behavior of optimization algorithms \cite{10.1145/937503.937505}. One such tool is related to \glspl{stn}, a graph-based tool for the visualization of \gls{mh} behavior \cite{OCHOA2021107492}. While these visualizations are very informative, they require prior knowledge to be produced (e.g., parameters for search space partitioning) and interpreted (e.g., which algorithm is superior). To bridge this gap, \citet{sartori2024large} enriched \glspl {stn} with \gls{llm}-generated explanations and suggestions. 

\new{The visual investigation of solutions is enabled by \citet{Da2024}, who also allows solution validation. The visual capabilities of \gls{llm} are also explored by \citet{make6030093}, who directly employ them to solve problems, presenting instances in visual form.}

An increasingly relevant topic in \gls{co} is explainability of results \cite{swan-2022-mil} as a way for enhancing confidence and trust in the solutions. Interesting questions include identifying which solution components most significantly influence the final result and understanding the characteristics of high-quality solutions, among others. Providing informative answers to these questions requires in-depth knowledge of the application domain, the solution method, and the specific instance being addressed. To reduce this human-intensive task, \citet{10.1007/978-981-97-2259-4_3} proposed a post hoc \gls{llm}-based explanation framework aimed at clarifying the decision-making process of \glspl{vrp}. \new{Similarly, also other study included an explainability level \cite{hu2024scalableaccurategraphreasoning,RePEc:spr:snopef:v:4:y:2023:i:4:d:10.1007_s43069-023-00277-6}.}

\new{A total of 7 studies focused on benchmarking (3 on visual analysis and 4 on explainability)}. 

\subsection{Large Language Models}
\label{sec:large-language-models-analysis}

\new{The \glspl{llm} available today exhibit different capabilities: some are designed for processing instruction-like prompts, others specialize in generating programming code, etc. Understanding the role of each \gls{llm} in the reviewed studies is valuable for developing future approaches to solving \glspl{cop}, complementing the findings in \Cref{sec:optimization-process-analysis}.}

The \numllm  \glspl{llm} used in the \numstudiesincluded studies are based on \numarchitectures different architectures. 
Despite the diverse capabilities of the architectures employed, most \glspl{llm} have been used in \gls{co} to enhance problem modeling 
and solution methods. 
\new{In the following, we provide a description of how \glspl{llm} are used within the included studies, considering their architectures, general tasks, and related activities. Note that the numbers for activities may not sum to those at the main task level, as a given \gls{llm} can be used for multiple purposes. For a tabular overview of the \glspl{llm}, we refer the interested reader to \Cref{app:llms}.}

\subsubsection{LLM Architectures}
\label{sec:large-language-models-analysis:architectures}

\numarchgenerative out of the \numarchitectures architectures focus on generative approaches. 
\new{Introduced in 2022, \texttt{GPT-3.5} \cite{NEURIPS2020_1457c0d6} is particularly effective for tasks requiring fluent text generation. \texttt{GLM} \cite{du2022glm} emerged as a balanced approach to generative tasks, leveraging both bidirectional and autoregressive capabilities. During 2023, several architectures were released: \texttt{LLaMa 2} \cite{touvron2023llama2}, optimized for efficiency and suitable for tasks involving scalability and quick inference times; \texttt{PaLM 2} \cite{anil2023palm}, which performs strongly in scenarios requiring comprehension and contextual text generation; \texttt{Mistral} \cite{jiang2023mistral}, designed for efficient inference and competitive performance in text generation, making it suitable for resource-constrained deployments; and \texttt{Qwen} \cite{bai2023qwentechnicalreport}, focusing on general-purpose text generation with multilingual support and fine-tuned reasoning capabilities.}
\new{In the same year, \texttt{GPT-4} \cite{openai2024gpt4} appeared as a particularly strong model for fluent text generation, alongside \texttt{LLaMa 3} \cite{dubey2024llama3herdmodels}, which is optimized for instruction-following and structured text understanding. \texttt{DeepSeek} \cite{deepseekai2024deepseekllmscalingopensource} also adopts generative approaches with a particular focus on the Chinese language, while \texttt{Claude 3.5} \cite{anthropic2024claude35} excels in dialogue-based applications and knowledge-intensive tasks. \texttt{Qwen2} \cite{yang2024qwen2technicalreport} extends its predecessor with multilingual capabilities and refined reasoning. \texttt{Cohere}’s \texttt{Command-R+} \cite{cohere2024commandrplus} is optimized for \gls{rag}, improving factual accuracy and knowledge recall. Finally, \texttt{Yi} \cite{yi2024open} is tailored for diverse text generation tasks, integrating efficient training strategies for improved performance.}
A total of 62 studies rely on these architectures, namely 
\new{24 on \texttt{GPT-3.5} (2022), 2 on \texttt{GLM} (2022), 9 on \texttt{LLaMa 2} (2023), 3 on \texttt{PaLM 2} (2023), 2 on \texttt{Mistral} (2023), 3 on \texttt{Qwen} (2023), 43 on \texttt{GPT-4} (2023), 7 on \texttt{LLaMa 3} (2024), 6 on \texttt{DeepSeek} (2024), 7 on \texttt{Claude 3.5} (2024), 1 on \texttt{Qwen2} (2024), 1 on \texttt{Cohere} (2024), and 1 on \texttt{Yi} (2024).}

A total of \numarchtextual architectures are designed to handle textual input. 
\new{\texttt{T5} \cite{raffel2023exploringlimitstransferlearning} (2019) treats all text-based tasks as text-to-text problems, offering significant flexibility across a range of natural language processing applications.
\texttt{BART} \cite{lewis-etal-2020-bart} (2019) is particularly effective at transforming textual input into structured outputs, making it suitable for complex textual tasks.
\texttt{UnixCoder} \cite{guo2022unixcoder} (2022) is an architecture specialized in programming-related tasks, such as code generation, code summarization, and code translation.}
\new{Additionally, \texttt{LLaMa 2} \cite{touvron2023llama2} (2023) and its successor, \texttt{LLaMa 3} \cite{dubey2024llama3herdmodels} (2024) provide \glspl{llm} optimized for instruction-following (e.g., \texttt{LLaMa 3-70B-Instruct}) and are fine-tuned for tasks requiring structured text understanding.}
A total of \new{9} studies rely on these architectures: 
\new{3 on \texttt{T5} (2019), 3 on \texttt{BART} (2019), 1 on \texttt{UnixCoder} (2022), 1 on \texttt{LLaMa 2} (2023), and 1 on \texttt{LLaMa 3} (2024).}

Additionally, \numarchmultimodal architectures focus on multimodal data. 
\new{All these architectures (or their multimodal models) were introduced in 2024. \texttt{Gemini} emphasizes both multimodal capabilities and generative approaches, making it ideal for tasks that require integrating and generating text, images, and possibly other forms of data. 
\texttt{GPT-4-Vision-Preview} extends the \texttt{GPT-4} \cite{openai2024gpt4} architecture by incorporating visual input processing, allowing it to handle tasks involving image understanding and text-image reasoning. 
\texttt{Mixtral} \cite{jiang2024mixtral} leverages an ensemble of models designed to handle multimodal tasks, such as text-to-image and image-to-text conversions. 
\texttt{OpenCoder} \cite{huang2024opencoder} is optimized for code-related tasks and supports multimodal inputs, particularly focusing on enhancing software development workflows with a combination of textual and structural data processing. 
A total of 9 studies rely on these architectures: 5 on \texttt{Gemini}, 2 on \texttt{Mixtral}, 1 on \texttt{GPT-4}, and 1 on \texttt{OpenCoder}.}

\new{
At the time of writing, several \glspl{llm} analyzed in this study have been deprecated or replaced by newer versions. OpenAI’s \texttt{Text-Davinci-003} and \texttt{Text-Davinci-Edit-001}, based on \texttt{GPT-3.5}, have been phased out in favor of \texttt{GPT-4-Turbo} and \texttt{GPT-4o}. Google rebranded \texttt{Bard} as \texttt{Gemini}, transitioning from \texttt{PaLM 2} to the \texttt{Gemini 1.x} and \texttt{2.x} series. Meta’s \texttt{LLaMa 2-13B} has been largely replaced by \texttt{LLaMa 3}, while OpenAI’s \texttt{GPT-4} evolved into \texttt{GPT-4o}, incorporating multimodal capabilities. Similarly, \texttt{Mixtral-8x7B-Instruct-v0.1}, \texttt{Qwen}, and \texttt{DeepSeek} have advanced to \texttt{Mixtral}, \texttt{Qwen2}, and \texttt{DeepSeek-V2}, respectively. Anthropic’s \texttt{Claude 3.5} is also expected to be succeeded by newer iterations. Given the rapid evolution of \glspl{llm}, readers should consider the latest available models when interpreting our findings.
}

\new{
Given the varying performance of \glspl{llm} across architectures, access to their source code remains a key issue to be analyzed. Most high-performing models, including OpenAI’s \texttt{GPT-3.5} and \texttt{GPT-4}, Google’s \texttt{PaLM 2} and \texttt{Gemini}, as well as Anthropic’s \texttt{Claude 3.5} and Cohere’s \texttt{Command-R+}, are closed source and require paid access. \texttt{Mixtral}, while open-weight, is commercialized via paid APIs, and \texttt{DeepSeek V2} adopts a more restrictive licensing model.
In contrast, Meta’s \texttt{LLaMa 2} and \texttt{LLaMa 3} are open source under specific licenses, while Mistral AI offers both open-weight and commercial models. \texttt{Qwen} and \texttt{Qwen2} provide open-source variants alongside proprietary versions. Fully open-source models include \texttt{BART}, \texttt{T5}, and certain versions of \texttt{LLaMa}, \texttt{Mistral}, \texttt{StarCoder}, and \texttt{OpenCoder} (the latter two specialized for code generation). Given the evolving nature of access policies and licensing conditions, these constraints may influence model selection in research.
}

We count separately \new{both} specific implementations designed for conversational purposes and versions of the same model updated to a specific timestamp, such as \texttt{GPT-4-0613} \new{(referring to a version of \texttt{GPT-4} released or fine-tuned on June 13, 2023).} This approach aims to enhance clarity and improve reproducibility. However, there is no guarantee of consistent responses even when \new{using} different versions of the same models \cite{Kochanek2024-cw, FREIRE2024659e1}, especially \new{since} chat completions are not deterministic by default \cite{openai_reproducible_outputs}.

When using \glspl{llm} in \new{a research setting}, several key aspects must be addressed to ensure \new{both} reliable and reproducible results. Documenting the exact prompts used is crucial, as \glspl{llm} are highly sensitive to prompt variations, which can lead to significantly different outcomes \cite{anagnostidis2024susceptible, errica2024did}. Additionally, detailed reporting on the model version, configuration, and any fine-tuning is essential for accuracy and replicability. Notably, a limited number of studies \new{(7)} that conducted experiments in this area did not provide any details about the \gls{llm} employed \new{\cite{LIU2023100520, teukam2024integrating, michailidis_et_al:LIPIcs.CP.2024.20, regin_et_al:LIPIcs.CP.2024.25, 10803039, 10720437, 10711695}}.
Understanding the training data used in an \gls{llm} can help identify potential data leakage, biases, or knowledge gaps that might \new{affect} results \cite{balloccu2024leak}. Moreover, documenting any additional steps, such as data preprocessing or post-processing of model outputs, is critical for ensuring that findings can be replicated across different studies and datasets. By addressing these factors, research involving \glspl{llm} remains transparent, reproducible, and robust.

\new{Finally, when evaluating \glspl{llm} in the context of \glspl{cop}, no universally shared metrics emerge. Accuracy is the most commonly used metric, but its interpretation varies by task. Researchers sometimes define a problem-specific \lq\lq score\rq\rq{}, while other metrics remain domain-specific. } 
\new{Indeed, many studies evaluate \glspl{llm} based on the objective values of the solutions they produce. As discussed in \Cref{sec:combinatorial-optimization}, this evaluation is problem-specific, as each problem has distinct objective functions. For instance, in \gls{tsp}, the goal is to minimize travel distance, whereas in \gls{pfsp}, it is typically to minimize tardiness or completion time. Since the optimal solution may not always be available, evaluations often compare the obtained solutions to the best-known ones. A common metric for this comparison is the optimality gap (or relative deviation), calculated as:  
$\text{gap} = 100 \times (Z_{\text{llm}} - Z_{\text{best}})/ Z_{\text{best}}$
where $Z_{\text{llm}}$ represents the objective value of the \gls{llm} solution, and $Z_{\text{best}}$ denotes the best-known objective value.}

\subsubsection{Problem Modeling}
\label{sec:large-language-models-analysis-subsec:problem-modeling}

A total of \numllmproblemmodeling \glspl{llm} out of \numllm\xspace \new{(57\%)} have been used for problem modeling. Among them, \numllmproblemmodelingmodelcreation \glspl{llm} have focused on enhancing \emph{model creation}. A common approach involves creating models from unstructured \gls{nl}. Some of these approaches address optimization problems more broadly, utilizing \glspl{llm} such as \texttt{GPT-3.5-Turbo} (an optimized variant of \texttt{GPT-3.5} for faster
performance and lower cost~\cite{xiao2024chainofexperts}), \texttt{GPT-3.5-turbo-0613} (a June 2023 release of \texttt{GPT-3.5-Turbo}), \texttt{GPT-4-0613} (a June 2023 release of \texttt{GPT-4}), and
\texttt{GPT-4} itself~\cite{10738100}, as well as \texttt{LLaMa 2-7b}, a version of \texttt{LLaMa 2} with 7 billion parameters~\cite{ahmed2024lm4opt}, and \texttt{T5-Base}, the base model built on the \texttt{T5} architecture~\cite{he2022linear}. \new{Additional optimized or specialized  variants, including \texttt{GPT-4o}, an improved multimodal version of \texttt{GPT-4} \cite{ahmaditeshnizi2024optimus03usinglargelanguage}, \texttt{Qwen1.5-14B}, \texttt{DeepSeek-V2} \cite{yang2024optibenchmeetsresocraticmeasure}, \texttt{Mistral-7B}, \texttt{DeepSeek-Math-7B}, \texttt{LLaMa 3-8B}, and \texttt{Qwen2.5-7B}~\cite{huang2025orlmcustomizableframeworktraining}, have also been used for broad optimization tasks, leveraging improved reasoning capabilities and efficiency. Moreover, instruction-tuned models such as \texttt{Code Llama-Instruct} and \texttt{Zephyr-7b-beta}, a \texttt{7B}-parameter model designed for instruction-following, have been used for structured task execution~\cite{Mostajabdaveh04112024}. Furthermore, both \texttt{GPT-4o} and \texttt{Claude 3.5 Sonnet} have been employed in a zero-shot fashion to generate problem formulations directly from user input, 
thus reducing the need for extensive prompts or examples~\cite{hao2025planningrigorgeneralpurposezeroshot}.
Other approaches focus on specific formulations. For instance, \glspl{llm} have been used to automate the generation of \gls{lp} models with \texttt{ChatGPT-3.5}~\cite{li2023synthesizing}, \texttt{GPT-4}~\cite{abdullin-etal-2023-synthetic}, and the \texttt{BART} architecture, including both \texttt{BART-Base}~\cite{gangwar2023highlighting,ramamonjison-etal-2022-augmenting} and \texttt{BART-Large} variants~\cite{gangwar2023highlighting,jang2022tag}. Furthermore, \texttt{ChatGPT-3.5} has also been employed in \gls{milp} problems \cite{amarasinghe2023aicopilot}\new{, as have \texttt{Bard}, a conversational model built on the \texttt{PaLM-2} architecture~\cite{li2023synthesizing}, and \texttt{LLaMa 3-70B}~\cite{ju-etal-2024-globe}. Instruction-tuned \texttt{Code Llama-Instruct} and \texttt{Zephyr-7b-beta} have likewise been applied to \gls{qp} tasks by translating user directives
into valid constraints and objective functions.}}

\new{Concerning \emph{entity recognition}, \numllmproblemmodelingentityrecognition~\glspl{llm} have been used to identify and extract specific elements from model representations. Specifically, \texttt{GPT-4} and \texttt{Text-Davinci-003}, based on \texttt{GPT-3.5} and optimized for a wide range of tasks, have been used to translate user input into constraints that the underlying \gls{cp} model can process \cite{lawless2024i}. This approach has also been applied to \gls{lp} and \gls{milp} problems using \texttt{GPT-3.5} \cite{ahmaditeshnizi2024optimus, zhang-etal-2024-solving}, \texttt{GPT-4}~\cite{zhang-etal-2024-solving}, \texttt{ChatGPT-4} \cite{ALIPOURVAEZI2024110574}, \texttt{LLaMa 3-70B}~\cite{ju-etal-2024-globe,Mostajabdaveh04112024,ahmaditeshnizi2024optimus03usinglargelanguage}, \texttt{ChatGPT-4o} \cite{hao2025planningrigorgeneralpurposezeroshot,jiang2025llmopt,ahmaditeshnizi2024optimus03usinglargelanguage}, and \texttt{Claude 3.5 Sonnet}~\cite{hao2025planningrigorgeneralpurposezeroshot}. Meanwhile, \texttt{CodeLlama-Instruct} and \texttt{Zephyr-7b-beta} have also been used for \gls{qp} problems. Additionally, \texttt{ChatGPT-3.5} and \texttt{Bard} have been used exclusively for \gls{milp} 
\cite{li2023synthesizing}, while \texttt{T5-Base} has been employed for \gls{lp} \cite{he2022linear}. Furthermore, several \glspl{llm} have been applied to general optimization problems, including 
\texttt{GPT-3.5-Turbo-0163}, \texttt{GPT-4}, \texttt{LLaMa 2-7b}, \texttt{LLaMa 2-13b}, and \texttt{LLaMa 3-70B} \cite{ahmed2024lm4opt, Da2024}, as well as \texttt{GPT-4o-mini} \cite{li2025graphteamfacilitatinglargelanguage}, \texttt{BART-Base}, \texttt{BART-Large} 
\cite{ramamonjison-etal-2022-augmenting, jang2022tag}, and \texttt{Gemini 1.0 Pro}~\cite{huang2024words}.}

\new{Additionally, there are \numllmproblemmodelingdomainknowledge applications of \glspl{llm} for extracting \emph{domain-specific knowledge}. For instance, \texttt{GPT-4} and \texttt{GPT-4-0163} have been used to extract general domain knowledge in household financial planning \cite{ai5010006, ju-etal-2024-globe}. \texttt{GPT-4o}, \texttt{Claude 3.5 Opus}, \texttt{Command-R+}, and \texttt{Mixtral-8x2B} have been applied to a social network problem \cite{10818476}, and \texttt{GPT-4o} has also been used in robotics \cite{10675146}. \texttt{LLaMa2-7B}, \texttt{LLaMa2-13B}, \texttt{GPT-3.5}, and \texttt{GPT-4} have been employed to model knowledge for the \gls{vrp} problem \cite{Da2024}. Moreover, a single \gls{llm}, \texttt{ChatGPT-4}, has been used to model domain knowledge in the form of knowledge graphs
\cite{smartcities7050094}.}

\subsubsection{Solution Methods}
\label{sec:large-language-models-analysis-subsec:solution-methods}

\glspl{llm} has been employed in \numllmsolutionmethod out of \numllm\new{\xspace(85\%)} works for what concerns the solution method task. Specifically, \numllmsolutionmethodsolutiongeneration \glspl{llm} have been used for \emph{solution generation}. One approach involves creating candidate solutions for well-known \glspl{cop}, such as a specific class of the \gls{vrp}, by leveraging both GPT-based models like \texttt{GPT-4} \cite{ahmaditeshnizi2024optimus}, \new{\texttt{GPT-4-Turbo}, and \texttt{GPT-4o} \cite{Khan_2024}}, and LLaMa-based or T5-based models, including \texttt{LLaMa 2} and \texttt{T5-Base} \cite{chin2024learning}. Another approach uses \glspl{llm} to combine problem descriptions and previously generated solutions within a meta-prompt processed by \texttt{ChatGPT-3.5-Turbo}, \texttt{PaLM 2-L}, \texttt{PaLM 2-L-IT}, and \texttt{text-bison} \cite{yang2024large}. The multimodal capabilities of \texttt{GPT-4-Vision-Preview} and \new{\texttt{ChatGPT-4o}} have also been employed to generate solutions through visual prompts \cite{huang2024multimodal, make6030093}. Moreover, \texttt{ChatGPT-4} has been adopted for finance-related solutions~\cite{RePEc:spr:snopef:v:4:y:2023:i:4:d:10.1007_s43069-023-00277-6} or automated sequence planning in robot-based assembly \cite{buildings13071772}. 
\new{Other \glspl{llm} used for domain-specific problems include \texttt{GPT-4} for travel planning~\cite{delarosa2024trippaltravelplanningguarantees}, program scheduling~\cite{10.1145/3664646.3665084}, and traffic simulation \cite{Da2024}. \texttt{LLaMa 2-7b} and \texttt{LLaMa 2-13b} have also been applied to traffic simulation. \texttt{Claude 3.5 Sonnet} has been used in molecular biology \cite{reinhart-2024}, and \texttt{ChatGPT-4-Turbo} together with \texttt{ChatGPT-4o-mini} has been used to coordinate computation in a graph reasoning scenario~\cite{hu2024scalableaccurategraphreasoning}. \texttt{GPT-3.5-Turbo} has found utility in industry-related problems~\cite{xiao2024chainofexperts}, while \texttt{GPT-4-Turbo}, \texttt{GPT-4o}, \texttt{Gemini 1.5} (Pro and Flash), and \texttt{Gemma 2 27B} have been employed in planning tasks~\cite{bohnet2024exploringbenchmarkingplanningcapabilities, 10675146}.} Furthermore, in the context of \glspl{ga}, \texttt{GPT-3.5-Turbo-0613} has been used to select parent solutions from the current population and perform crossover and mutation~\cite{liu2024large1}, as well as to perform mutation and crossover only~\cite{mao2024identify}.

Additionally, \numllmsolutionmethodcodegeneration \glspl{llm} leverage approaches for \emph{code generation}. \texttt{GPT-3.5}, \texttt{GPT-4}, \texttt{GPT-4o}, and \texttt{Qwen (LoRA Fine-Tuned)}, which is a version of the \texttt{Qwen} family fine-tuned using LoRA~\cite{hu2021lora}, have been used to generate code specifically for \gls{lp}, \gls{mip}, and \gls{milp} approaches to \glspl{cop}~\cite{ahmaditeshnizi2024optimus, zhang-etal-2024-solving, 10738100, li2024foundationmodelsmixedinteger}, whereas \texttt{CodeLlama-Instruct} and \texttt{Zephyr-7b-beta} have been applied to both \gls{milp} and \gls{qp} \cite{Mostajabdaveh04112024}. Various methods for automating the generation of heuristic algorithms rely on a range of \glspl{llm}, each optimized for different aspects of code generation. For example, \texttt{CodeLlama}, a code-oriented variant of \texttt{LLaMa 2}~\cite{10.1007/978-3-031-70068-2_12}, and \texttt{StarCoder}, trained on extensive code-related datasets, are fine-tuned for specific programming tasks, while \texttt{DeepSeek-LLM-7B-Base}, \texttt{GPT-3.5-Turbo}, and \texttt{GPT-4-Turbo} are optimized for speed and efficiency \cite{liu2024evolution, ye2024large, liu2023algorithm, liu2024large, Romera-Paredes2024, ju-etal-2024-globe, liu2024llm4adplatformalgorithmdesign, yu2024autornetautomaticallyoptimizingheuristics, yao2024multiobjectiveevolutionheuristicusing}. \texttt{GPT-3.5-Turbo-0613} has likewise been employed for generating crossover and mutation implementations in Python \cite{mao2024identify} within \gls{ea} contexts. \new{Multimodal models, such as \texttt{GPT-4o}, and advanced reasoning models like \texttt{Claude 3.5 Opus} and \texttt{Claude 3.5 Haiku}, have also been utilized in heuristic generation tasks, and \texttt{DeepSeek-Coder} along with its updated \texttt{DeepSeek-Coder-V2} focus on high-performance code synthesis. Similarly, \texttt{GLM-3-Turbo}, \texttt{OpenCoder-8B-Instruct}, and \texttt{Yi-34b-Chat} provide structured, optimized code generation \cite{yatong2024tseohedgeservertask, liu2024llm4adplatformalgorithmdesign}. For large-scale problem-solving, models such as \texttt{Gemini 1.0 Pro} and \texttt{Codey}, both built on \texttt{PaLM 2}, have shown strong performance in complex coding scenarios, while the latest iterations of foundation models (e.g., \texttt{LLaMa 3-70B}, \texttt{LLaMa3-70B-Instruct}, \texttt{LLaMa 3.1-8B}, \texttt{GPT-3.5-Turbo}, \texttt{GPT-4o-Mini}, and \texttt{Qwen-Turbo}) have been refined for specialized programming applications \cite{10.1007/978-3-031-70068-2_12, yatong2024tseohedgeservertask, liu2024llm4adplatformalgorithmdesign}. Other approaches harness a self-reflection mechanism to directly generate executable Python code from natural language, exemplified by \texttt{GPT-4} and \texttt{Gemini 1.0 Pro}~\cite{huang2024words}. A comprehensive code-generating framework for business optimization has also been developed using \texttt{CodeT5-finetuned\_CodeRL} \cite{le2022coderlmasteringcodegeneration}, a model that employs reinforcement learning on top of \texttt{CodeT5} \cite{amarasinghe2023aicopilot}. Additionally, \texttt{Text-Davinci-Edit-001} has been employed to generate MiniZinc-specific representations~\cite{almonacid2023automatic}, while \texttt{Text-Davinci-003} and \texttt{GPT-4} have produced Gurobi-based code for supply chain optimization \cite{li2023large}. \new{\texttt{GPT-3.5-Turbo}, \texttt{Mistral-7B}, \texttt{DeepSeek-Math-7B}, \texttt{LLaMA 3-8B}, and \texttt{Qwen2.5-7B} have been broadly used for industry-related problems \cite{xiao2024chainofexperts, huang2025orlmcustomizableframeworktraining}, and a specialized version called \texttt{GPT-4o-2024-08-06} has been introduced to optimize code for SAT solvers \cite{sun2024autosatautomaticallyoptimizesat}. Moreover, \texttt{GPT-4o} and \texttt{Claude 3.5 Sonnet} have addressed supply chain and robot logistics tasks \cite{hao2025planningrigorgeneralpurposezeroshot}, and \texttt{ChatGPT-4o-mini} has been applied to multi-agent solution design \cite{li2025graphteamfacilitatinglargelanguage}.}}

\new{Moving to \emph{parameter tuning}, we identified \numllmsolutionmethodparametertuning applications of \glspl{llm}. \texttt{ChatGPT-4o} has been adopted to model user preferences by adjusting an optimizer's weights \cite{a17120582}, whereas \texttt{GPT-3.5}, \texttt{GPT-4}, \texttt{Gemini}, and \texttt{Le Chat} (the chatbot interface for Mistral models) have been used to set \glspl{mh} parameters~\cite{DBLP:conf/esann/MartinekLG24}.}

\new{A single application aimed at enhancing \emph{algorithm selection}: \texttt{UnixCoder}, a code representation model designed for programming tasks, has been used to extract features linked to the underlying optimization algorithms~\cite{10.24963/ijcai.2024/579}.}

\subsubsection{Validation}
\label{sec:large-language-models-analysis-subsec:val}

\new{
We found \numllmtvalidation applications of \glspl{llm} out of \numllm (14\%), in the context of validation for optimization models. In particular, \numllmtvalidationsolutionvalidation \glspl{llm} have been involved in \emph{solution validation}. \texttt{GPT-4} and \texttt{Gemini 1.0 Pro} have been used to validate solutions generated for the \gls{vrp} \cite{huang2024words}, while \texttt{GPT-3.5}, \texttt{LLaMa 2-7b}, \texttt{LLaMa 2-13b}, and \texttt{ChatGPT-4o} have been applied to broader validation tasks \cite{Da2024, make6030093}. Additionally, \texttt{Qwen (LoRA Fine-Tuned)} has been leveraged for solution validation~\cite{zhang-etal-2024-solving}. 

As for \emph{model validation}, \numllmtvalidationmodelvalidation \glspl{llm} have played a role. \texttt{GPT-4} has been employed to diagnose infeasible \gls{milp} models through interactive sessions \cite{chen2023diagnosing}, while \texttt{GPT-3.5-Turbo} has been used to validate models built from natural language \cite{xiao2024chainofexperts}. Similarly, \texttt{GPT-3.5}, \texttt{GPT-4}, and \texttt{Qwen (LoRA Fine-Tuned)} have also been applied in this context \cite{zhang-etal-2024-solving, chen2023diagnosing}, with \texttt{GPT-4o} and \texttt{Claude 3.5 Sonnet} reported for model validation tasks~\cite{hao2025planningrigorgeneralpurposezeroshot}. In addition, code-focused models have been utilized to ensure correctness: \texttt{CodeLlama-Instruct} and \texttt{Zephyr-7b-beta} have been adopted to verify that the generated code remains consistent with the original model formulations \cite{Mostajabdaveh04112024}.
}

\subsubsection{Benchmarking}
\label{sec:large-language-models-analysis-subsec:bench}

As for benchmarking, \numllmtbenchmarking out of \numllm\new{\xspace(14\%)} \glspl{llm} have been used to enhance it. Specifically, \numllmtbenchmarkingvisualanalysis \glspl{llm} \new{use \emph{visual analysis}, as they are integrated, for instance, with a tool designed to visualize the behavior of various algorithms} applied to specific instances of a \gls{cop} \cite{sartori2024large}. The models used by \citet{sartori2024large} include \texttt{Mixtral-8x7B-Instruct-v0.1}, which is optimized for instructional and guided tasks, \texttt{GPT-4-Turbo}, and \texttt{Tulu-v2-dpo-7b}, a fine-tuned version of \texttt{LLaMa 2} that was trained on a mix of publicly available, synthetic, and human-generated datasets using a parametrization of the RLHF algorithm known as Direct Preference Optimization \cite{rafailov2024directpreferenceoptimizationlanguage}. \new{Furthermore, \texttt{GPT-3.5}, \texttt{GPT-4}, \texttt{LLaMa 2-7b}, and \texttt{LLaMa 2-13b} have been employed to visualize solutions for a domain-specific problem \cite{Da2024}, while \texttt{ChatGPT-4o} has been used to interpret visual inputs in lieu of purely numerical data \cite{make6030093}.}

We also \new{identified \numllmtbenchmarkingexplainability applications} aimed at improving \emph{explainability}. \texttt{ChatGPT-4} has been employed to describe the decision-making process for the \gls{vrp} \cite{10.1007/978-981-97-2259-4_3}, \new{while \texttt{ChatGPT-4-Turbo}, \texttt{ChatGPT-4o}, \texttt{ChatGPT-4o-mini}, and \texttt{Claude 3.5 Sonnet} have been used for instances of both the \gls{tsp} and m\gls{tsp} \cite{Da2024, reinhart-2024, hu2024scalableaccurategraphreasoning, a17120582}}.

\subsubsection{Platforms for Supporting LLMs-Based Approaches}
\label{sec:large-language-models-analysis-subsec:bonus}

While reviewing the studies considered for inclusion, we \new{came across} two platforms used by \citet{sartori2024large} to support the design of their approach. Although these platforms are general-purpose and facilitate the use of \glspl{llm} \new{broadly rather than specifically} in the context of \gls{co}, we believe it is \new{beneficial} to describe them briefly. 

Chatbot Arena \cite{chiang2024chatbotarenaopenplatform} is an open platform for evaluating \glspl{llm} based on human preferences. It offers access to over twenty \glspl{llm}, including both proprietary and open-source options, and provides a leaderboard to compare results. Chat2Vis \cite{10121440}, on the other hand, focuses on generating visualizations directly from natural language text. It uses various \glspl{llm} and shows that a set of proposed prompts offers a reliable approach to rendering visualizations from natural language queries, even when they are highly misspecified or underspecified.

\subsection{Benchmark Datasets}
\label{sec:datasets}

\new{The methodologies proposed in the studies have been evaluated using well-known \gls{co} instances, related problem suits (e.g., MIPLIB\footnote{\url{https://miplib.zib.de/}} and ASLIB),\footnote{\url{https://www.coseal.net/aslib/}} or datasets specifically developed for the case of \glspl{llm}.} 
\new{Considering this latter point, 29} studies assess the efficacy and efficiency of leveraging \gls{llm} within \gls{co} \new{with} specific benchmark datasets developed for this purpose. \new{We now describe these datasets and report a tabular overview in \Cref{app:benchmark-datasets-analysis}}.

The most frequently used benchmark dataset is the LPWP, also referred to as the NL4Opt dataset, which has been employed in \new{16} studies \new{\cite{ahmed2024lm4opt,li2023synthesizing,ahmaditeshnizi2024optimus,xiao2024chainofexperts,wang2023opdnl4opt,doan2022vtccnlp,ning2023novel,gangwar2023highlighting,he2022linear,abdullin-etal-2023-synthetic,jang2022tag,ramamonjison-etal-2022-augmenting,michailidis_et_al:LIPIcs.CP.2024.20,jiang2025llmopt,zhang-etal-2024-solving,huang2025orlmcustomizableframeworktraining}}. Initially introduced by \citet{ramamonjison-etal-2022-augmenting} and later expanded by \citet{li2023synthesizing}, this dataset has been used in the \gls{nl4opt} competition. The original dataset includes 4,216 \gls{nl} problem declarations derived from 1,101 \gls{lp} problems across six different domains. The extension enhances the dataset by introducing new problem descriptions and constraint types, such as logic constraints and binary variables. 
The second most used benchmark dataset is the ComplexOR dataset \cite{xiao2024chainofexperts}, which has been employed in \new{3} studies \new{\cite{xiao2024chainofexperts,ahmaditeshnizi2024optimus,jiang2025llmopt}} \new{and has been introduced to complement the NL4Opt dataset}. The dataset consists of \gls{nl} descriptions of 37 problems across different domains, sourced from academic studies and real-world scenarios, encompassing 25 \gls{lp} formulations and  12 \gls{milp} formulations. \new{As ComplexOR also the NLP4LP dataset \cite{ahmaditeshnizi2024optimus,ahmaditeshnizi2024optimus03usinglargelanguage} has been used in 3 studies \cite{ahmaditeshnizi2024optimus,jiang2025llmopt,ahmaditeshnizi2024optimus03usinglargelanguage}}. It encompasses \gls{nl} descriptions of 67 problems across various domains and including both \gls{lp} (54) and \gls{milp} (13) formulations. \new{This same dataset has been enriched to up to more than 200 optimization problems by \citet{ahmaditeshnizi2024optimus03usinglargelanguage}}.
\new{\citet{huang2025orlmcustomizableframeworktraining} proposed OR-Instruct, a pipeline for creating synthetic data for optimization data. To test the capabilities of such a pipeline, they created IndustryOR, which has been used in 2 studies \cite{huang2025orlmcustomizableframeworktraining,jiang2025llmopt}. As the name suggests, it focuses on industrial problems (a total of 100 real-world problems from eight industries) covering \gls{lp}, \gls{ilp}, \gls{milp}, and non-linear programming.}
\new{\citet{huang2025llmsmathematicalmodelingbridging} introduced Mamo, a dataset for \gls{lp} modeling. It includes 652 easy and 211 complex \gls{lp} problems, each paired with its corresponding optimal solution, sourced from various academic materials. Mamo has been used in 2 studies \cite{jiang2025llmopt,huang2025orlmcustomizableframeworktraining}.}
\new{\citet{luo2024graphinstructempoweringlargelanguage} introduced GraphInstruct, a dataset comprising 21 reasoning problems on the topic of graphs and networks, including \glspl{cop}. GraphInstruct has been used in 2 studies \cite{hu2024scalableaccurategraphreasoning,li2025graphteamfacilitatinglargelanguage} . Similar to this dataset, there is Talk Like A Graph \cite{fatemi2023talklikegraphencoding}, LLM4DyG \cite{zhang2024llm4dyglargelanguagemodels}, GraphViz \cite{chen2024graphwizinstructionfollowinglanguagemodel}, NLGraph \cite{wang2024languagemodelssolvegraph}, and GNN-AutoGL \cite{li2025graphteamfacilitatinglargelanguage} (note that these datasets have been used only by one study \cite{li2025graphteamfacilitatinglargelanguage}). }

The remaining datasets have been used exclusively in \new{one out of the retrieved studies -- many times this being the study proposing the dataset on the first place.}
\citet{amarasinghe2023aicopilot} introduced the AI-copilot-data dataset, which includes 100 \gls{nl} descriptions of problems within the production scheduling domain.
\citet{huang2024words} introduced the homonym dataset,\footnote{\new{We use the term ``homonym'' when the authors did not provide a specific name for the dataset.}} which comprises 80 \gls{nl} descriptions of routing problems, including variants for single-robot and multi-robot routing.
\citet{almonacid2023automatic} introduced a homonym dataset, which comprises 10 \gls{nl} instructions for the creation of MiniZinc models involving discrete variables and arrays of discrete variables, thus \gls{ilp} problems.
\citet{chen2023diagnosing} introduced the OptiChat dataset, which comprises 63 \new{infeasible (i.e., inconsistent)} \gls{milp} model formulations across various domains. This dataset was derived from feasible models expressed through the Python library Pyomo \cite{Hart2011} and sourced from a collection of libraries and textbooks. The formulations were created by modifying one or more model parameters (e.g., minimum inventory, demand, maximum capacity) or adding constraints (e.g., maximum cost, minimum demand for a particular product) \new{so to make the instances infeasible (i.e., their set of \new{feasible} solutions is empty)}.
\citet{lawless2024i} introduced two datasets, Safeguard and Code Generation. They both are strictly connected to the framework the authors proposed. 
The Safeguard dataset is a binary classification dataset designed to evaluate whether the system \new{at hand (i.e., a system that integrates \gls{cp} and \glspl{llm} to schedule meetings in a company)} has sufficient data sources \new{(e.g., information related to the meeting attendees)} to handle given \new{\gls{nl}} constraints. 
The Code Generation dataset contains a collection of \new{\gls{nl}} constraints that can be translated into Python code using the data structures available in the system. \new{An example of such constraints is “The team has a no-meeting policy on \{WEEKDAY\}”.} This dataset aims to test the capability of generating executable Python code that satisfies the specified constraints.
\new{While employing NL4Opt and other existing dataset from the \gls{ml} community, \citet{michailidis_et_al:LIPIcs.CP.2024.20} also introduced a homonym \gls{cp}-based benchmark. The dataset was built using 18 \gls{cp} problems from a university-level \gls{cp} modelling course.}
\new{\citet{yang2024optibenchmeetsresocraticmeasure} introduced OptiBench and ReSocratic-29k. OptiBench includes 816 real-world optimization problems spanning multiple domains, focusing on linear and mixed-integer programming and introducing a graph-based evaluation method to assess model correctness. ReSocratic-29k consists of 29,000 optimization problem demonstrations, generated through a reverse synthesis approach that first constructs step-by-step formulations before back-translating them into \gls{nl} questions. These datasets provide a targeted benchmark for assessing and improving LLMs in formulating and solving optimization tasks.}
\new{\citet{borazjanizadeh2024navigatinglabyrinthevaluatingenhancing} introduced SearchBench, a dataset encompassing five problem categories and 1,107 instances, primarily focused on puzzles and general combinatorics. Each problem type corresponds to a well-known \gls{cop}, but the constraints have been slightly modified to reduce the likelihood that \glspl{llm} encountered identical problems during their training phase. }
\new{\citet{Mostajabdaveh04112024} presented a new dataset to complement the existing NL4Opt and ComplexOR benchmarks, aiming to provide less structured input and more complex optimization scenarios. This benchmark includes problems related to \gls{lp}, \gls{milp}, and \gls{qp}. Unlike NL4Opt, which expects a formal model as output, and ComplexOR, which requires Python code, this dataset outputs solutions in Zimple code.}
\new{Similarly, \citet{zhang-etal-2024-solving} enriched the NL4Opt dataset with English and Chinese problem descriptions. }
\new{\citet{ju-etal-2024-globe} introduced a dataset of 173,700 training and 21,800 test samples related to travel planning.}

\new{Finally, a different dataset is ORQUA \cite{mostajabdaveh2024evaluatingllmreasoningoperations}. It is designed to evaluate the extent of \gls{co} knowledge in \glspl{llm}. Given a problem description, the dataset assesses the model's ability to understand and identify, for example, the appropriate type of mathematical modeling the problem corresponds to.}

\begin{table}[ht]
    \centering
    \caption{\new{Examples of mathematical reasoning tasks from various benchmark datasets.}}
    \label{tab:dataset-examples}
    \resizebox{1\textwidth}{!}{%
    \begin{tabular}{p{2.5cm} p{2.9cm} p{10.3cm}}
    \toprule
    \footnotesize \textbf{\new{Dataset}} & \footnotesize \textbf{\new{Task}} & \footnotesize \textbf{\new{Example}} \\
    \midrule
    \footnotesize \new{GSM8K} \new{\cite{cobbe2021trainingverifierssolvemath}} & \footnotesize \new{Arithmetic Word Problem} & \footnotesize \emph{\new{Natalia sold clips to 48 of her friends in April, and then she sold half as many clips in May. How many clips did Natalia sell altogether in April and May?}} \\
    
    \footnotesize \new{MultiArith} \new{\cite{hosseini-etal-2014-learning}} & \footnotesize \new{Multi-Step Arithmetic} & \footnotesize \emph{\new{John has 3 apples. He buys 5 more apples and then eats 2. How many apples does he have left?}} \\
    
    \footnotesize \new{AquA} \new{\cite{ling-etal-2017-program}} & \footnotesize \new{Probability} & \footnotesize \emph{\new{From a pack of 52 cards, two cards are drawn together at random. What is the probability of both the cards being kings?}} \\
    
    \footnotesize \new{BIG-Bench} \new{\cite{srivastava2023beyond}} & \footnotesize \new{Logical Reasoning} & \footnotesize \emph{\new{If a snail climbs a 10-meter pole at a rate of 2 meters per day but slides back 1 meter each night, how many days will it take to reach the top?}} \\
    
    \footnotesize \new{BIG-Bench Hard}~\new{\cite{suzgun2022challengingbigbenchtaskschainofthought}} & \footnotesize \new{Combinatorial Reasoning} & \footnotesize \emph{\new{A school has 6 different clubs. Each student must join exactly 2 clubs. How many unique pairs of clubs can be formed?}} \\
    
    \bottomrule
    \end{tabular}}
\end{table}

\new{Similar to the platforms discussed in \Cref{sec:large-language-models-analysis-subsec:bonus}, we also identified general-purpose datasets in the reviewed studies. These datasets are valuable for developing \glspl{llm}-based approaches to optimization, particularly by providing math-related problems expressed in unstructured natural language which are useful especially for problem modeling (e.g., handling addition, multiplication, and data structures like sets). These resources can support researchers and practitioners in designing new applications of \glspl{llm} in \gls{co}. We briefly outline their characteristics and report some examples in \Cref{tab:dataset-examples}. Notably, all of these resources were used by \citet{ahmed2024lm4opt}, while GSM8K was also employed by \citet{yang2024large}. }

\new{
GSM8K \cite{cobbe2021trainingverifierssolvemath} consists of 8,500  linguistically diverse grade-school human-generated math word problems. Unlike simpler arithmetic datasets like AddSub \cite{hosseini-etal-2014-learning} (395 problems) and SingleOp \cite{koncel-kedziorski-etal-2015-parsing} (562 problems), GSM8K requires multi-step reasoning. Its emphasis on stepwise numerical reasoning aligns with \gls{co}, where problems often require sequential computations and recursive decision-making.

MultiArith \cite{hosseini-etal-2014-learning} consists of around 600 multi-step arithmetic problems that require sequential operations (e.g., addition, subtraction, and multiplication). It was partially generated from existing math problems and structured for \gls{ml} applications. Its focus on structured numerical reasoning makes it relevant to \gls{co} techniques such as branch-and-bound and constraint satisfaction, which rely on constructive decision-making.

AquA \cite{ling-etal-2017-program} presents 100,000 human- and machine-generated algebraic word problems in a multiple-choice format along with rationales explaining each answer. It requires the computation of correct answers and their selection from distractors. Many \glspl{cop} involve symbolic reasoning and equation solving, making AquA useful for assessing a model’s ability to handle algebraic structures and optimization constraints.

BIG-Bench \cite{srivastava2023beyond} is a large-scale benchmark evaluating language models across more than 200 tasks, covering mathematics, reasoning, linguistics, commonsense understanding, and code generation. The dataset was created through a combination of human-designed tasks and algorithmically generated problem sets, allowing for a broad evaluation of logical reasoning and heuristic problem-solving. This makes it valuable for studying how models approach optimization-based decision-making.
A more challenging subset, BIG-Bench Hard \cite{suzgun2022challengingbigbenchtaskschainofthought}, focuses on 23 particularly difficult tasks from BIG-Bench that remain unsolved or challenging for state-of-the-art models.  
Many of these challenges relate to \gls{co}, where finding optimal solutions in complex search spaces is a key difficulty.
}

\subsection{Application Domains}
\label{sec:application}

\gls{co} has been applied to a wide range of problems across various domains, and over the years, these problems have been formalized into well-known prototypical models, \new{such as} the \gls{tsp}, the \gls{vrp}, and the \gls{cvpr} in the field of routing.
Among the \numstudiesincluded studies, \new{64} explicitly address problems within specific domains or problem formulations \new{-- which we overview in this section}. In this analysis, literature reviews that provide information on the use of \gls{llm} without offering actual implementations are excluded, such as those by \citet{fan2024artificial}, \citet{wu2024evolutionary}, and \citet{zhao2024survey}. 
Exceptions apply to reviews like the one by \citet{SAKA2024100300} that also verified the application of \glspl{llm} in \gls{co} within the construction domain.

The majority of these studies (\new{26}) focus on routing problems, particularly within the contexts of the \gls{tsp} \cite{yang2024large,liu2024large,LIU2023100520,liu2023algorithm,liu2024evolution,ye2024large,make6030093,Khan_2024,huang2024exploringtruepotentialevaluating,yao2024multiobjectiveevolutionheuristicusing,vanstein2024intheloophyperparameteroptimizationllmbased,sui-2024,chen2024uberuncertaintybasedevolutionlarge,liu2024llm4adplatformalgorithmdesign,DBLP:conf/esann/MartinekLG24,10.1007/978-3-031-70068-2_12}, \gls{vrp}  \cite{10.1007/978-981-97-2259-4_3,huang2024words,huang2024multimodal,chin2024learning,ye2024large,a17120582, Da2024,Khan_2024,liu2024llm4adplatformalgorithmdesign,wu2024neuralcombinatorialoptimizationalgorithms}, orienteering \cite{ye2024large} , and \new{traveling} \cite{ju-etal-2024-globe,delarosa2024trippaltravelplanningguarantees,10704489} . \new{While the first three \glspl{cop} are well-known in the \gls{co} field, the latter refers to the modeling of the traveling activity (i.e., visiting a new city) as a \gls{cop}. It cannot be considered as a \gls{tsp} variant in all cases, as it is not given that the goal is to find a Hamiltonian path.}
Two other domains that have garnered significant attention are scheduling\new{/planning (14 studies) and networks and graphs (10).} 
\new{In the first domain, studies have focused on variants of well-known benchmark problems, like the \gls{pfsp} \cite{amarasinghe2023aicopilot,liu2024evolution} and planning \cite{buildings13071772,10.1609/icaps.v34i1.31503,10803039,10675146,10711695,jiang2024largelanguagemodelscombinatorial,bohnet2024exploringbenchmarkingplanningcapabilities,hao2025planningrigorgeneralpurposezeroshot,Wang_2024}, as well as specific scheduling problems, like server \cite{yatong2024tseohedgeservertask} and meeting/conference \cite{lawless2024i,10.1145/3664646.3665084} scheduling.} 
\new{Also for what concerns the Network and Graph domains, studies focused on general network design \cite{yu2024autornetautomaticallyoptimizingheuristics,10829820,hu2024scalableaccurategraphreasoning,li2025graphteamfacilitatinglargelanguage,smartcities7050094}, but also on classical problems like coloring \cite{DBLP:conf/esann/MartinekLG24}, social networks \citet{10818476}, critical node identification \cite{mao2024identify}, and path finding \footnotesize \cite{Khan_2024,borazjanizadeh2024navigatinglabyrinthevaluatingenhancing}. } 

\new{The remaining studies focused on packing (9), combinatorics (4), engineering (3), finance (3), bioinformatics (3), supply chain (1), and strings (1). For a breakdown of specific \glspl{cop}, we refer the interested reader to \Cref{app:application-analysis}.}  

A limited number of studies (\new{9}) explore multiple problem domains and formulations to demonstrate the capabilities of \glspl{llm} across various contexts \new{\cite{ye2024large, liu2024evolution,
Romera-Paredes2024,
yao2024multiobjectiveevolutionheuristicusing,
borazjanizadeh2024navigatinglabyrinthevaluatingenhancing,
chen2024uberuncertaintybasedevolutionlarge,
vanstein2024intheloophyperparameteroptimizationllmbased,
DBLP:conf/esann/MartinekLG24,
10.1007/978-3-031-70068-2_12}}. 
\new{Additionally, \citet{Khan_2024} tackled a diverse set of problems, which however can all be reduced to graph-based ones, while studies like the one by \citet{lawless2024llmscoldstartcuttingplane} inherently address multiple problem domains as they are based on library of problems (initially not thought for applications in the field of \glspl{llm}), like the MIPLIB dataset.}

\subsection{Positions from Non-experimental Literature}
\label{sec:position-papers}

Among the \numstudiesincluded selected studies, \new{11} are literature reviews, \new{7} are position papers, and 1 is a technical report. 

\citet{fan2024artificial}, \citet{wu2024evolutionary}, \citet{huang2024large}, \new{\citet{han-2024-syrvey}, \citet{10.1145/3638530.3664086}}, \new{and \citet{liu2024systematicsurveylargelanguage}} presented literature reviews on the topic of \gls{llm} and optimization \new{or on strictly related fields, such as algorithm design} (\Cref{sec:related-work}).
\citet{SAKA2024100300}, \citet{zhao2024survey}, \new{\citet{wu2024neuralcombinatorialoptimizationalgorithms}, \citet{10.1609/icaps.v34i1.31503}, \citet{sui-2024}, and \citet{10829820}} presented literature reviews that deal with \gls{co} in specific application domains (i.e., engineering, planning, \new{\gls{vrp}, planning and scheduling, \gls{tsp}, and network optimization}, respectively) and \new{in some cases} involve (preliminary) studies and experiments \new{with \gls{llm}}. 

\citet{wasserkrug2024large} put forward a position paper advocating for the usage of \new{an} \gls{llm} within optimization considering all the optimization steps. Similarly, \citet{tsouros2023holy} proposed a \gls{llm}-aided optimization pipeline in the context of \gls{cp} modeling.
\citet{Freuder_2024} highlighted the potential of \glspl{llm} in facilitating the discussion between optimization and domain experts.
\new{\citet{srivastava2024casedevelopingfoundationmodel}, \citet{yu2024deepinsightsautomatedoptimization}, and \citet{10720437} presented insights on the usage of \glspl{llm} within \glspl{ea}, planning-like tasks, and automated \gls{milp} configuration. }

The only technical report identified is by \citet{ramamonjison-etal-2022-augmenting}. The report includes insights and comparisons related to the \gls{nl4opt} challenge (\Cref{sec:optimization-process-analysis,sec:datasets}).

\section{Future Research Directions} 
\label{sec:future-research-directions}

\new{While various aspects of the optimization process have been addressed in the existing literature (\Cref{sec:analysis}), certain areas still call for further exploration. We have identified several future research directions:}

\begin{itemize}[label={--}, leftmargin=2em]
    \item \textit{Metaheuristics}: The adoption of \glspl{llm} in \gls{mh} frameworks has been limited and scattered. Future research could investigate how \glspl{llm} can be used to dynamically adjust \gls{mh} strategies, optimizing parameters \new{(as anticipated in some studies  \cite{a17120582,DBLP:conf/esann/MartinekLG24,lawless2024llmscoldstartcuttingplane})} or switching strategies based on the current state \new{of the search}. This could enhance the flexibility and effectiveness of \glspl{mh}. Future studies could explore how \glspl{llm} might expand local search neighborhoods by suggesting structures or transition operators.

    \item \textit{Algorithm Selection}: Utilizing \glspl{llm} to explore the search space of algorithm selection might be winning (\new{as anticipated by \citet{10.24963/ijcai.2024/579}}) and \glspl{llm} could help pinpoint scenarios where current solvers are less effective. Additionally, \glspl{llm} could be used to generate instances specifically designed to test the strengths and weaknesses of these solvers, leading to improved performance insights. 

    \item \textit{Synthetic Instance Generation}: \glspl{llm} could be leveraged to create synthetic instances that replicate the complexities of real-world problems or that are specifically crafted to challenge existing algorithms. This approach \new{has been anticipated in the IndustryOR dataset \cite{huang2025orlmcustomizableframeworktraining}}.

    \item \textit{\glspl{llm} behavior w.r.t. \gls{nlp} problem description}: \new{While \gls{llm} capabilities w.r.t. visual representations in \gls{co} has been addressed \cite{make6030093}, it remains unclear to what extent \glspl{llm} adjust their responses depending on the input style thus representing a promising research area}. 

    \item \textit{Evaluation Protocol}: \new{Evaluating the performance of \glspl{llm} on \glspl{cop} remains challenging due to several factors: problems can have multiple representations, various encoding methods, and often lack known optimal solutions. While some efforts exist, a promising research direction involves developing standardized evaluation protocols and identifying suitable metrics.}  

    \item \textit{Agent-based System and \glspl{cop}}: \new{Recent studies have explored how \glspl{llm} can function as autonomous agents \cite{LeiWang_2024}. While some of the studies already use an ensemble of \glspl{llm}, a promising research direction would be to investigate thoroughly how these autonomous agents can be effectively utilized within \gls{co}.} 
\end{itemize}

Additional \new{more general} considerations for future research include ethical \new{and bias} considerations.
As \glspl{llm} are integrated into optimization frameworks, research should also focus on ensuring that these technologies are used responsibly. This includes considering the environmental impact of the resources required by \glspl{llm} and developing more sustainable approaches to large-scale optimization. 
Future research should examine the potential biases in \gls{llm}-generated optimization solutions. More generally, strategies for ensuring fairness as well as incorporating ethical considerations into the optimization process could be explored.

\section{Limitations}
\label{sec:limitations}

\new{While our systematic review offers valuable insights into the use of \glspl{llm} in \gls{co}, it has some limitations.  
First, research on \glspl{llm} and their application to optimization is advancing rapidly, with new studies emerging continuously. }
Consequently, while this systematic review attempts to be as comprehensive as possible, it inevitably cannot cover all existing studies.
Also, systematic reviews depend on data gathered from other studies, meaning the quality and bias level of the evidence in a systematic review are directly tied to those of the data sources \cite{DRUCKER2016e109}. Therefore, we acknowledge that following \gls{prisma} guidelines might have led to the inclusion of too many studies that report positive outcomes or successful applications of \gls{llm} over those that do not, potentially overstating the effectiveness or applicability of \gls{llm} in this field due to inherent selection bias.

Then, our review focuses solely on how \glspl{llm} can be applied to \gls{co}. An equally significant aspect not covered in this study is how \gls{co} techniques can be employed to enhance \glspl{llm}, which is dual to our main focus. By not addressing this aspect, which can be pursued in future work, our review may miss relationships and interdependencies between the two research fields. \new{Examples of such works include \citet{pan-etal-2024-plum}, who experiment with using \glspl{mh} to engineer \gls{llm} prompts. More specifically, they use -- among other methods -- \gls{hc} and \gls{sa}, to discover and learn new efficient prompts. Another example is the work by \citet{singla2023biobjective}, who model the trade-off between response quality and inference cost of \gls{llm} as a bi-objective combinatorial optimization problem.}
Additionally, this study focuses on \gls{co}, disregarding other types of optimizations, such as \acrlong{col}. Examples of such works are those by \citeauthor{10394233} \cite{10394233, 10.1145/3583133.3596401}.

We intentionally included a significant number of pre-prints alongside peer-reviewed publications \new{(\Cref{sec:methodology})}. Such a decision is driven by the fast-paced nature of research in the field of \glspl{llm}, where discoveries and advancements are rapidly shared through pre\new{-}prints before formal publication (see, for example, \citet{devlin-etal-2019-bert}). Pre-prints allow for timely access to the latest research, innovations, and discussions and are widely used within the \gls{nlp} community. We take no position on the content of pre\new{-}prints found on platforms like \arxiv; instead, we include these documents to maximize the recall of our systematic review. 

Eventually, our systematic review includes studies based on closed-source \glspl{llm}, like ChatGPT. The proprietary nature of these systems often restricts access to their full methodologies and inner workings, which creates significant challenges for understanding and replicating the reported findings. Despite these downsides, we include works dealing both with open- and closed-source \glspl{llm} to maximize the recall of our systematic review.

\section{Conclusions and Future Work}
\label{sec:conclusions}

In this systematic review, we examined the application of \glspl{llm} to \gls{co}. 
Our study summarizes the literature leveraging the \gls{prisma} 2020 guidelines.
To our knowledge, this is the first attempt to comprehensively study the application of \glspl{llm} to \gls{co} and \gls{cop}.
Out of over \new{2,000} collected publications, we included \numstudiesincluded studies in our analysis. These studies were classified based on the task \glspl{llm} performed within the optimization process, the implementation details of the \glspl{llm}, the most commonly used datasets, and the application domains where \glspl{llm}  have been employed in \gls{co} thus far. Additionally, we highlighted the caveats associated with using \glspl{llm} in the field of optimization, outlined future research directions, and exposed the limitations of the present review.

The research on \glspl{llm} is rapidly evolving, thus continuously identifying areas for further research is crucial. 
\new{W}e plan to periodically update our systematic review (as advocated by the \gls{prisma} guidelines \cite{Pagen71}). This process ensures that our review remains relevant in this fast-moving field and allows us to update those studies initially polished as pre-prints (thus solving one of the limitations of our work, \Cref{sec:limitations}).
Future research will also account for the aspects neglected by this paper, such as the integration of \glspl{llm} into optimization paradigms other than \gls{co} and for the dual aspect of optimization used for enhancing \glspl{llm}.

With much of the state-of-the-art work being conducted relying on proprietary models such as ChatGPT, future studies will also focus on developing methods for assessing and reporting on such models to improve the transparency and replicability of the presented studies. This could involve creating frameworks for sharing results that do not compromise proprietary data but still provide sufficient detail for academic reproducibility/replicability (as also advocated by many researchers in the optimization field \cite{swan-2015-research-agenda,swan-2022-mil}).

\new{As \glspl{llm} are deployed in critical areas, their ethical implications, particularly in sensitive optimization tasks, require closer examination especially under the ethical point of view. Finally, addressing biases in \gls{llm} outputs remains a significant concern and is essential for ensuring fair and responsible applications of \glspl{llm} in \gls{co}.
}

\section*{Declaration of Competing Interest}

The authors declare that they have no known competing interests.

\section*{Data Availability}

The data supporting the findings of this study are available within the paper \new{and its appendices}. 

\section*{Acknowledgments}

This research is partially supported by the PRIN 2022 Project – ``MoT—The Measure of Truth: An Evaluation-Centered Machine-Human Hybrid Framework for Assessing Information Truthfulness'' – Code No. 20227F2ZN3, CUP No. G53D23002800006 Funded by the European Union – Next Generation EU – PNRR M4 C2 I1.1., by the Strategic Plan of the University of Udine--Interdepartment Project on Artificial Intelligence (2020-2025), and by the Strategic Plan of the University of Udine (2022-2025).

\bibliographystyle{mystyle}
\bibliography{references}

\begin{thebibliography}{254}
\providecommand{\natexlab}[1]{#1}
\providecommand{\url}[1]{\texttt{#1}}
\providecommand{\urlprefix}{URL }
\expandafter\ifx\csname urlstyle\endcsname\relax
  \providecommand{\doi}[1]{doi:\discretionary{}{}{}#1}\else
  \providecommand{\doi}{doi:\discretionary{}{}{}\begingroup \urlstyle{rm}\Url}\fi

\bibitem[{Abdullin et~al.(2023)Abdullin, Molla, Ofoghi, Yearwood, and Li}]{abdullin-etal-2023-synthetic}
Abdullin, Y., Molla, D., Ofoghi, B., Yearwood, J., Li, Q.: {Synthetic Dialogue Dataset Generation using LLM Agents}. In: Proceedings of the Third Workshop on Natural Language Generation, Evaluation, and Metrics (GEM), pp. 181--191, ACL, Singapore (12 2023), \urlprefix\url{https://aclanthology.org/2023.gem-1.16}

\bibitem[{AhmadiTeshnizi et~al.(2024{\natexlab{a}})AhmadiTeshnizi, Gao, Brunborg, Talaei, and Udell}]{ahmaditeshnizi2024optimus03usinglargelanguage}
AhmadiTeshnizi, A., Gao, W., Brunborg, H., Talaei, S., Udell, M.: {OptiMUS-0.3: Using Large Language Models to Model and Solve Optimization Problems at Scale}. arXiv (2024{\natexlab{a}}), \doi{10.48550/arXiv.2407.19633}

\bibitem[{AhmadiTeshnizi et~al.(2024{\natexlab{b}})AhmadiTeshnizi, Gao, and Udell}]{ahmaditeshnizi2024optimus}
AhmadiTeshnizi, A., Gao, W., Udell, M.: {OptiMUS: Scalable Optimization Modeling with (MI)LP Solvers and Large Language Models}. In: Proceedings of the 41st International Conference on Machine Learning, pp. 1234--1245, ICML '24, JMLR.org, Vienna, Austria (2024{\natexlab{b}}), \doi{10.5555/3692070.3692094}

\bibitem[{Ahmed and Choudhury(2024)}]{ahmed2024lm4opt}
Ahmed, T., Choudhury, S.: {LM4OPT: Unveiling the potential of Large Language Models in formulating mathematical optimization problems}. INFOR: Information Systems and Operational Research \textbf{62}(4), 559--572 (2024), \doi{10.1080/03155986.2024.2388452}

\bibitem[{AI(2024)}]{cohere2024commandrplus}
AI, C.: {Command R+: A Large Language Model for Enterprise AI} (2024), \urlprefix\url{https://docs.cohere.com/v2/docs/command-r-plus}

\bibitem[{Alipour-Vaezi and Tsui(2024)}]{ALIPOURVAEZI2024110574}
Alipour-Vaezi, M., Tsui, K.L.: {Data-driven portfolio management for motion pictures industry: A new data-driven optimization methodology using a large language model as the expert}. Computers \& Industrial Engineering \textbf{197}, 110574 (2024), \doi{10.1016/j.cie.2024.110574}

\bibitem[{Almonacid(2023)}]{almonacid2023automatic}
Almonacid, B.: {Towards an Automatic Optimisation Model Generator Assisted with Generative Pre-trained Transformer}. arXiv (2023), \doi{10.48550/arXiv.2305.05811}

\bibitem[{Amarasinghe et~al.(2023)Amarasinghe, Nguyen, Sun, and Alahakoon}]{amarasinghe2023aicopilot}
Amarasinghe, P.T., Nguyen, S., Sun, Y., Alahakoon, D.: {AI-Copilot for Business Optimisation: A Framework and A Case Study in Production Scheduling}. arXiv (2023), \doi{10.48550/arXiv.2309.13218}

\bibitem[{Anagnostidis and Bulian(2024)}]{anagnostidis2024susceptible}
Anagnostidis, S., Bulian, J.: {How Susceptible are LLMs to Influence in Prompts?} arXiv (2024), \doi{10.48550/arXiv.2408.11865}

\bibitem[{Anthropic(2024)}]{anthropic2024claude35}
Anthropic: {Claude 3.5 Sonnet Model Card Addendum} (2024), \urlprefix\url{https://www.anthropic.com/news/claude-3-5-sonnet}

\bibitem[{{Anthropic Team}(2024)}]{anthropic2024claude}
{Anthropic Team}: {Introducing the next generation of Claude} (March 2024), \urlprefix\url{https://www.anthropic.com/news/claude-3-family}, accessed: 2024-06-18

\bibitem[{Bahdanau et~al.(2016)Bahdanau, Cho, and Bengio}]{bahdanau2014neural}
Bahdanau, D., Cho, K., Bengio, Y.: {Neural Machine Translation by Jointly Learning to Align and Translate}. arXiv (2016), \doi{10.48550/arXiv.1409.0473}

\bibitem[{Balloccu et~al.(2024)Balloccu, Schmidtov{\'a}, Lango, and Dusek}]{balloccu2024leak}
Balloccu, S., Schmidtov{\'a}, P., Lango, M., Dusek, O.: {Leak, Cheat, Repeat: Data Contamination and Evaluation Malpractices in Closed-Source LLMs}. In: Proceedings of the 18th Conference of the European Chapter of the ACL (Volume 1: Long Papers), pp. 67--93, ACL, St. Julian{'}s, Malta (3 2024), \urlprefix\url{https://aclanthology.org/2024.eacl-long.5}

\bibitem[{Bengio et~al.(2021)Bengio, Lodi, and Prouvost}]{BENGIO2021405}
Bengio, Y., Lodi, A., Prouvost, A.: Machine learning for combinatorial optimization: A methodological tour d’horizon. European Journal of Operational Research \textbf{290}(2), 405--421 (2021), ISSN 0377-2217, \doi{10.1016/j.ejor.2020.07.063}

\bibitem[{Besta et~al.(2024)Besta, Blach, Kubicek, Gerstenberger, Podstawski, Gianinazzi, Gajda, Lehmann, Niewiadomski, Nyczyk, and Hoefler}]{Besta_Blach_Kubicek_Gerstenberger_Podstawski_Gianinazzi_Gajda_Lehmann_Niewiadomski_Nyczyk_Hoefler_2024}
Besta, M., Blach, N., Kubicek, A., Gerstenberger, R., Podstawski, M., Gianinazzi, L., Gajda, J., Lehmann, T., Niewiadomski, H., Nyczyk, P., Hoefler, T.: Graph of thoughts: Solving elaborate problems with large language models. Proceedings of the AAAI Conference on Artificial Intelligence \textbf{38}(16), 17682--17690 (3 2024), \doi{10.1609/aaai.v38i16.29720}

\bibitem[{Bifulco et~al.(2024)Bifulco, Errica, Sanvito, and Siracusano}]{errica2024did}
Bifulco, R., Errica, F., Sanvito, D., Siracusano, G.: {What Did I Do Wrong? Quantifying LLMs' Sensitivity and Consistency to Prompt Engineering}. arXiv (2024), \doi{10.48550/arXiv.2406.12334}, \urlprefix\url{https://arxiv.org/abs/2406.12334}

\bibitem[{Blank and Deb(2020)}]{9078759}
Blank, J., Deb, K.: {Pymoo: Multi-Objective Optimization in Python}. IEEE Access \textbf{8}, 89497--89509 (2020), \doi{10.1109/ACCESS.2020.2990567}

\bibitem[{Blum and Roli(2003)}]{10.1145/937503.937505}
Blum, C., Roli, A.: Metaheuristics in combinatorial optimization: Overview and conceptual comparison. ACM Computing Surveys \textbf{35}(3), 268–308 (9 2003), ISSN 0360-0300, \doi{10.1145/937503.937505}

\bibitem[{Bohnet et~al.(2024)Bohnet, Nova, Parisi, Swersky, Goshvadi, Dai, Schuurmans, Fiedel, and Sedghi}]{bohnet2024exploringbenchmarkingplanningcapabilities}
Bohnet, B., Nova, A., Parisi, A.T., Swersky, K., Goshvadi, K., Dai, H., Schuurmans, D., Fiedel, N., Sedghi, H.: {Exploring and Benchmarking the Planning Capabilities of Large Language Models}. arXiv (2024), \doi{10.48550/arXiv.2406.13094}

\bibitem[{Borazjanizadeh et~al.(2024)Borazjanizadeh, Herzig, Darrell, Feris, and Karlinsky}]{borazjanizadeh2024navigatinglabyrinthevaluatingenhancing}
Borazjanizadeh, N., Herzig, R., Darrell, T., Feris, R., Karlinsky, L.: {Navigating the Labyrinth: Evaluating and Enhancing LLMs' Ability to Reason About Search Problems}. arXiv (2024), \doi{10.48550/arXiv.2406.12172}

\bibitem[{Brown et~al.(2020)Brown, Mann, Ryder, Subbiah, Kaplan, Dhariwal, Neelakantan, Shyam, Sastry, Askell et~al.}]{NEURIPS2020_1457c0d6}
Brown, T., Mann, B., Ryder, N., Subbiah, M., Kaplan, J.D., Dhariwal, P., Neelakantan, A., Shyam, P., Sastry, G., Askell, A., et~al.: {Language Models are Few-Shot Learners}. In: Advances in Neural Information Processing Systems, vol.~33, pp. 1877--1901, Curran Associates, Inc., Virtual Conference (2020), \urlprefix\url{https://proceedings.neurips.cc/paper_files/paper/2020/file/1457c0d6bfcb4967418bfb8ac142f64a-Paper.pdf}

\bibitem[{Bubeck et~al.(2023)Bubeck, Chandrasekaran, Eldan, Gehrke, Horvitz, Kamar, Lee, Lee, Li, Lundberg, Nori, Palangi, Ribeiro, and Zhang}]{bubeck2023sparks}
Bubeck, S., Chandrasekaran, V., Eldan, R., Gehrke, J., Horvitz, E., Kamar, E., Lee, P., Lee, Y.T., Li, Y., Lundberg, S., Nori, H., Palangi, H., Ribeiro, M.T., Zhang, Y.: {Sparks of Artificial General Intelligence: Early experiments with GPT-4}. arXiv (2023), \doi{10.48550/arXiv.2303.12712}

\bibitem[{Cai et~al.(2024{\natexlab{a}})Cai, Xu, Li, Yamauchi, Iba, and Tei}]{10.1145/3638530.3664086}
Cai, J., Xu, J., Li, J., Yamauchi, T., Iba, H., Tei, K.: {Exploring the Improvement of Evolutionary Computation via Large Language Models}. In: Proceedings of the Genetic and Evolutionary Computation Conference Companion, p. 83–84, GECCO '24 Companion, ACM, New York, NY, USA (2024{\natexlab{a}}), ISBN 9798400704956, \doi{10.1145/3638530.3664086}

\bibitem[{Cai et~al.(2024{\natexlab{b}})Cai, Cao, Chen, Chen, Chen, Chen, Chen, Chen, Chen, Chu et~al.}]{cai2024internlm2technicalreport}
Cai, Z., Cao, M., Chen, H., Chen, K., Chen, K., Chen, X., Chen, X., Chen, Z., Chen, Z., Chu, P., et~al.: {InternLM2 Technical Report}. arXiv (2024{\natexlab{b}}), \doi{10.48550/arXiv.2403.17297}

\bibitem[{Ceschia et~al.(2023{\natexlab{a}})Ceschia, Di~Gaspero, Mazzaracchio, Policante, and Schaerf}]{CESCHIA2023100379}
Ceschia, S., Di~Gaspero, L., Mazzaracchio, V., Policante, G., Schaerf, A.: Solving a real-world nurse rostering problem by simulated annealing. Operations Research for Health Care \textbf{36}, 100379 (2023{\natexlab{a}}), ISSN 2211-6923, \doi{10.1016/j.orhc.2023.100379}

\bibitem[{Ceschia et~al.(2023{\natexlab{b}})Ceschia, {Di Gaspero}, and Schaerf}]{CESCHIA20231}
Ceschia, S., {Di Gaspero}, L., Schaerf, A.: {Educational timetabling: Problems, benchmarks, and state-of-the-art results}. European Journal of Operational Research \textbf{308}(1), 1--18 (2023{\natexlab{b}}), ISSN 0377-2217, \doi{10.1016/j.ejor.2022.07.011}

\bibitem[{Ceschia and Schaerf(2024)}]{CESCHIA2024109858}
Ceschia, S., Schaerf, A.: {Multi-neighborhood simulated annealing for the capacitated facility location problem with customer incompatibilities}. Computers \& Industrial Engineering \textbf{188}, 109858 (2024), ISSN 0360-8352, \doi{10.1016/j.cie.2023.109858}

\bibitem[{Chac\'{o}n~Sartori et~al.(2024)Chac\'{o}n~Sartori, Blum, and Ochoa}]{sartori2024large}
Chac\'{o}n~Sartori, C., Blum, C., Ochoa, G.: {Large Language Models for the Automated Analysis of Optimization Algorithms}. In: Proceedings of the Genetic and Evolutionary Computation Conference, pp. 160--168, GECCO '24, ACM, New York, NY, USA (2024), \doi{10.1145/3638529.3654086}

\bibitem[{Chen et~al.(2024{\natexlab{a}})Chen, Li, Tang, and Li}]{chen2024graphwizinstructionfollowinglanguagemodel}
Chen, N., Li, Y., Tang, J., Li, J.: {GraphWiz: An Instruction-Following Language Model for Graph Problems}. arXiv (2024{\natexlab{a}}), \doi{10.48550/arXiv.2402.16029}

\bibitem[{Chen et~al.(2024{\natexlab{b}})Chen, Zhou, Lu, Xu, Pan, and Lan}]{chen2024uberuncertaintybasedevolutionlarge}
Chen, Z., Zhou, Z., Lu, Y., Xu, R., Pan, L., Lan, Z.: {UBER: Uncertainty-Based Evolution with Large Language Models for Automatic Heuristic Design}. arXiv (2024{\natexlab{b}}), \doi{10.48550/arXiv.2412.20694}

\bibitem[{Chiang et~al.(2024)Chiang, Zheng, Sheng, Angelopoulos, Li, Li, Zhang, Zhu, Jordan, Gonzalez, Stoica et~al.}]{chiang2024chatbotarenaopenplatform}
Chiang, W.L., Zheng, L., Sheng, Y., Angelopoulos, A.N., Li, T., Li, D., Zhang, H., Zhu, B., Jordan, M., Gonzalez, J.E., Stoica, I., et~al.: {Chatbot Arena: An Open Platform for Evaluating LLMs by Human Preference}. arXiv (2024), \doi{10.48550/arXiv.2403.04132}

\bibitem[{Chin et~al.(2024)Chin, Winkenbach, and Srivastava}]{chin2024learning}
Chin, S.J.K., Winkenbach, M., Srivastava, A.: {Learning to Deliver: a Foundation Model for the Montreal Capacitated Vehicle Routing Problem}. arXiv (2024), \doi{10.48550/arXiv.2403.00026}

\bibitem[{Chowdhery et~al.(2024)Chowdhery, Narang, Devlin, Bosma, Mishra, Roberts, Barham, Chung, Sutton, Gehrmann et~al.}]{10.5555/3648699.3648939}
Chowdhery, A., Narang, S., Devlin, J., Bosma, M., Mishra, G., Roberts, A., Barham, P., Chung, H.W., Sutton, C., Gehrmann, S., et~al.: {PaLM: scaling language modeling with pathways}. Journal of Machine Learning Research \textbf{24}(1) (3 2024), ISSN 1532-4435, \urlprefix\url{http://jmlr.org/papers/v24/22-1144.html}

\bibitem[{Chung et~al.(2024)Chung, Hou, Longpre, Zoph, Tay, Fedus, Li, Wang, Dehghani, Brahma,  et~al.}]{JMLR:v25:23-0870}
Chung, H.W., Hou, L., Longpre, S., Zoph, B., Tay, Y., Fedus, W., Li, Y., Wang, X., Dehghani, M., Brahma, S., , et~al.: Scaling instruction-finetuned language models. Journal of Machine Learning Research \textbf{25}(70), 1--53 (2024), \urlprefix\url{http://jmlr.org/papers/v25/23-0870.html}

\bibitem[{Cobbe et~al.(2021)Cobbe, Kosaraju, Bavarian, Chen, Jun, Kaiser, Plappert, Tworek, Hilton, Nakano, Hesse, and Schulman}]{cobbe2021trainingverifierssolvemath}
Cobbe, K., Kosaraju, V., Bavarian, M., Chen, M., Jun, H., Kaiser, L., Plappert, M., Tworek, J., Hilton, J., Nakano, R., Hesse, C., Schulman, J.: {Training Verifiers to Solve Math Word Problems}. arXiv (2021), \doi{10.48550/arXiv.2110.14168}

\bibitem[{Cooper et~al.(2017)Cooper, Booth, Britten, and Garside}]{Cooper2017}
Cooper, C., Booth, A., Britten, N., Garside, R.: {A comparison of results of empirical studies of supplementary search techniques and recommendations in review methodology handbooks: a methodological review}. Systematic Reviews \textbf{6}(1), 234 (Nov 2017), ISSN 2046-4053, \doi{10.1186/s13643-017-0625-1}

\bibitem[{Da et~al.(2024)Da, Liou, Chen, Zhou, Luo, Yang, and Wei}]{Da2024}
Da, L., Liou, K., Chen, T., Zhou, X., Luo, X., Yang, Y., Wei, H.: {Open-ti: open traffic intelligence with augmented language model}. International Journal of Machine Learning and Cybernetics \textbf{15}(10), 4761--4786 (Oct 2024), ISSN 1868-808X, \doi{10.1007/s13042-024-02190-8}

\bibitem[{Dakle et~al.(2023)Dakle, Kad{\i}o{\u{g}}lu, Uppuluri, Politi, Raghavan, Rallabandi, and Srinivasamurthy}]{10.1007/978-3-031-33271-5_20}
Dakle, P.P., Kad{\i}o{\u{g}}lu, S., Uppuluri, K., Politi, R., Raghavan, P., Rallabandi, S., Srinivasamurthy, R.: {Ner4Opt: Named Entity Recognition for Optimization Modelling from Natural Language}. In: Integration of Constraint Programming, Artificial Intelligence, and Operations Research, pp. 299--319, Springer Nature Switzerland, Cham (2023), ISBN 978-3-031-33271-5, \doi{10.1007/978-3-031-33271-5_20}

\bibitem[{{De La Rosa} et~al.(2024){De La Rosa}, Gopalakrishnan, Pozanco, Zeng, and Borrajo}]{delarosa2024trippaltravelplanningguarantees}
{De La Rosa}, T., Gopalakrishnan, S., Pozanco, A., Zeng, Z., Borrajo, D.: {TRIP-PAL: Travel Planning with Guarantees by Combining Large Language Models and Automated Planners}. arXiv (2024), \doi{10.48550/arXiv.2406.10196}

\bibitem[{De~Zarzà et~al.(2024)De~Zarzà, De~Curtò, Roig, and Calafate}]{ai5010006}
De~Zarzà, I., De~Curtò, J., Roig, G., Calafate, C.T.: {Optimized Financial Planning: Integrating Individual and Cooperative Budgeting Models with LLM Recommendations}. AI \textbf{5}(1), 91--114 (2024), ISSN 2673-2688, \doi{10.3390/ai5010006}

\bibitem[{DeepSeek-AI et~al.(2024{\natexlab{a}})DeepSeek-AI, Bi, Chen, Chen, Chen, Dai, Deng, Ding, Dong, and Du}]{deepseekai2024deepseekllmscalingopensource}
DeepSeek-AI, Bi, X., Chen, D., Chen, G., Chen, S., Dai, D., Deng, C., Ding, H., Dong, K., Du, Q.: {DeepSeek LLM: Scaling Open-Source Language Models with Longtermism}. arXiv (2024{\natexlab{a}}), \doi{10.48550/arXiv.2401.02954}

\bibitem[{DeepSeek-AI et~al.(2024{\natexlab{b}})DeepSeek-AI, Liu, Feng, Wang, Wang, Liu, Zhao, Deng, Ruan, and Dai}]{deepseekai2024deepseekv2strongeconomicalefficient}
DeepSeek-AI, Liu, A., Feng, B., Wang, B., Wang, B., Liu, B., Zhao, C., Deng, C., Ruan, C., Dai, D.: {DeepSeek-V2: A Strong, Economical, and Efficient Mixture-of-Experts Language Model}. arXiv (2024{\natexlab{b}}), \doi{10.48550/arXiv.2405.04434}

\bibitem[{DeepSeek-AI et~al.(2024{\natexlab{c}})DeepSeek-AI, Zhu, Guo, Shao, Yang, Wang, Xu, Wu, Li, and Gao}]{deepseekai2024deepseekcoderv2breakingbarrierclosedsource}
DeepSeek-AI, Zhu, Q., Guo, D., Shao, Z., Yang, D., Wang, P., Xu, R., Wu, Y., Li, Y., Gao, H.: {DeepSeek-Coder-V2: Breaking the Barrier of Closed-Source Models in Code Intelligence}. arXiv (2024{\natexlab{c}}), \doi{10.48550/arXiv.2406.11931}

\bibitem[{Devlin et~al.(2019)Devlin, Chang, Lee, and Toutanova}]{devlin-etal-2019-bert}
Devlin, J., Chang, M.W., Lee, K., Toutanova, K.: {BERT: Pre-training of Deep Bidirectional Transformers for Language Understanding}. In: Proceedings of the 2019 Conference of the North {A}merican Chapter of the ACL: Human Language Technologies, Volume 1 (Long and Short Papers), pp. 4171--4186, ACL, Minneapolis, Minnesota (6 2019), \doi{10.18653/v1/N19-1423}

\bibitem[{Dhanaraj et~al.(2024)Dhanaraj, Jeon, Kang, Nikolaidis, and Gupta}]{10711695}
Dhanaraj, N., Jeon, M., Kang, J.H., Nikolaidis, S., Gupta, S.K.: {Preference Elicitation and Incorporation for Human-Robot Task Scheduling}. In: 2024 IEEE 20th International Conference on Automation Science and Engineering (CASE), pp. 3103--3110, IEEE, Bari, Italy (2024), \doi{10.1109/CASE59546.2024.10711695}

\bibitem[{Diamond and Boyd(2016)}]{10.5555/2946645.3007036}
Diamond, S., Boyd, S.: {CVXPY: a python-embedded modeling language for convex optimization}. {The Journal of Machine Learning Research} \textbf{17}(1), 2909–2913 (1 2016), ISSN 1532-4435, \doi{10.5555/2946645.3007036}

\bibitem[{Doan(2022)}]{doan2022vtccnlp}
Doan, X.D.: {VTCC-NLP at NL4Opt competition subtask 1: An Ensemble Pre-trained language models for Named Entity Recognition}. arXiv (2022), \doi{10.48550/arXiv.2212.07219}

\bibitem[{Dorigo et~al.(2006)Dorigo, Birattari, and Stutzle}]{dorigo-2004-aco}
Dorigo, M., Birattari, M., Stutzle, T.: {Ant colony optimization}. IEEE Computational Intelligence Magazine \textbf{1}(4), 28--39 (2006), \doi{10.1109/MCI.2006.329691}

\bibitem[{Driess et~al.(2023)Driess, Xia, Sajjadi, Lynch, Chowdhery, Ichter, Wahid, Tompson, Vuong, Yu et~al.}]{10.5555/3618408.3618748}
Driess, D., Xia, F., Sajjadi, M.S.M., Lynch, C., Chowdhery, A., Ichter, B., Wahid, A., Tompson, J., Vuong, Q., Yu, T., et~al.: {PaLM-E: an embodied multimodal language model}. In: Proceedings of the 40th International Conference on Machine Learning, ICML'23, JMLR.org, Honolulu, Hawaii, USA (2023), \doi{10.5555/3618408.3618748}

\bibitem[{Drucker et~al.(2016)Drucker, Fleming, and Chan}]{DRUCKER2016e109}
Drucker, A.M., Fleming, P., Chan, A.W.: {Research Techniques Made Simple: Assessing Risk of Bias in Systematic Reviews}. Journal of Investigative Dermatology \textbf{136}(11), e109--e114 (2016), ISSN 0022-202X, \doi{10.1016/j.jid.2016.08.021}

\bibitem[{Du et~al.(2022)Du, Qian, Liu, Ding, Qiu, Yang, and Tang}]{du2022glm}
Du, Z., Qian, Y., Liu, X., Ding, M., Qiu, J., Yang, Z., Tang, J.: {{GLM}: General Language Model Pretraining with Autoregressive Blank Infilling}. In: Proceedings of the 60th Annual Meeting of the ACL (Volume 1: Long Papers), pp. 320--335, ACL, Dublin, Ireland (2022), \doi{10.18653/v1/2022.acl-long.26}

\bibitem[{Elhenawy et~al.(2024)Elhenawy, Abutahoun, Alhadidi, Jaber, Ashqar, Jaradat, Abdelhay, Glaser, and Rakotonirainy}]{make6030093}
Elhenawy, M., Abutahoun, A., Alhadidi, T.I., Jaber, A., Ashqar, H.I., Jaradat, S., Abdelhay, A., Glaser, S., Rakotonirainy, A.: {Visual Reasoning and Multi-Agent Approach in Multimodal Large Language Models (MLLMs): Solving TSP and mTSP Combinatorial Challenges}. Machine Learning and Knowledge Extraction \textbf{6}(3), 1894--1920 (2024), \doi{10.3390/make6030093}

\bibitem[{Fan et~al.(2024)Fan, Ghaddar, Wang, Xing, Zhang, and Zhou}]{fan2024artificial}
Fan, Z., Ghaddar, B., Wang, X., Xing, L., Zhang, Y., Zhou, Z.: {Artificial Intelligence for Operations Research: Revolutionizing the Operations Research Process}. arXiv (2024), \doi{10.48550/arXiv.2401.03244}

\bibitem[{Fatemi et~al.(2023)Fatemi, Halcrow, and Perozzi}]{fatemi2023talklikegraphencoding}
Fatemi, B., Halcrow, J., Perozzi, B.: {Talk like a Graph: Encoding Graphs for Large Language Models}. arXiv (2023), \doi{10.48550/arXiv.2310.04560}

\bibitem[{Franzin and Stützle(2019)}]{franzin-2019-sa-component}
Franzin, A., Stützle, T.: {Revisiting simulated annealing: A component-based analysis}. Computers \& Operations Research \textbf{104}, 191--206 (2019), ISSN 0305-0548, \doi{10.1016/j.cor.2018.12.015}

\bibitem[{Freire et~al.(2024)Freire, Santamaría~Laorden, Orejas~Pérez, Gómez~Sánchez, Díaz-Flores~García, and Suárez}]{FREIRE2024659e1}
Freire, Y., Santamaría~Laorden, A., Orejas~Pérez, J., Gómez~Sánchez, M., Díaz-Flores~García, V., Suárez, A.: {ChatGPT performance in prosthodontics: Assessment of accuracy and repeatability in answer generation}. {The Journal of Prosthetic Dentistry} \textbf{131}(4), 659.e1--659.e6 (2024), ISSN 0022-3913, \doi{10.1016/j.prosdent.2024.01.018}

\bibitem[{Freuder(2024)}]{Freuder_2024}
Freuder, E.C.: {Conversational Modeling for Constraint Satisfaction}. Proceedings of the AAAI Conference on Artificial Intelligence \textbf{38}(20), 22592--22597 (Mar 2024), \doi{10.1609/aaai.v38i20.30268}

\bibitem[{Gangwar and Kani(2023)}]{gangwar2023highlighting}
Gangwar, N., Kani, N.: {Highlighting Named Entities in Input for Auto-formulation of Optimization Problems}. In: Intelligent Computer Mathematics, pp. 130--141, Springer Nature Switzerland, Cham (2023), ISBN 978-3-031-42753-4, \doi{10.1007/978-3-031-42753-4_9}

\bibitem[{Garey et~al.(1976)Garey, Johnson, and Sethi}]{Garey1976117}
Garey, M.R., Johnson, D.S., Sethi, R.: {The Complexity of Flowshop and Jobshop Scheduling}. Mathematics of Operations Research \textbf{1}(2), 117--129 (1976), ISSN 0364-765X, 1526-5471, \urlprefix\url{http://www.jstor.org/stable/3689278}

\bibitem[{{Gecode Team}(2006)}]{gecode}
{Gecode Team}: {Gecode: Generic Constraint Development Environment} (2006), \urlprefix\url{http://www.gecode.org}, accessed: 27-06-2024

\bibitem[{GenAI(2023{\natexlab{a}})}]{ivison2023camelschangingclimateenhancing}
GenAI, M.: {Camels in a Changing Climate: Enhancing LM Adaptation with Tulu 2}. arXiv (2023{\natexlab{a}}), \doi{10.48550/arXiv.2311.10702}

\bibitem[{GenAI(2023{\natexlab{b}})}]{touvron2023llama2}
GenAI, M.: {Llama 2: Open Foundation and Fine-Tuned Chat Models}. arXiv (2023{\natexlab{b}}), \doi{10.48550/arXiv.2307.09288}

\bibitem[{Ghiani et~al.(2024)Ghiani, Solazzo, and Elia}]{a17120582}
Ghiani, G., Solazzo, G., Elia, G.: {Integrating Large Language Models and Optimization in Semi-Structured Decision Making: Methodology and a Case Study}. Algorithms \textbf{17}(12) (2024), \doi{10.3390/a17120582}

\bibitem[{Gjergji and Musliu(2024)}]{10.1007/978-3-031-62922-8_11}
Gjergji, I., Musliu, N.: {Large Neighborhood Search for the Capacitated P-Median Problem}. In: Metaheuristics, pp. 158--173, Springer Nature Switzerland, Cham (2024), ISBN 978-3-031-62922-8, \doi{10.1007/978-3-031-62922-8_11}

\bibitem[{Glover(1997)}]{glover-1997-ts-book}
Glover, F.: {Tabu Search}. Springer, New York, NY (1997), ISBN 978-0-7923-9965-0, \doi{10.1007/978-1-4615-6089-0}

\bibitem[{Google(2023)}]{anil2023palm}
Google: {PaLM 2 Technical Report}. arXiv (2023), \doi{10.48550/arXiv.2305.10403}

\bibitem[{Guns(2019)}]{guns2019increasing}
Guns, T.: {Increasing Modeling Language Convenience with a Universal N-Dimensional Array: CPpy as a Python-Embedded Example}. In: Proceedings of the 18th Workshop on Constraint Modelling and Reformulation at CP (ModRef 2019), ACP, Stamford, CT, USA (2019), \urlprefix\url{https://modref.github.io/ModRef2019.html}

\bibitem[{Guo et~al.(2022)Guo, Lu, Duan, Wang, Zhou, and Yin}]{guo2022unixcoder}
Guo, D., Lu, S., Duan, N., Wang, Y., Zhou, M., Yin, J.: {UniXcoder: Unified Cross-Modal Pre-training for Code Representation}. In: Proceedings of the 60th Annual Meeting of the ACL (Volume 1: Long Papers), pp. 7212--7225, ACL, Dublin, Ireland (may 2022), \doi{10.18653/v1/2022.acl-long.499}

\bibitem[{Guo et~al.(2024)Guo, Chen, Tsai, and Lin}]{guo2024optimizinglargelanguagemodels}
Guo, P.F., Chen, Y.H., Tsai, Y.D., Lin, S.D.: {Towards Optimizing with Large Language Models}. arXiv (2024), \doi{10.48550/arXiv.2310.05204}

\bibitem[{{Gurobi Optimization, LLC}(2023)}]{gurobi}
{Gurobi Optimization, LLC}: {Gurobi Optimizer Reference Manual} (2023), \urlprefix\url{https://www.gurobi.com}

\bibitem[{Hao et~al.(2024)Hao, Zhang, and Fan}]{hao2025planningrigorgeneralpurposezeroshot}
Hao, Y., Zhang, Y., Fan, C.: {Planning Anything with Rigor: General-Purpose Zero-Shot Planning with LLM-based Formalized Programming}. arXiv (2024), \doi{10.48550/arXiv.2410.12112}

\bibitem[{Hao~Chen and Li(2024)}]{chen2023diagnosing}
Hao~Chen, G.E.C.F., Li, C.: {Diagnosing infeasible optimization problems using large language models}. INFOR: Information Systems and Operational Research \textbf{62}(4), 573--587 (2024), \doi{10.1080/03155986.2024.2385189}

\bibitem[{Hart et~al.(2011)Hart, Watson, and Woodruff}]{Hart2011}
Hart, W.E., Watson, J.P., Woodruff, D.L.: Pyomo: modeling and solving mathematical programs in python. Mathematical Programming Computation \textbf{3}(3), 219--260 (Sep 2011), ISSN 1867-2957, \doi{10.1007/s12532-011-0026-8}

\bibitem[{He et~al.(2022)He, N, Vignesh, Kumar, and Uppal}]{he2022linear}
He, J., N, M., Vignesh, S., Kumar, D., Uppal, A.: {Linear programming word problems formulation using EnsembleCRF NER labeler and text generator with data augmentations}. arXiv (2022), \doi{10.48550/arXiv.2212.14657}

\bibitem[{Heinrich et~al.(2022)Heinrich, Hofmann, Baurecht, Kreuzer, Kn{\"u}ttel, Leitzmann, and Seliger}]{Heinrich2022}
Heinrich, M., Hofmann, L., Baurecht, H., Kreuzer, P.M., Kn{\"u}ttel, H., Leitzmann, M.F., Seliger, C.: {Suicide risk and mortality among patients with cancer}. Nature Medicine \textbf{28}(4), 852--859 (Apr 2022), ISSN 1546-170X, \doi{10.1038/s41591-022-01745-y}

\bibitem[{Holland(1992)}]{holland-1992-genetic-algorithm}
Holland, J.H.: {Genetic Algorithms}. Scientific American \textbf{267}(1), 66--73 (1992), ISSN 00368733, 19467087, \urlprefix\url{http://www.jstor.org/stable/24939139}

\bibitem[{Hoos(2012)}]{hoos-2012-pbo}
Hoos, H.H.: Programming by optimization. Communications of the ACM \textbf{55}(2), 70–80 (feb 2012), ISSN 0001-0782, \doi{10.1145/2076450.2076469}

\bibitem[{Hosseini et~al.(2014)Hosseini, Hajishirzi, Etzioni, and Kushman}]{hosseini-etal-2014-learning}
Hosseini, M.J., Hajishirzi, H., Etzioni, O., Kushman, N.: {Learning to Solve Arithmetic Word Problems with Verb Categorization}. In: Proceedings of the 2014 Conference on Empirical Methods in Natural Language Processing ({EMNLP}), pp. 523--533, ACL, Doha, Qatar (Oct 2014), \doi{10.3115/v1/D14-1058}

\bibitem[{Hu et~al.(2021)Hu, Shen, Wallis, Allen-Zhu, Li, Wang, Wang, and Chen}]{hu2021lora}
Hu, E.J., Shen, Y., Wallis, P., Allen-Zhu, Z., Li, Y., Wang, S., Wang, L., Chen, W.: {LoRA: Low-Rank Adaptation of Large Language Models}. arXiv (2021), \doi{10.48550/arXiv.2106.09685}

\bibitem[{Hu et~al.(2024)Hu, Lei, Huang, Wei, and Liu}]{hu2024scalableaccurategraphreasoning}
Hu, Y., Lei, R., Huang, X., Wei, Z., Liu, Y.: {Scalable and Accurate Graph Reasoning with LLM-based Multi-Agents}. arXiv (2024), \doi{10.48550/arXiv.2410.05130}

\bibitem[{Huang et~al.(2024{\natexlab{a}})Huang, Wu, Zhou, Wu, Feng, Cheng, and Tan}]{huang2024exploringtruepotentialevaluating}
Huang, B., Wu, X., Zhou, Y., Wu, J., Feng, L., Cheng, R., Tan, K.C.: {Exploring the True Potential: Evaluating the Black-box Optimization Capability of Large Language Models}. arXiv (2024{\natexlab{a}}), \doi{10.48550/arXiv.2404.06290}

\bibitem[{Huang et~al.(2024{\natexlab{b}})Huang, Tang, Hu, Jiang, Zheng, Ge, Wang, and Wang}]{huang2025orlmcustomizableframeworktraining}
Huang, C., Tang, Z., Hu, S., Jiang, R., Zheng, X., Ge, D., Wang, B., Wang, Z.: {ORLM: A Customizable Framework in Training Large Models for Automated Optimization Modeling}. arXiv (2024{\natexlab{b}}), \doi{10.48550/arXiv.2405.17743}

\bibitem[{Huang et~al.(2024{\natexlab{c}})}]{huang2024opencoder}
Huang, H., et~al.: {The Open Cookbook for Top-Tier Code Large Language Models}. arXiv (2024{\natexlab{c}}), \doi{10.48550/arXiv.2411.04905}

\bibitem[{Huang et~al.(2024{\natexlab{d}})Huang, Yang, Qi, and Wang}]{huang2024large}
Huang, S., Yang, K., Qi, S., Wang, R.: {When large language model meets optimization}. Swarm and Evolutionary Computation \textbf{90}, 101663 (2024{\natexlab{d}}), \doi{10.1016/j.swevo.2024.101663}

\bibitem[{Huang et~al.(2025)Huang, Shen, Hu, Gao, and Wang}]{huang2025llmsmathematicalmodelingbridging}
Huang, X., Shen, Q., Hu, Y., Gao, A., Wang, B.: {LLMs for Mathematical Modeling: Towards Bridging the Gap between Natural and Mathematical Languages}. arXiv (2025), \doi{10.48550/arXiv.2405.13144}

\bibitem[{Huang et~al.(2024{\natexlab{e}})Huang, Zhang, Feng, Wu, and Tan}]{huang2024multimodal}
Huang, Y., Zhang, W., Feng, L., Wu, X., Tan, K.C.: {How Multimodal Integration Boost the Performance of LLM for Optimization: Case Study on Capacitated Vehicle Routing Problems}. arXiv (2024{\natexlab{e}}), \doi{10.48550/arXiv.2403.01757}

\bibitem[{Huang et~al.(2024{\natexlab{f}})Huang, Shi, and Sukhatme}]{huang2024words}
Huang, Z., Shi, G., Sukhatme, G.S.: {From Words to Routes: Applying Large Language Models to Vehicle Routing}. arXiv (2024{\natexlab{f}}), \doi{10.48550/arXiv.2403.10795}

\bibitem[{IBM(2017)}]{IBMsched2017}
IBM: {IBM ILOG CPLEX Optimization Studio, Getting Started with Scheduling in CPLEX Studio}. IBM (2017), \urlprefix\url{https://www.ibm.com/docs/en/icos/20.1.0?topic=kit-getting-started-scheduling-in-cplex-studio}, accessed: 27-06-2024

\bibitem[{Jang(2022)}]{jang2022tag}
Jang, S.: {Tag Embedding and Well-defined Intermediate Representation improve Auto-Formulation of Problem Description}. arXiv (2022), \doi{10.48550/arXiv.2212.03575}

\bibitem[{Jiang et~al.(2023)Jiang, Sablayrolles, Mensch, Bamford, Chaplot, de~las Casas, Bressand, Lengyel, Lample, Saulnier, Renard~Lavaud, Lachaux, Stock, Le~Scao, Lavril, Wang, Lacroix, and El~Sayed}]{jiang2023mistral}
Jiang, A.Q., Sablayrolles, A., Mensch, A., Bamford, C., Chaplot, D.S., de~las Casas, D., Bressand, F., Lengyel, G., Lample, G., Saulnier, L., Renard~Lavaud, L., Lachaux, M.A., Stock, P., Le~Scao, T., Lavril, T., Wang, T., Lacroix, T., El~Sayed, W.: {Mistral 7B}. arXiv (2023), \doi{10.48550/arXiv.2310.06825}

\bibitem[{Jiang et~al.(2024{\natexlab{a}})Jiang, Sablayrolles, Roux, Mensch, Savary, Bamford, Chaplot, de~las Casas, Bou~Hanna, Bressand, Lengyel, Bour, Lample, Renard~Lavaud, Saulnier, Lachaux, Stock, Subramanian, Yang, Antoniak, Le~Scao, Gervet, Lavril, Wang, Lacroix, and El~Sayed}]{jiang2024mixtral}
Jiang, A.Q., Sablayrolles, A., Roux, A., Mensch, A., Savary, B., Bamford, C., Chaplot, D.S., de~las Casas, D., Bou~Hanna, E., Bressand, F., Lengyel, G., Bour, G., Lample, G., Renard~Lavaud, L., Saulnier, L., Lachaux, M.A., Stock, P., Subramanian, S., Yang, S., Antoniak, S., Le~Scao, T., Gervet, T., Lavril, T., Wang, T., Lacroix, T., El~Sayed, W.: {Mixtral of Experts}. arXiv (2024{\natexlab{a}}), \doi{10.48550/arXiv.2401.04088}

\bibitem[{Jiang et~al.(2025)Jiang, Shu, Qian, Lu, Zhou, Zhou, and Yu}]{jiang2025llmopt}
Jiang, C., Shu, X., Qian, H., Lu, X., Zhou, J., Zhou, A., Yu, Y.: {{LLMOPT}: Learning to Define and Solve General Optimization Problems from Scratch}. In: Proceedings of the Thirteenth International Conference on Learning Representations, pp. 3160--3172, Conference Organizers, Singapore (2025), \urlprefix\url{https://openreview.net/forum?id=9OMvtboTJg}

\bibitem[{Jiang et~al.(2024{\natexlab{b}})Jiang, Xie, and Luo}]{jiang2024largelanguagemodelscombinatorial}
Jiang, S., Xie, M., Luo, J.: {Large Language Models for Combinatorial Optimization of Design Structure Matrix}. arXiv (2024{\natexlab{b}}), \doi{10.48550/arXiv.2411.12571}

\bibitem[{Jin et~al.(2024)Jin, Sel, Hardeep, and Yin}]{10738100}
Jin, M., Sel, B., Hardeep, F., Yin, W.: {Democratizing Energy Management with LLM-Assisted Optimization Autoformalism}. In: 2024 IEEE International Conference on Communications, Control, and Computing Technologies for Smart Grids (SmartGridComm), pp. 258--263, IEEE Communications Society, Oslo, Norway (2024), \doi{10.1109/SmartGridComm60555.2024.10738100}

\bibitem[{Jobson and Li(2024)}]{10.1145/3664646.3665084}
Jobson, D., Li, Y.: {Investigating the Potential of Using Large Language Models for Scheduling}. In: Proceedings of the 1st ACM International Conference on AI-Powered Software, p. 170–171, AIware 2024, ACM, New York, NY, USA (2024), ISBN 9798400706851, \doi{10.1145/3664646.3665084}

\bibitem[{Ju et~al.(2024)Ju, Jiang, Cohen, Foss, Mitts, Zharmagambetov, Amos, Li, Kao, Fazel-Zarandi, and Tian}]{ju-etal-2024-globe}
Ju, D., Jiang, S., Cohen, A., Foss, A., Mitts, S., Zharmagambetov, A., Amos, B., Li, X., Kao, J.T., Fazel-Zarandi, M., Tian, Y.: {To the Globe ({TTG}): Towards Language-Driven Guaranteed Travel Planning}. In: Proceedings of the 2024 Conference on Empirical Methods in Natural Language Processing: System Demonstrations, pp. 240--249, ACL, Miami, Florida, USA (Nov 2024), \doi{10.18653/v1/2024.emnlp-demo.25}

\bibitem[{Karimi-Mamaghan et~al.(2022)Karimi-Mamaghan, Mohammadi, Meyer, Karimi-Mamaghan, and Talbi}]{KARIMIMAMAGHAN2022393}
Karimi-Mamaghan, M., Mohammadi, M., Meyer, P., Karimi-Mamaghan, A.M., Talbi, E.G.: {Machine learning at the service of meta-heuristics for solving combinatorial optimization problems: A state-of-the-art}. European Journal of Operational Research \textbf{296}(2), 393--422 (2022), ISSN 0377-2217, \doi{10.1016/j.ejor.2021.04.032}

\bibitem[{Khan and Hamad(2024)}]{Khan_2024}
Khan, M.A., Hamad, L.: {On the Capability of LLMs in Combinatorial Optimization}. TechRxiv (November 2024), \doi{10.36227/techrxiv.173092026.60478567/v1}

\bibitem[{Kikuta et~al.(2024)Kikuta, Ikeuchi, Tajiri, and Nakano}]{10.1007/978-981-97-2259-4_3}
Kikuta, D., Ikeuchi, H., Tajiri, K., Nakano, Y.: {RouteExplainer: An Explanation Framework for Vehicle Routing Problem}. In: Advances in Knowledge Discovery and Data Mining, pp. 30--42, Springer Nature Singapore, Singapore (2024), ISBN 978-981-97-2259-4, \doi{10.1007/978-981-97-2259-4_3}

\bibitem[{Kirkpatrick et~al.(1983)Kirkpatrick, Gelatt, and Vecchi}]{kirkpatrick-1983-sa}
Kirkpatrick, S., Gelatt, C.D., Vecchi, M.P.: {Optimization by Simulated Annealing}. Science \textbf{220}(4598), 671--680 (1983), \doi{10.1126/science.220.4598.671}

\bibitem[{Kletzander and Musliu(2020)}]{KLETZANDER2020104794}
Kletzander, L., Musliu, N.: {Solving the general employee scheduling problem}. Computers \& Operations Research \textbf{113}, 104794 (2020), ISSN 0305-0548, \doi{10.1016/j.cor.2019.104794}

\bibitem[{Kletzander and Musliu(2024)}]{KLETZANDER2024104172}
Kletzander, L., Musliu, N.: {Hyper-heuristics for personnel scheduling domains}. Artificial Intelligence \textbf{334}, 104172 (2024), ISSN 0004-3702, \doi{10.1016/j.artint.2024.104172}

\bibitem[{Kochanek et~al.(2024)Kochanek, Skarzynski, and Jedrzejczak}]{Kochanek2024-cw}
Kochanek, K., Skarzynski, H., Jedrzejczak, W.W.: {Accuracy and Repeatability of ChatGPT Based on a Set of Multiple-Choice Questions on Objective Tests of Hearing}. Cureus \textbf{16}(5) (5 2024), \doi{10.7759/cureus.59857}

\bibitem[{Kojima et~al.(2023)Kojima, Gu, Reid, Matsuo, and Iwasawa}]{kojima2022large}
Kojima, T., Gu, S.S., Reid, M., Matsuo, Y., Iwasawa, Y.: {Large Language Models are Zero-Shot Reasoners}. arXiv (2023), \doi{10.48550/arXiv.2205.11916}

\bibitem[{Koncel-Kedziorski et~al.(2015)Koncel-Kedziorski, Roy, Amini, Kushman, and Hajishirzi}]{koncel-kedziorski-etal-2015-parsing}
Koncel-Kedziorski, R., Roy, S., Amini, A., Kushman, N., Hajishirzi, H.: {Parsing Algebraic Word Problems into Equations}. Transactions of the ACL \textbf{3}, 585--597 (2015), \doi{10.1162/tacl_a_00153}

\bibitem[{Kubiak(2021)}]{KUBIAK202126}
Kubiak, W.: On a conjecture for the university timetabling problem. Discrete Applied Mathematics \textbf{299}, 26--49 (2021), ISSN 0166-218X, \doi{10.1016/j.dam.2021.04.010}

\bibitem[{Lackner et~al.(2023)Lackner, Mrkvicka, Musliu, Walkiewicz, and Winter}]{Lackner2023}
Lackner, M.L., Mrkvicka, C., Musliu, N., Walkiewicz, D., Winter, F.: {Exact methods for the Oven Scheduling Problem}. Constraints \textbf{28}(2), 320--361 (Jun 2023), ISSN 1572-9354, \doi{10.1007/s10601-023-09347-2}

\bibitem[{Laguna et~al.(2023)Laguna, Martí, Martinez-Gavara, Perez-Peló, and Resende}]{laguna202320}
Laguna, M., Martí, R., Martinez-Gavara, A., Perez-Peló, S., Resende, M.G.C.: {20 years of Greedy Randomized Adaptive Search Procedures with Path Relinking}. arXiv (2023), \doi{10.48550/arXiv.2312.12663}

\bibitem[{Lai et~al.(2024)Lai, Wang, Liu, He, Zhang, Liu, and Chen}]{han-2024-syrvey}
Lai, H., Wang, B., Liu, J., He, F., Zhang, C., Liu, H., Chen, H.: {Solving Mathematical Problems Using Large Language Models: A Survey.} SSRN (2024), \doi{10.2139/ssrn.5002356}

\bibitem[{Lawless et~al.(2024{\natexlab{a}})Lawless, Li, Wikum, Udell, and Vitercik}]{lawless2024llmscoldstartcuttingplane}
Lawless, C., Li, Y., Wikum, A., Udell, M., Vitercik, E.: {LLMs for Cold-Start Cutting Plane Separator Configuration}. arXiv (2024{\natexlab{a}}), \doi{10.48550/arXiv.2412.12038}

\bibitem[{Lawless et~al.(2024{\natexlab{b}})Lawless, Schoeffer, Le, Rowan, Sen, St.~Hill, Suh, and Sarrafzadeh}]{lawless2024i}
Lawless, C., Schoeffer, J., Le, L., Rowan, K., Sen, S., St.~Hill, C., Suh, J., Sarrafzadeh, B.: {``I Want It That Way'': Enabling Interactive Decision Support Using Large Language Models and Constraint Programming}. ACM Transactions on Interactive Intelligent Systems \textbf{14}(3), 8432--8448 (Sep 2024{\natexlab{b}}), \doi{10.1145/3685053}

\bibitem[{Le et~al.(2022)Le, Wang, Gotmare, Savarese, and Hoi}]{le2022coderlmasteringcodegeneration}
Le, H., Wang, Y., Gotmare, A.D., Savarese, S., Hoi, S.C.: {CodeRL: mastering code generation through pretrained models and deep reinforcement learning} (2022), \doi{10.5555/3600270.3601819}

\bibitem[{Lewis et~al.(2019)Lewis, Liu, Goyal, Ghazvininejad, Mohamed, Levy, Stoyanov, and Zettlemoyer}]{lewis-etal-2020-bart}
Lewis, M., Liu, Y., Goyal, N., Ghazvininejad, M., Mohamed, A., Levy, O., Stoyanov, V., Zettlemoyer, L.: {BART: Denoising Sequence-to-Sequence Pre-training for Natural Language Generation, Translation, and Comprehension}. arXiv (2019), \urlprefix\url{https://arxiv.org/abs/1910.13461}

\bibitem[{Li et~al.(2023{\natexlab{a}})Li, Mellou, Zhang, Pathuri, and Menache}]{li2023large}
Li, B., Mellou, K., Zhang, B., Pathuri, J., Menache, I.: {Large Language Models for Supply Chain Optimization}. arXiv (2023{\natexlab{a}}), \doi{10.48550/arXiv.2307.03875}

\bibitem[{Li et~al.(2024{\natexlab{a}})Li, Zhang, Sun, and Zou}]{10704489}
Li, B., Zhang, K., Sun, Y., Zou, J.: {Research on Travel Route Planning Optimization based on Large Language Model}. In: 2024 6th International Conference on Data-driven Optimization of Complex Systems (DOCS), pp. 352--357, IEEE, Hangzhou, China (2024{\natexlab{a}}), \doi{10.1109/DOCS63458.2024.10704489}

\bibitem[{Li et~al.(2023{\natexlab{b}})Li, Zhang, and Mak-Hau}]{li2023synthesizing}
Li, Q., Zhang, L., Mak-Hau, V.: {Synthesizing mixed-integer linear programming models from natural language descriptions}. arXiv (2023{\natexlab{b}}), \doi{10.48550/arXiv.2311.15271}

\bibitem[{Li et~al.(2023{\natexlab{c}})Li, Allal, Zi, Muennighoff, Kocetkov, Mou, Marone, Akiki, Li, Chim et~al.}]{li2023starcoder}
Li, R., Allal, L.B., Zi, Y., Muennighoff, N., Kocetkov, D., Mou, C., Marone, M., Akiki, C., Li, J., Chim, J., et~al.: {StarCoder: May the Source Be with You!} arXiv (2023{\natexlab{c}}), \doi{10.48550/arXiv.2305.06161}

\bibitem[{Li et~al.(2024{\natexlab{b}})Li, Kulkarni, Menache, Wu, and Li}]{li2024foundationmodelsmixedinteger}
Li, S., Kulkarni, J., Menache, I., Wu, C., Li, B.: {Towards Foundation Models for Mixed Integer Linear Programming}. arXiv (2024{\natexlab{b}}), \doi{10.48550/arXiv.2410.08288}

\bibitem[{Li et~al.(2024{\natexlab{c}})Li, Chu, Chen, Liu, Liu, Yu, Chen, Qian, Shi, and Yang}]{li2025graphteamfacilitatinglargelanguage}
Li, X., Chu, Q., Chen, Y., Liu, Y., Liu, Y., Yu, Z., Chen, W., Qian, C., Shi, C., Yang, C.: {Facilitating Large Language Model-based Graph Analysis via Multi-Agent Collaboration}. arXiv (2024{\natexlab{c}}), \doi{10.48550/arXiv.2410.18032}

\bibitem[{Lin(2004)}]{lin-2004-rouge}
Lin, C.Y.: {ROUGE: A Package for Automatic Evaluation of Summaries}. In: Text Summarization Branches Out, pp. 74--81, ACL, Barcelona, Spain (Jul 2004), \urlprefix\url{https://aclanthology.org/W04-1013}

\bibitem[{Lin et~al.(2017)Lin, Zhang, Zheng, and Zheng}]{00005792-201711270-00083}
Lin, H., Zhang, L., Zheng, R., Zheng, Y.: {The prevalence, metabolic risk and effects of lifestyle intervention for metabolically healthy obesity: a systematic review and meta-analysis: A PRISMA-compliant article}. Medicine \textbf{96}(47) (2017), \doi{10.1097/MD.0000000000008838}

\bibitem[{Lindauer et~al.(2022)Lindauer, Eggensperger, Feurer, Biedenkapp, Deng, Benjamins, Ruhkopf, Sass, and Hutter}]{JMLR:v23:21-0888}
Lindauer, M., Eggensperger, K., Feurer, M., Biedenkapp, A., Deng, D., Benjamins, C., Ruhkopf, T., Sass, R., Hutter, F.: {SMAC3: A Versatile Bayesian Optimization Package for Hyperparameter Optimization}. Journal of Machine Learning Research \textbf{23}(54), 1--9 (2022), \urlprefix\url{http://jmlr.org/papers/v23/21-0888.html}

\bibitem[{Ling et~al.(2017)Ling, Yogatama, Dyer, and Blunsom}]{ling-etal-2017-program}
Ling, W., Yogatama, D., Dyer, C., Blunsom, P.: {Program Induction by Rationale Generation: Learning to Solve and Explain Algebraic Word Problems}. In: Proceedings of the 55th Annual Meeting of the ACL (Volume 1: Long Papers), pp. 158--167, ACL, Vancouver, Canada (Jul 2017), \doi{10.18653/v1/P17-1015}

\bibitem[{Liu et~al.(2024{\natexlab{a}})Liu, Lin, Wang, Yao, Tong, Yuan, and Zhang}]{liu2024large}
Liu, F., Lin, X., Wang, Z., Yao, S., Tong, X., Yuan, M., Zhang, Q.: {Large Language Model for Multi-objective Evolutionary Optimization}. arXiv (2024{\natexlab{a}}), \doi{10.48550/arXiv.2310.12541}

\bibitem[{Liu et~al.(2024{\natexlab{b}})Liu, Tong, Yuan, Lin, Luo, Wang, Lu, and Zhang}]{liu2024evolution}
Liu, F., Tong, X., Yuan, M., Lin, X., Luo, F., Wang, Z., Lu, Z., Zhang, Q.: {Evolution of Heuristics: Towards Efficient Automatic Algorithm Design Using Large Language Model}. arXiv (2024{\natexlab{b}}), \doi{10.48550/arXiv.2401.02051}

\bibitem[{Liu et~al.(2024{\natexlab{c}})Liu, Tong, Yuan, Lin, Luo, Wang, Lu, and Zhang}]{10.5555/3692070.3693374}
Liu, F., Tong, X., Yuan, M., Lin, X., Luo, F., Wang, Z., Lu, Z., Zhang, Q.: {Evolution of Heuristics: Towards Efficient Automatic Algorithm Design Using Large Language Model}. In: Proceedings of the 41st International Conference on Machine Learning, ICML'24, JMLR.org, Vienna, Austria (2024{\natexlab{c}}), \urlprefix\url{https://openreview.net/forum?id=BwAkaxqiLB}

\bibitem[{Liu et~al.(2023{\natexlab{a}})Liu, Tong, Yuan, and Zhang}]{liu2023algorithm}
Liu, F., Tong, X., Yuan, M., Zhang, Q.: {Algorithm Evolution Using Large Language Model}. arXiv (2023{\natexlab{a}}), \doi{10.48550/arXiv.2311.15249}

\bibitem[{Liu et~al.(2024{\natexlab{d}})Liu, Yao, Guo, Yang, Zhao, Lin, Tong, Yuan, Lu, Wang, and Zhang}]{liu2024systematicsurveylargelanguage}
Liu, F., Yao, Y., Guo, P., Yang, Z., Zhao, Z., Lin, X., Tong, X., Yuan, M., Lu, Z., Wang, Z., Zhang, Q.: {A Systematic Survey on Large Language Models for Algorithm Design}. arXiv (2024{\natexlab{d}}), \doi{10.48550/arXiv.2410.14716}

\bibitem[{Liu et~al.(2024{\natexlab{e}})Liu, Zhang, Xie, Sun, Li, Lin, Wang, Lu, and Zhang}]{liu2024llm4adplatformalgorithmdesign}
Liu, F., Zhang, R., Xie, Z., Sun, R., Li, K., Lin, X., Wang, Z., Lu, Z., Zhang, Q.: {LLM4AD: A Platform for Algorithm Design with Large Language Model}. arXiv (2024{\natexlab{e}}), \doi{10.48550/arXiv.2412.17287}

\bibitem[{Liu et~al.(2024{\natexlab{f}})Liu, Chen, Qu, Tang, and Ong}]{liu2024large1}
Liu, S., Chen, C., Qu, X., Tang, K., Ong, Y.S.: {Large Language Models as Evolutionary Optimizers}. arXiv (2024{\natexlab{f}}), \doi{10.48550/arXiv.2310.19046}

\bibitem[{Liu et~al.(2023{\natexlab{b}})Liu, Wu, Liu, Wang, Wang, and Qu}]{LIU2023100520}
Liu, Y., Wu, F., Liu, Z., Wang, K., Wang, F., Qu, X.: {Can language models be used for real-world urban-delivery route optimization?} The Innovation \textbf{4}(6), 100520 (2023{\natexlab{b}}), ISSN 2666-6758, \doi{10.1016/j.xinn.2023.100520}

\bibitem[{Llama~Team(2024)}]{dubey2024llama3herdmodels}
Llama~Team, A.a.M.: {The Llama 3 Herd of Models}. arXiv (2024), \doi{10.48550/arXiv.2407.21783}

\bibitem[{Londe et~al.(2024)Londe, Pessoa, Andrade, and Resende}]{LONDE2024}
Londe, M.A., Pessoa, L.S., Andrade, C.E., Resende, M.G.: {Biased random-key genetic algorithms: A review}. European Journal of Operational Research \textbf{318}(2) (2024), ISSN 0377-2217, \doi{10.1016/j.ejor.2024.03.030}

\bibitem[{Long et~al.(2025)Long, Tan, Mao, Tang, Li, Zhao, and Kato}]{10829820}
Long, S., Tan, J., Mao, B., Tang, F., Li, Y., Zhao, M., Kato, N.: {A Survey on Intelligent Network Operations and Performance Optimization Based on Large Language Models}. IEEE Communications Surveys \& Tutorials pp. 1--1 (2025), \doi{10.1109/COMST.2025.3526606}

\bibitem[{Lopes~Silva et~al.(2018)Lopes~Silva, de~Souza, Freitas~Souza, and de~França~Filho}]{lopes-silva-2018-survey}
Lopes~Silva, M.A., de~Souza, S.R., Freitas~Souza, M.J., de~França~Filho, M.F.: {Hybrid metaheuristics and multi-agent systems for solving optimization problems: A review of frameworks and a comparative analysis}. Applied Soft Computing \textbf{71}, 433--459 (2018), ISSN 1568-4946, \doi{10.1016/j.asoc.2018.06.050}

\bibitem[{Lozhkov et~al.(2024)Lozhkov, Li, Allal, Cassano, Lamy-Poirier, Tazi, Tang, Pykhtar, Liu, Wei, and et~al.}]{lozhkov2024starcoder2stackv2}
Lozhkov, A., Li, R., Allal, L.B., Cassano, F., Lamy-Poirier, J., Tazi, N., Tang, A., Pykhtar, D., Liu, J., Wei, Y., et~al.: {StarCoder 2 and The Stack v2: The Next Generation}. arXiv (2024), \doi{10.48550/arXiv.2402.19173}

\bibitem[{Luo et~al.(2024)Luo, Song, Huang, Lian, Zhang, Jiang, and Xie}]{luo2024graphinstructempoweringlargelanguage}
Luo, Z., Song, X., Huang, H., Lian, J., Zhang, C., Jiang, J., Xie, X.: {GraphInstruct: Empowering Large Language Models with Graph Understanding and Reasoning Capability}. arXiv (2024), \urlprefix\url{https://arxiv.org/abs/2403.04483}

\bibitem[{López-Ibáñez et~al.(2016)López-Ibáñez, Dubois-Lacoste, Pérez~Cáceres, Birattari, and Stützle}]{lopex-ibanez-2016-irace}
López-Ibáñez, M., Dubois-Lacoste, J., Pérez~Cáceres, L., Birattari, M., Stützle, T.: {The irace package: Iterated racing for automatic algorithm configuration}. Operations Research Perspectives \textbf{3}, 43--58 (2016), ISSN 2214-7160, \doi{10.1016/j.orp.2016.09.002}

\bibitem[{Maddigan and Susnjak(2023)}]{10121440}
Maddigan, P., Susnjak, T.: {Chat2VIS: Generating Data Visualizations via Natural Language Using ChatGPT, Codex and GPT-3 Large Language Models}. IEEE Access \textbf{11}, 45181--45193 (2023), \doi{10.1109/ACCESS.2023.3274199}

\bibitem[{Manica et~al.(2023)Manica, Born, Cadow, Christofidellis, Dave, Clarke, Teukam, Giannone, Hoffman, Buchan, Chenthamarakshan, Donovan, Hsu, Zipoli, Schilter, Kishimoto, Hamada, Padhi, Wehden, McHugh, Khrabrov, Das, Takeda, and Smith}]{Manica2023}
Manica, M., Born, J., Cadow, J., Christofidellis, D., Dave, A., Clarke, D., Teukam, Y.G.N., Giannone, G., Hoffman, S.C., Buchan, M., Chenthamarakshan, V., Donovan, T., Hsu, H.H., Zipoli, F., Schilter, O., Kishimoto, A., Hamada, L., Padhi, I., Wehden, K., McHugh, L., Khrabrov, A., Das, P., Takeda, S., Smith, J.R.: Accelerating material design with the generative toolkit for scientific discovery. npj Computational Materials \textbf{9}(1), 69 (5 2023), ISSN 2057-3960, \doi{10.1038/s41524-023-01028-1}

\bibitem[{Manyika and Hsiao(2023)}]{manyika2023overview}
Manyika, J., Hsiao, S.: {An overview of Bard: an early experiment with generative AI}. AI. Google Static Documents (2023), \urlprefix\url{https://ai.google/static/documents/google-about-bard.pdf}, accessed: 2024-06-27

\bibitem[{Mao et~al.(2024)Mao, Zou, Sheng, Liu, Gao, Wang, and Li}]{mao2024identify}
Mao, J., Zou, D., Sheng, L., Liu, S., Gao, C., Wang, Y., Li, Y.: {Identify Critical Nodes in Complex Network with Large Language Models}. arXiv (2024), \doi{10.48550/arXiv.2403.03962}

\bibitem[{Martinek et~al.(2024)Martinek, Łukasik, and Gandomi}]{DBLP:conf/esann/MartinekLG24}
Martinek, A., Łukasik, S., Gandomi, A.H.: {Large Language Models as Tuning Agents of Metaheuristics}. In: 32nd European Symposium on Artificial Neural Networks, Computational Intelligence and Machine Learning (ESANN 2024), pp. 631--636, i6doc.com, Bruges, Belgium (2024), \doi{10.14428/ESANN/2024.ES2024-209}

\bibitem[{Martí et~al.(2024)Martí, Sevaux, and Sörensen}]{marti-2024-50-history}
Martí, R., Sevaux, M., Sörensen, K.: {Fifty years of metaheuristics}. European Journal of Operational Research \textbf{318}(2) (2024), ISSN 0377-2217, \doi{10.1016/j.ejor.2024.04.004}

\bibitem[{Meadows and Freitas(2024)}]{meadows2024survey}
Meadows, J., Freitas, A.: {A Survey in Mathematical Language Processing}. arXiv (2024), \doi{10.48550/arXiv.2205.15231}

\bibitem[{Michailidis et~al.(2024)Michailidis, Tsouros, and Guns}]{michailidis_et_al:LIPIcs.CP.2024.20}
Michailidis, K., Tsouros, D., Guns, T.: {Constraint Modelling with LLMs Using In-Context Learning}. In: 30th International Conference on Principles and Practice of Constraint Programming (CP 2024), Leibniz International Proceedings in Informatics (LIPIcs), vol. 307, pp. 20:1--20:27, Schloss Dagstuhl -- Leibniz-Zentrum f{\"u}r Informatik, Dagstuhl, Germany (2024), \doi{10.4230/LIPIcs.CP.2024.20}

\bibitem[{Mo et~al.(2024)Mo, Liu, Shen, Xu, Xu, Su, and Zhang}]{10675146}
Mo, K., Liu, W., Shen, F., Xu, X., Xu, L., Su, X., Zhang, Y.: {Precision Kinematic Path Optimization for High-DoF Robotic Manipulators Utilizing Advanced Natural Language Processing Models}. In: 2024 5th International Conference on Electronic Communication and Artificial Intelligence (ICECAI), pp. 649--654, IEEE, Shenzhen, China (2024), \doi{10.1109/ICECAI62591.2024.10675146}

\bibitem[{Moher et~al.(1999)Moher, Cook, Eastwood, Olkinm, Rennie, and Stroup}]{MOHER19991896}
Moher, D., Cook, D.J., Eastwood, S., Olkinm, I., Rennie, D., Stroup, D.F.: {Improving The Quality of Reports of Meta-analyses of Randomised Controlled Trials: The QUOROM Statement}. The Lancet \textbf{354}(9193), 1896--1900 (1999), ISSN 0140-6736, \doi{10.1016/S0140-6736(99)04149-5}

\bibitem[{Moher et~al.(2009)Moher, Liberati, Tetzlaff, Altman, and Group}]{moher2009preferred}
Moher, D., Liberati, A., Tetzlaff, J., Altman, D.G., Group, T.P.: {Preferred Reporting Items for Systematic Reviews and Meta-Analyses: The PRISMA Statement}. PLOS Medicine \textbf{6}(7), 1--6 (07 2009), \doi{10.1371/journal.pmed.1000097}

\bibitem[{Moser et~al.(2022)Moser, Musliu, Schaerf, and Winter}]{moser2022exact}
Moser, M., Musliu, N., Schaerf, A., Winter, F.: {Exact and metaheuristic approaches for unrelated parallel machine scheduling}. Journal of Scheduling \textbf{25}(5), 507--534 (2022), \doi{10.1007/s10951-021-00714-6}

\bibitem[{Mostajabdaveh et~al.(2024{\natexlab{a}})Mostajabdaveh, Yu, Dash, Ramamonjison, Byusa, Carenini, Zhou, and Zhang}]{mostajabdaveh2024evaluatingllmreasoningoperations}
Mostajabdaveh, M., Yu, T.T., Dash, S.C.B., Ramamonjison, R., Byusa, J.S., Carenini, G., Zhou, Z., Zhang, Y.: {Evaluating LLM Reasoning in the Operations Research Domain with ORQA}. arXiv (2024{\natexlab{a}}), \doi{10.48550/arXiv.2412.17874}

\bibitem[{Mostajabdaveh et~al.(2024{\natexlab{b}})Mostajabdaveh, Yu, Ramamonjison, Carenini, Zhou, and Zhang}]{Mostajabdaveh04112024}
Mostajabdaveh, M., Yu, T.T., Ramamonjison, R., Carenini, G., Zhou, Z., Zhang, Y.: {Optimization modeling and verification from problem specifications using a multi-agent multi-stage LLM framework}. INFOR: Information Systems and Operational Research \textbf{62}(4), 599--617 (2024{\natexlab{b}}), \doi{10.1080/03155986.2024.2381306}

\bibitem[{Nana~Teukam et~al.(2024)Nana~Teukam, Zipoli, Laino, Criscuolo, Grisoni, and Manica}]{teukam2024integrating}
Nana~Teukam, Y.G., Zipoli, F., Laino, T., Criscuolo, E., Grisoni, F., Manica, M.: {Integrating Genetic Algorithms and Language Models for Enhanced Enzyme Design}. Briefings in Bioinformatics \textbf{26}(1), bbae675 (Nov 2024), \doi{10.1093/bib/bbae675}

\bibitem[{Nassen et~al.(2023)Nassen, Vandebosch, Poels, and Karsay}]{NASSEN2023101980}
Nassen, L.M., Vandebosch, H., Poels, K., Karsay, K.: {Opt-out, abstain, unplug. A systematic review of the voluntary digital disconnection literature}. Telematics and Informatics \textbf{81}, 101980 (2023), ISSN 0736-5853, \doi{https://doi.org/10.1016/j.tele.2023.101980}

\bibitem[{Nethercote et~al.(2007)Nethercote, Stuckey, Becket, Brand, Duck, and Tack}]{nethercote_minizinc_2007}
Nethercote, N., Stuckey, P.J., Becket, R., Brand, S., Duck, G.J., Tack, G.: {MiniZinc: Towards a Standard CP Modelling Language}. In: Principles and Practice of Constraint Programming -- CP 2007, pp. 529--543, Springer Berlin Heidelberg, Berlin, Heidelberg (2007), ISBN 978-3-540-74970-7, \doi{10.1007/978-3-540-74970-7_38}

\bibitem[{Ning et~al.(2023)Ning, Liu, Qin, Xiao, Xue, Huang, Liu, Chen, and Wu}]{ning2023novel}
Ning, Y., Liu, J., Qin, L., Xiao, T., Xue, S., Huang, Z., Liu, Q., Chen, E., Wu, J.: {A Novel Approach for Auto-Formulation of Optimization Problems}. arXiv (2023), \doi{10.48550/arXiv.2302.04643}

\bibitem[{Obata et~al.(2025)Obata, Aoki, Horii, Taniguchi, and Nagai}]{10803039}
Obata, K., Aoki, T., Horii, T., Taniguchi, T., Nagai, T.: {LiP-LLM: Integrating Linear Programming and Dependency Graph With Large Language Models for Multi-Robot Task Planning}. IEEE Robotics and Automation Letters \textbf{10}(2), 1122--1129 (2025), \doi{10.1109/LRA.2024.3518105}

\bibitem[{{Object Management Group}(2011)}]{omg2011bpmn}
{Object Management Group}: {Business Process Model and Notation (BPMN), Version 2.0} (January 2011), \urlprefix\url{https://www.omg.org/spec/BPMN/2.0/PDF}, retrieved from the Object Management Group website

\bibitem[{Ochoa et~al.(2021)Ochoa, Malan, and Blum}]{OCHOA2021107492}
Ochoa, G., Malan, K.M., Blum, C.: {Search trajectory networks: A tool for analysing and visualising the behaviour of metaheuristics}. Applied Soft Computing \textbf{109}, 107492 (2021), ISSN 1568-4946, \doi{10.1016/j.asoc.2021.107492}

\bibitem[{OpenAI(2024)}]{openai2024gpt4}
OpenAI: {GPT-4 Technical Report}. arXiv (2024), \doi{10.48550/arXiv.2303.08774}

\bibitem[{{OpenAI Team}(2024)}]{openai_reproducible_outputs}
{OpenAI Team}: Reproducible outputs (2024), \urlprefix\url{https://platform.openai.com/docs/advanced-usage/reproducible-outputs}, accessed: 2024-08-26

\bibitem[{Ouyang et~al.(2022)Ouyang, Wu, Jiang, Almeida, Wainwright, Mishkin, Zhang, Agarwal, Slama, Ray et~al.}]{ouyang2022training}
Ouyang, L., Wu, J., Jiang, X., Almeida, D., Wainwright, C., Mishkin, P., Zhang, C., Agarwal, S., Slama, K., Ray, A., et~al.: Training language models to follow instructions with human feedback. Advances in Neural Information Processing Systems \textbf{35}, 27730--27744 (2022), \urlprefix\url{https://proceedings.neurips.cc/paper_files/paper/2022/file/b1efde53be364a73914f58805a001731-Paper-Conference.pdf}

\bibitem[{Page et~al.(2021{\natexlab{a}})Page, McKenzie, Bossuyt, Boutron, Hoffmann, Mulrow, Shamseer, Tetzlaff, Akl, Brennan, Chou, Glanville, Grimshaw, Hr{\'o}bjartsson, Lalu, Li, Loder, Mayo-Wilson, McDonald, McGuinness, Stewart, Thomas, Tricco, Welch, Whiting, and Moher}]{Pagen71}
Page, M.J., McKenzie, J.E., Bossuyt, P.M., Boutron, I., Hoffmann, T.C., Mulrow, C.D., Shamseer, L., Tetzlaff, J.M., Akl, E.A., Brennan, S.E., Chou, R., Glanville, J., Grimshaw, J.M., Hr{\'o}bjartsson, A., Lalu, M.M., Li, T., Loder, E.W., Mayo-Wilson, E., McDonald, S., McGuinness, L.A., Stewart, L.A., Thomas, J., Tricco, A.C., Welch, V.A., Whiting, P., Moher, D.: {The PRISMA 2020 Statement: An Updated Guideline For Reporting Systematic Reviews}. BMJ \textbf{372} (2021{\natexlab{a}}), \doi{10.1136/bmj.n71}

\bibitem[{Page et~al.(2021{\natexlab{b}})Page, Moher, Bossuyt, Boutron, Hoffmann, Mulrow, Shamseer, Tetzlaff, Akl, Brennan, Chou, Glanville, Grimshaw, Hr{\'o}bjartsson, Lalu, Li, Loder, Mayo-Wilson, McDonald, McGuinness, Stewart, Thomas, Tricco, Welch, Whiting, and McKenzie}]{Pagen160}
Page, M.J., Moher, D., Bossuyt, P.M., Boutron, I., Hoffmann, T.C., Mulrow, C.D., Shamseer, L., Tetzlaff, J.M., Akl, E.A., Brennan, S.E., Chou, R., Glanville, J., Grimshaw, J.M., Hr{\'o}bjartsson, A., Lalu, M.M., Li, T., Loder, E.W., Mayo-Wilson, E., McDonald, S., McGuinness, L.A., Stewart, L.A., Thomas, J., Tricco, A.C., Welch, V.A., Whiting, P., McKenzie, J.E.: {PRISMA 2020 Explanation and Elaboration: Updated Guidance And Exemplars For Reporting Systematic Reviews}. BMJ \textbf{372} (2021{\natexlab{b}}), \doi{10.1136/bmj.n160}

\bibitem[{Pagnozzi and Stützle(2021)}]{PAGNOZZI2021100180}
Pagnozzi, F., Stützle, T.: {Automatic design of hybrid stochastic local search algorithms for permutation flowshop problems with additional constraints}. Operations Research Perspectives \textbf{8}, 100180 (2021), ISSN 2214-7160, \doi{https://doi.org/10.1016/j.orp.2021.100180}

\bibitem[{Pallagani et~al.(2025)Pallagani, Muppasani, Roy, Fabiano, Loreggia, Murugesan, Srivastava, Rossi, Horesh, and Sheth}]{10.1609/icaps.v34i1.31503}
Pallagani, V., Muppasani, B.C., Roy, K., Fabiano, F., Loreggia, A., Murugesan, K., Srivastava, B., Rossi, F., Horesh, L., Sheth, A.: {On the prospects of incorporating large language models (LLMs) in automated planning and scheduling (APS)}. In: Proceedings of the Thirty-Fourth International Conference on Automated Planning and Scheduling, ICAPS '24, AAAI Press, Alberta, Canada (2025), ISBN 1-57735-889-9, \doi{10.1609/icaps.v34i1.31503}

\bibitem[{Pan et~al.(2024)Pan, Xing, Diao, Sun, Liu, Shum, Zhang, Pi, and Zhang}]{pan-etal-2024-plum}
Pan, R., Xing, S., Diao, S., Sun, W., Liu, X., Shum, K., Zhang, J., Pi, R., Zhang, T.: Plum: Prompt learning using metaheuristics. In: Findings of the ACL: ACL 2024, pp. 2177--2197, ACL, Bangkok, Thailand (Aug 2024), \doi{10.18653/v1/2024.findings-acl.129}

\bibitem[{Parejo et~al.(2012)Parejo, Ruiz-Cort\'{e}s, Lozano, and Fernandez}]{parejo-2012-survey}
Parejo, J.A., Ruiz-Cort\'{e}s, A., Lozano, S., Fernandez, P.: {Metaheuristic optimization frameworks: a survey and benchmarking}. Soft Computing \textbf{16}(3), 527–561 (mar 2012), ISSN 1432-7643, \doi{10.1007/s00500-011-0754-8}

\bibitem[{Perron and Furnon(2024)}]{ortools}
Perron, L., Furnon, V.: {OR-Tools} (2024), \urlprefix\url{https://developers.google.com/optimization/}, software

\bibitem[{Peters et~al.(2018)Peters, Neumann, Iyyer, Gardner, Clark, Lee, and Zettlemoyer}]{peters-etal-2018-deep}
Peters, M.E., Neumann, M., Iyyer, M., Gardner, M., Clark, C., Lee, K., Zettlemoyer, L.: {Deep Contextualized Word Representations}. In: Proceedings of the 2018 Conference of the North {A}merican Chapter of the ACL: Human Language Technologies, Volume 1 (Long Papers), pp. 2227--2237, ACL, New Orleans, Louisiana (Jun 2018), \doi{10.18653/v1/N18-1202}

\bibitem[{Pluhacek et~al.(2023{\natexlab{a}})Pluhacek, Kazikova, Kadavy, Viktorin, and Senkerik}]{10.1145/3583133.3596401}
Pluhacek, M., Kazikova, A., Kadavy, T., Viktorin, A., Senkerik, R.: Leveraging large language models for the generation of novel metaheuristic optimization algorithms. In: Proceedings of the Companion Conference on Genetic and Evolutionary Computation, pp. 1812--1820, GECCO '23 Companion, ACM, New York, NY, USA (2023{\natexlab{a}}), ISBN 9798400701207, \doi{10.1145/3583133.3596401}

\bibitem[{Pluhacek et~al.(2023{\natexlab{b}})Pluhacek, Kazikova, Viktorin, Kadavy, and Senkerik}]{10394233}
Pluhacek, M., Kazikova, A., Viktorin, A., Kadavy, T., Senkerik, R.: {Investigating the Potential of AI-Driven Innovations for Enhancing Differential Evolution in Optimization Tasks}. In: 2023 IEEE International Conference on Systems, Man, and Cybernetics (SMC), pp. 1070--1075, IEEE, Honolulu, Oahu, HI, USA (2023{\natexlab{b}}), \doi{10.1109/SMC53992.2023.10394233}

\bibitem[{Pluhacek et~al.(2024)Pluhacek, Kovac, Viktorin, Janku, Kadavy, and Senkerik}]{10.1145/3638530.3664181}
Pluhacek, M., Kovac, J., Viktorin, A., Janku, P., Kadavy, T., Senkerik, R.: {Using LLM for Automatic Evolvement of Metaheuristics from Swarm Algorithm SOMA}. In: Proceedings of the Genetic and Evolutionary Computation Conference Companion, p. 2018–2022, GECCO '24 Companion, ACM, New York, NY, USA (2024), ISBN 9798400704956, \doi{10.1145/3638530.3664181}

\bibitem[{Pop et~al.(2024)Pop, Cosma, Sabo, and Pop~Sitar}]{POP2024819}
Pop, P.C., Cosma, O., Sabo, C., Pop~Sitar, C.: {A comprehensive survey on the generalized traveling salesman problem}. European Journal of Operational Research \textbf{314}(3), 819--835 (2024), ISSN 0377-2217, \doi{10.1016/j.ejor.2023.07.022}

\bibitem[{Qwen~Team(2023)}]{bai2023qwentechnicalreport}
Qwen~Team, A.G.: {Qwen Technical Report}. arXiv (2023), \doi{10.48550/arXiv.2309.16609}

\bibitem[{Qwen~Team(2024)}]{yang2024qwen2technicalreport}
Qwen~Team, A.G.: {Qwen2 Technical Report}. arXiv (2024), \doi{10.48550/arXiv.2407.10671}

\bibitem[{Rafailov et~al.(2024)Rafailov, Sharma, Mitchell, Ermon, Manning, and Finn}]{rafailov2024directpreferenceoptimizationlanguage}
Rafailov, R., Sharma, A., Mitchell, E., Ermon, S., Manning, C.D., Finn, C.: {Direct Preference Optimization: Your Language Model is Secretly a Reward Model}. arXiv (2024), \doi{10.48550/arXiv.2305.18290)}

\bibitem[{Raffel et~al.(2023)Raffel, Shazeer, Roberts, Lee, Narang, Matena, Zhou, Li, Liu et~al.}]{raffel2023exploringlimitstransferlearning}
Raffel, C., Shazeer, N., Roberts, A., Lee, K., Narang, S., Matena, M., Zhou, Y., Li, W., Liu, P.J., et~al.: {Exploring the Limits of Transfer Learning with a Unified Text-to-Text Transformer}. arXiv (2023), \doi{10.48550/arXiv.1910.10683}

\bibitem[{Ramamonjison et~al.(2022)Ramamonjison, Li, Yu, He, Rengan, Banitalebi-dehkordi, Zhou, and Zhang}]{ramamonjison-etal-2022-augmenting}
Ramamonjison, R., Li, H., Yu, T., He, S., Rengan, V., Banitalebi-dehkordi, A., Zhou, Z., Zhang, Y.: {Augmenting Operations Research with Auto-Formulation of Optimization Models From Problem Descriptions}. In: Proceedings of the 2022 Conference on Empirical Methods in Natural Language Processing: Industry Track, pp. 29--62, ACL, Abu Dhabi, UAE (Dec 2022), \doi{10.18653/v1/2022.emnlp-industry.4}

\bibitem[{Ramamonjison et~al.(2023)Ramamonjison, Yu, Li, Li, Carenini, Ghaddar, He, Mostajabdaveh, Banitalebi-Dehkordi, Zhou, and Zhang}]{ramamonjison2023nl4opt}
Ramamonjison, R., Yu, T.T., Li, R., Li, H., Carenini, G., Ghaddar, B., He, S., Mostajabdaveh, M., Banitalebi-Dehkordi, A., Zhou, Z., Zhang, Y.: {NL4Opt Competition: Formulating Optimization Problems Based on Their Natural Language Descriptions}. arXiv (2023), \doi{10.48550/arXiv.2303.08233}

\bibitem[{R\'{e}gin et~al.(2024)R\'{e}gin, De~Maria, and Bonlarron}]{regin_et_al:LIPIcs.CP.2024.25}
R\'{e}gin, F., De~Maria, E., Bonlarron, A.: {Combining Constraint Programming Reasoning with Large Language Model Predictions}. In: 30th International Conference on Principles and Practice of Constraint Programming (CP 2024), Leibniz International Proceedings in Informatics (LIPIcs), vol. 307, pp. 25:1--25:18, Schloss Dagstuhl -- Leibniz-Zentrum f{\"u}r Informatik, Dagstuhl, Germany (2024), \doi{10.4230/LIPIcs.CP.2024.25}

\bibitem[{Reinhart and Statt(2024)}]{reinhart-2024}
Reinhart, W.F., Statt, A.: {Large language models design sequence-defined macromolecules via evolutionary optimization}. NPJ Computational Materials \textbf{10}(1), 262 (2024), \doi{10.1038/s41524-024-01449-6}

\bibitem[{Romanko et~al.(2023)Romanko, Narayan, and Kwon}]{RePEc:spr:snopef:v:4:y:2023:i:4:d:10.1007_s43069-023-00277-6}
Romanko, O., Narayan, A., Kwon, R.H.: {ChatGPT-Based Investment Portfolio Selection}. SN Operations Research Forum \textbf{4}(4), 1--27 (December 2023), \doi{10.1007/s43069-023-00277-6}

\bibitem[{Romera-Paredes et~al.(2024)Romera-Paredes, Barekatain, Novikov, Balog, Kumar, Dupont, Ruiz, Ellenberg, Wang, Fawzi, Kohli, and Fawzi}]{Romera-Paredes2024}
Romera-Paredes, B., Barekatain, M., Novikov, A., Balog, M., Kumar, M.P., Dupont, E., Ruiz, F.J.R., Ellenberg, J.S., Wang, P., Fawzi, O., Kohli, P., Fawzi, A.: {Mathematical discoveries from program search with large language models}. Nature \textbf{625}(7995), 468--475 (1 2024), ISSN 1476-4687, \doi{10.1038/s41586-023-06924-6}

\bibitem[{Ross et~al.(2016)Ross, Stevenson, Lau, and Murray}]{Ross2016}
Ross, J., Stevenson, F., Lau, R., Murray, E.: {Factors that influence the implementation of e-health: a systematic review of systematic reviews (an update)}. Implementation Science \textbf{11}(1), 146 (Oct 2016), ISSN 1748-5908, \doi{10.1186/s13012-016-0510-7}

\bibitem[{Rossi et~al.(2006)Rossi, Van~Beek, and Walsh}]{rossi-2006-handbook-cp}
Rossi, F., Van~Beek, P., Walsh, T.: Handbook of constraint programming. Elsevier, Amsterdam (2006), ISBN 9780444527264, \urlprefix\url{https://www.dcs.gla.ac.uk/~pat/cpM/papers/CP_Handbook-20060315-final.pdf}

\bibitem[{Saka et~al.(2024)Saka, Taiwo, Saka, Salami, Ajayi, Akande, and Kazemi}]{SAKA2024100300}
Saka, A., Taiwo, R., Saka, N., Salami, B.A., Ajayi, S., Akande, K., Kazemi, H.: {GPT models in construction industry: Opportunities, limitations, and a use case validation}. Developments in the Built Environment \textbf{17}, 100300 (2024), ISSN 2666-1659, \doi{https://doi.org/10.1016/j.dibe.2023.100300}

\bibitem[{Sartori et~al.(2025)Sartori, Blum, Bistaffa, and Rodr{\'\i}guez~Corominas}]{10818476}
Sartori, C.C., Blum, C., Bistaffa, F., Rodr{\'\i}guez~Corominas, G.: {Metaheuristics and Large Language Models Join Forces: Toward an Integrated Optimization Approach}. IEEE Access \textbf{13}, 2058--2079 (2025), \doi{10.1109/ACCESS.2024.3524176}

\bibitem[{Schäfer et~al.(2024)Schäfer, Nadi, Eghbali, and Tip}]{10329992}
Schäfer, M., Nadi, S., Eghbali, A., Tip, F.: {An Empirical Evaluation of Using Large Language Models for Automated Unit Test Generation}. IEEE Transactions on Software Engineering \textbf{50}(1), 85--105 (2024), \doi{10.1109/TSE.2023.3334955}

\bibitem[{Shao et~al.(2024)Shao, Wang, Zhu, Xu, Song, Bi, Zhang, Zhang, Li, Wu, and Guo}]{shao2024deepseekmathpushinglimitsmathematical}
Shao, Z., Wang, P., Zhu, Q., Xu, R., Song, J., Bi, X., Zhang, H., Zhang, M., Li, Y.K., Wu, Y., Guo, D.: {DeepSeekMath: Pushing the Limits of Mathematical Reasoning in Open Language Models}. arXiv (2024), \doi{10.48550/arXiv.2402.03300}

\bibitem[{Shaw(1998)}]{shaw-1998-lns}
Shaw, P.: {Principles and Practice of Constraint Programming --- CP98}. In: {4th International Conference, CP98, Pisa, Italy, October 26-30, 1998: Proceedings}, Lecture Notes in Computer Science, vol. 1520, Springer, Berlin (1998), ISBN 978-3540659290, \doi{10.1007/3-540-49481-2}

\bibitem[{Singla et~al.(2023)Singla, Singh, and Kukreja}]{singla2023biobjective}
Singla, A., Singh, A., Kukreja, K.: {A bi-objective $\epsilon$-constrained framework for quality-cost optimization in language model ensembles}. arXiv (2023), \doi{10.48550/arXiv.2312.16119}

\bibitem[{Soprano et~al.(2024)Soprano, Roitero, {La Barbera}, Ceolin, Spina, Demartini, and Mizzaro}]{SOPRANO2024103672}
Soprano, M., Roitero, K., {La Barbera}, D., Ceolin, D., Spina, D., Demartini, G., Mizzaro, S.: {Cognitive Biases in Fact-Checking and Their Countermeasures: A Review}. Information Processing \& Management \textbf{61}(3), 103672 (2024), ISSN 0306-4573, \doi{10.1016/j.ipm.2024.103672}

\bibitem[{S{\"o}rensen et~al.(2018)S{\"o}rensen, Sevaux, and Glover}]{Sörensen2018}
S{\"o}rensen, K., Sevaux, M., Glover, F.: {A History of Metaheuristics}, pp. 791--808. Springer International Publishing, Cham (2018), ISBN 978-3-319-07124-4, \doi{10.1007/978-3-319-07124-4_4}

\bibitem[{Srivastava et~al.(2022)Srivastava, Rastogi, Rao, Shoeb, Abid, Fisch, Brown, Santoro, Gupta, Garriga-Alonso et~al.}]{srivastava2023beyond}
Srivastava, A., Rastogi, A., Rao, A., Shoeb, A.A.M., Abid, A., Fisch, A., Brown, A.R., Santoro, A., Gupta, A., Garriga-Alonso, A., et~al.: {Beyond the Imitation Game: Quantifying and Extrapolating the Capabilities of Language Models}. arXiv (2022), \doi{10.48550/arXiv.2206.04615}

\bibitem[{Srivastava and Pallagani(2024)}]{srivastava2024casedevelopingfoundationmodel}
Srivastava, B., Pallagani, V.: {The Case for Developing a Foundation Model for Planning-like Tasks from Scratch} (2024), \doi{10.48550/arXiv.2404.04540}

\bibitem[{van Stein and Bäck(2024)}]{10752628}
van Stein, N., Bäck, T.: {LLaMEA: A Large Language Model Evolutionary Algorithm for Automatically Generating Metaheuristics}. IEEE Transactions on Evolutionary Computation pp. 1--1 (2024), \doi{10.1109/TEVC.2024.3497793}

\bibitem[{van Stein et~al.(2024)van Stein, Vermetten, and B{\"a}ck}]{vanstein2024intheloophyperparameteroptimizationllmbased}
van Stein, N., Vermetten, D., B{\"a}ck, T.: {In-the-loop Hyper-Parameter Optimization for LLM-Based Automated Design of Heuristics}. arXiv (2024), \doi{10.48550/arXiv.2410.16309}

\bibitem[{Steiner et~al.(2024)Steiner, Pferschy, and Schaerf}]{Steiner2024}
Steiner, E., Pferschy, U., Schaerf, A.: {Curriculum-based university course timetabling considering individual course of studies}. {Central European Journal of Operations Research} \textbf{32} (6 2024), ISSN 1613-9178, \doi{10.1007/s10100-024-00923-2}

\bibitem[{Sui et~al.(2024)Sui, Ding, Huang, Yu, Liu, Xia, Ding, Xu, Zhang, Yu, and Bu}]{sui-2024}
Sui, J., Ding, S., Huang, X., Yu, Y., Liu, R., Xia, B., Ding, Z., Xu, L., Zhang, H., Yu, C., Bu, D.: {A survey on deep learning-based algorithms for the traveling salesman problem}. Frontiers of Computer Science \textbf{19}(6), 196322 (2024), \doi{10.1007/s11704-024-40490-y}

\bibitem[{Sun et~al.(2024)Sun, Ye, Zhang, Huang, Zhang, Wei, and Cai}]{sun2024autosatautomaticallyoptimizesat}
Sun, Y., Ye, F., Zhang, X., Huang, S., Zhang, B., Wei, K., Cai, S.: {AutoSAT: Automatically Optimize SAT Solvers via Large Language Models}. arXiv (2024), \doi{10.48550/arXiv.2402.10705}

\bibitem[{Suzgun et~al.(2022)Suzgun, Scales, Schärli, Gehrmann, Tay, Chung, Chowdhery, Le, Chi, Zhou, and Wei}]{suzgun2022challengingbigbenchtaskschainofthought}
Suzgun, M., Scales, N., Schärli, N., Gehrmann, S., Tay, Y., Chung, H.W., Chowdhery, A., Le, Q.V., Chi, E.H., Zhou, D., Wei, J.: {Challenging BIG-Bench Tasks and Whether Chain-of-Thought Can Solve Them}. arXiv (2022), \doi{10.48550/arXiv.2210.09261}

\bibitem[{Swan et~al.(2015)Swan, Adriaensen, Bishr, Burke, Clark, De~Causmaecker, Durillo, Hammond, Hart, Johnson, Kocsis, Kovitz, Krawiec, Martin, Merelo, Minku, Özcan, Pappa, Pesch, Garcia-Sánchez, Schaerf, Sim, Smith, Stützle, Vo, Wagner, and Yao}]{swan-2015-research-agenda}
Swan, J., Adriaensen, S., Bishr, M., Burke, E.K., Clark, J.A., De~Causmaecker, P., Durillo, J., Hammond, K., Hart, E., Johnson, C.G., Kocsis, Z.A., Kovitz, B., Krawiec, K., Martin, S., Merelo, J.J., Minku, L.L., Özcan, E., Pappa, G.L., Pesch, E., Garcia-Sánchez, P., Schaerf, A., Sim, K., Smith, J., Stützle, T., Vo, S., Wagner, S., Yao, X.: {A Research Agenda for Metaheuristic Standardization}. In: {Proceedings of the XI Metaheuristics International Conference (MIC 2015)}, pp. 1--3, University of Nottingham, Agadir, Morocco (June 2015), \doi{10.1007/978-3-031-62912-9}

\bibitem[{Swan et~al.(2022)Swan, Adriaensen, Brownlee, Hammond, Johnson, Kheiri, Krawiec, Merelo, Minku, Özcan, Pappa, García-Sánchez, Sörensen, Voß, Wagner, and White}]{swan-2022-mil}
Swan, J., Adriaensen, S., Brownlee, A.E., Hammond, K., Johnson, C.G., Kheiri, A., Krawiec, F., Merelo, J., Minku, L.L., Özcan, E., Pappa, G.L., García-Sánchez, P., Sörensen, K., Voß, S., Wagner, M., White, D.R.: {Metaheuristics “In the Large”}. European Journal of Operational Research \textbf{297}(2), 393--406 (2022), ISSN 0377-2217, \doi{10.1016/j.ejor.2021.05.042}

\bibitem[{Swan et~al.(2019)Swan, Adriænsen, Barwell, Hammond, and White}]{swan-2019-open-closed-princ}
Swan, J., Adriænsen, S., Barwell, A.D., Hammond, K., White, D.R.: {Extending the “Open-Closed Principle” to Automated Algorithm Configuration}. Evolutionary Computation \textbf{27}(1), 173--193 (03 2019), ISSN 1063-6560, \doi{10.1162/evco_a_00245}

\bibitem[{Team(2024{\natexlab{a}})}]{geminiteam2024gemini}
Team, G.: {Gemini 1.5: Unlocking multimodal understanding across millions of tokens of context}. arXiv (2024{\natexlab{a}}), \doi{10.48550/arXiv.2403.05530}

\bibitem[{Team(2024{\natexlab{b}})}]{geminiteam2024geminifamilyhighlycapable}
Team, G.: {Gemini: A Family of Highly Capable Multimodal Models}. arXiv (2024{\natexlab{b}}), \doi{10.48550/arXiv.2312.11805}

\bibitem[{Team(2024{\natexlab{c}})}]{gemmateam2024gemma2improvingopen}
Team, G.: {Gemma 2: Improving Open Language Models at a Practical Size}. arXiv (2024{\natexlab{c}}), \doi{10.48550/arXiv.2408.00118}

\bibitem[{Team(2024{\natexlab{d}})}]{roziere2024codellamaopenfoundation}
Team, M.A.: {Code Llama: Open Foundation Models for Code}. arXiv (2024{\natexlab{d}}), \doi{10.48550/arXiv.2308.12950}

\bibitem[{Team(2024{\natexlab{e}})}]{yi2024open}
Team, Y.: {Yi: Open Foundation Models by 01.AI}. arXiv (2024{\natexlab{e}}), \doi{10.48550/arXiv.2403.04652}

\bibitem[{Thieu and Mirjalili(2023)}]{van2023mealpy}
Thieu, N.V., Mirjalili, S.: {MEALPY: An Open-Source Library for Latest Meta-Heuristic Algorithms in Python}. Journal of Systems Architecture \textbf{139}, 102871 (2023), \doi{10.1016/j.sysarc.2023.102871}

\bibitem[{Touvron et~al.(2023)Touvron, Lavril, Izacard, Martinet, Lachaux, Lacroix, Rozière, Goyal, Hambro, Azhar, Rodriguez, Joulin, Grave, and Lample}]{touvron2023llama}
Touvron, H., Lavril, T., Izacard, G., Martinet, X., Lachaux, M.A., Lacroix, T., Rozière, B., Goyal, N., Hambro, E., Azhar, F., Rodriguez, A., Joulin, A., Grave, E., Lample, G.: {LLaMA: Open and Efficient Foundation Language Models}. arXiv (2023), \doi{10.48550/arXiv.2302.13971}

\bibitem[{Tran and Hy(2024)}]{10628050}
Tran, T.V.T., Hy, T.S.: {Protein Design by Directed Evolution Guided by Large Language Models}. IEEE Transactions on Evolutionary Computation pp. 1--1 (2024), \doi{10.1109/TEVC.2024.3439690}

\bibitem[{Tsouros et~al.(2023)Tsouros, Verhaeghe, Kadıoğlu, and Guns}]{tsouros2023holy}
Tsouros, D., Verhaeghe, H., Kadıoğlu, S., Guns, T.: {Holy Grail 2.0: From Natural Language to Constraint Models}. arXiv (2023), \doi{10.48550/arXiv.2308.01589}

\bibitem[{Tunstall et~al.(2023)Tunstall, Beeching, Lambert, Rajani, Rasul, Belkada, Huang, von Werra, Fourrier, Habib, Sarrazin, Sanseviero, Rush, and Wolf}]{tunstall2023zephyrdirectdistillationlm}
Tunstall, L., Beeching, E., Lambert, N., Rajani, N., Rasul, K., Belkada, Y., Huang, S., von Werra, L., Fourrier, C., Habib, N., Sarrazin, N., Sanseviero, O., Rush, A.M., Wolf, T.: {Zephyr: Direct Distillation of LM Alignment}. arXiv (2023), \doi{10.48550/arXiv.2310.16944}

\bibitem[{Tupayachi et~al.(2024)Tupayachi, Xu, Omitaomu, Camur, Sharmin, and Li}]{smartcities7050094}
Tupayachi, J., Xu, H., Omitaomu, O.A., Camur, M.C., Sharmin, A., Li, X.: {Towards Next-Generation Urban Decision Support Systems through AI-Powered Construction of Scientific Ontology Using Large Language Models---A Case in Optimizing Intermodal Freight Transportation}. Smart Cities \textbf{7}(5), 2392--2421 (2024), \doi{10.3390/smartcities7050094}

\bibitem[{Urdaneta-Ponte et~al.(2021)Urdaneta-Ponte, Mendez-Zorrilla, and Oleagordia-Ruiz}]{electronics10141611}
Urdaneta-Ponte, M.C., Mendez-Zorrilla, A., Oleagordia-Ruiz, I.: {Recommendation Systems for Education: Systematic Review}. Electronics \textbf{10}(14) (2021), ISSN 2079-9292, \doi{10.3390/electronics10141611}

\bibitem[{Ustyugov(2024)}]{10720437}
Ustyugov, V.: {On Different Methods For Automated MILP Solver Configuration}. In: 2024 20th International Asian School-Seminar on Optimization Problems of Complex Systems (OPCS), pp. 24--27, IEEE, Issyk-Kul Lake, Kyrgyzstan (2024), \doi{10.1109/OPCS63516.2024.10720437}

\bibitem[{Vass et~al.(2022)Vass, Lackner, Mrkvicka, Musliu, and Winter}]{Vass2022}
Vass, J., Lackner, M.L., Mrkvicka, C., Musliu, N., Winter, F.: {Exact and meta-heuristic approaches for the production leveling problem}. Journal of Scheduling \textbf{25}(3), 339--370 (Jun 2022), ISSN 1099-1425, \doi{10.1007/s10951-022-00721-1}

\bibitem[{Vaswani et~al.(2017)Vaswani, Shazeer, Parmar, Uszkoreit, Jones, Gomez, Kaiser, and Polosukhin}]{vaswani2017attention}
Vaswani, A., Shazeer, N., Parmar, N., Uszkoreit, J., Jones, L., Gomez, A.N., Kaiser, {\L}., Polosukhin, I.: {Attention is All You Need}. In: Advances in Neural Information Processing Systems, vol.~30, pp. 5998--6008, Curran Associates, Inc., Long Beach, CA, USA (2017), \urlprefix\url{https://proceedings.neurips.cc/paper_files/paper/2017/file/3f5ee243547dee91fbd053c1c4a845aa-Paper.pdf}

\bibitem[{Verduin et~al.(2023)Verduin, Weise, and van~den Berg}]{Verduin2023}
Verduin, K., Weise, T., van~den Berg, D.: Why is the traveling tournament problem not solved with genetic algorithms? In: Evo* 2023 -- Late-Breaking Abstracts Volume, pp. 13--18, Species, Brno, Czech Republic (04 2023), \urlprefix\url{https://arxiv.org/pdf/2403.13950}

\bibitem[{Voboril et~al.(2024)Voboril, Ramaswamy, and Szeider}]{voboril2025generatingstreamliningconstraintslarge}
Voboril, F., Ramaswamy, V.P., Szeider, S.: {Generating Streamlining Constraints with Large Language Models}. arXiv (2024), \doi{10.48550/arXiv.2408.10268}

\bibitem[{Wang et~al.(2024{\natexlab{a}})Wang, Feng, He, Tan, Han, and Tsvetkov}]{wang2024languagemodelssolvegraph}
Wang, H., Feng, S., He, T., Tan, Z., Han, X., Tsvetkov, Y.: {Can Language Models Solve Graph Problems in Natural Language?} arXiv (2024{\natexlab{a}}), \doi{10.48550/arXiv.2305.10037}

\bibitem[{Wang et~al.(2023)Wang, Chen, and Zheng}]{wang2023opdnl4opt}
Wang, K., Chen, Z., Zheng, J.: {OPD@NL4Opt: An ensemble approach for the NER task of the optimization problem}. arXiv (2023), \doi{10.48550/arXiv.2301.02459}

\bibitem[{Wang et~al.(2024{\natexlab{b}})Wang, Ma, Feng, Zhang, Yang, Zhang, Chen, Tang, Chen, Lin, Zhao, Wei, and Wen}]{LeiWang_2024}
Wang, L., Ma, C., Feng, X., Zhang, Z., Yang, H., Zhang, J., Chen, Z., Tang, J., Chen, X., Lin, Y., Zhao, W.X., Wei, Z., Wen, J.: {A Survey on Large Language Model Based Autonomous Agents}. Frontiers of Computer Science \textbf{18}(6) (March 2024{\natexlab{b}}), ISSN 2095-2236, \doi{10.1007/s11704-024-40231-1}

\bibitem[{Wang et~al.(2024{\natexlab{c}})Wang, Farooq, Ghazzai, and Setti}]{Wang_2024}
Wang, Y., Farooq, J., Ghazzai, H., Setti, G.: {Multi-UAV Placement for Integrated Access and Backhauling Using LLM-Driven Optimization}. TechRxive (2024{\natexlab{c}}), \doi{10.36227/techrxiv.172833400.07230719/v1}

\bibitem[{Wang et~al.(2024{\natexlab{d}})Wang, Sambasivan, Fu, and Mehrotra}]{wang2024pivoting}
Wang, Y., Sambasivan, L.K., Fu, M., Mehrotra, P.: {Pivoting Retail Supply Chain with Deep Generative Techniques: Taxonomy, Survey and Insights}. arXiv (2024{\natexlab{d}}), \doi{10.48550/arXiv.2403.00861}

\bibitem[{Wasserkrug et~al.(2024)Wasserkrug, Boussioux, Hertog, Mirzazadeh, Birbil, Kurtz, and Maragno}]{wasserkrug2024large}
Wasserkrug, S., Boussioux, L., Hertog, D.d., Mirzazadeh, F., Birbil, I., Kurtz, J., Maragno, D.: {From Large Language Models and Optimization to Decision Optimization CoPilot: A Research Manifesto}. arXiv (2024), \doi{10.48550/arXiv.2402.16269}

\bibitem[{Wei et~al.(2022{\natexlab{a}})Wei, Tay, Bommasani, Raffel, Zoph, Borgeaud, Yogatama, Bosma, Zhou, Metzler, Chi, Hashimoto, Vinyals, Liang, Dean, and Fedus}]{wei2022emergent}
Wei, J., Tay, Y., Bommasani, R., Raffel, C., Zoph, B., Borgeaud, S., Yogatama, D., Bosma, M., Zhou, D., Metzler, D., Chi, E.H., Hashimoto, T., Vinyals, O., Liang, P., Dean, J., Fedus, W.: {Emergent Abilities of Large Language Models}. arXiv (2022{\natexlab{a}}), \doi{10.48550/arXiv.2206.07682}

\bibitem[{Wei et~al.(2022{\natexlab{b}})Wei, Wang, Schuurmans, Bosma, Richter, Xia, Chi, Le, and Zhou}]{NEURIPS2022_9d560961}
Wei, J., Wang, X., Schuurmans, D., Bosma, M., Richter, B., Xia, F., Chi, E., Le, Q.V., Zhou, D.: Chain-of-thought prompting elicits reasoning in large language models. In: Advances in Neural Information Processing Systems, vol.~35, pp. 24824--24837, Curran Associates, Inc., Virtual Conference (2022{\natexlab{b}}), \urlprefix\url{https://proceedings.neurips.cc/paper_files/paper/2022/file/9d5609613524ecf4f15af0f7b31abca4-Paper-Conference.pdf}

\bibitem[{{Windras Mara} et~al.(2022){Windras Mara}, Norcahyo, Jodiawan, Lusiantoro, and Rifai}]{WINDRASMARA2022105903}
{Windras Mara}, S.T., Norcahyo, R., Jodiawan, P., Lusiantoro, L., Rifai, A.P.: {A survey of adaptive large neighborhood search algorithms and applications}. Computers \& Operations Research \textbf{146}, 105903 (2022), ISSN 0305-0548, \doi{https://doi.org/10.1016/j.cor.2022.105903}

\bibitem[{Winter et~al.(2019)Winter, Musliu, Demirovi{\'c}, and Mrkvicka}]{winter2019solution}
Winter, F., Musliu, N., Demirovi{\'c}, E., Mrkvicka, C.: Solution approaches for an automotive paint shop scheduling problem. In: Proceedings of the International Conference on Automated Planning and Scheduling, vol.~29, pp. 573--581, AAAI Press, Virtual Conference (2019), \doi{10.1609/icaps.v29i1.3524}

\bibitem[{Wolpert and Macready(1997)}]{wolpert-1997-no-free-lunch}
Wolpert, D., Macready, W.: {No free lunch theorems for optimization}. IEEE Transactions on Evolutionary Computation \textbf{1}(1), 67--82 (1997), \doi{10.1109/4235.585893}

\bibitem[{Wu et~al.(2021)Wu, Ge, Zhang, Du, He, Ji, and Lang}]{00005792-202101290-00042}
Wu, C., Ge, Y., Zhang, X., Du, Y., He, S., Ji, Z., Lang, H.: {The combined effects of Lamaze breathing training and nursing intervention on the delivery in primipara: A PRISMA systematic review meta-analysis}. Medicine \textbf{100}(4) (2021), \doi{10.1097/MD.0000000000023920}

\bibitem[{Wu et~al.(2024{\natexlab{a}})Wu, Wang, Wen, Xiao, Wu, Wu, Yu, Maskell, and Zhou}]{wu2024neuralcombinatorialoptimizationalgorithms}
Wu, X., Wang, D., Wen, L., Xiao, Y., Wu, C., Wu, Y., Yu, C., Maskell, D.L., Zhou, Y.: {Neural Combinatorial Optimization Algorithms for Solving Vehicle Routing Problems: A Comprehensive Survey with Perspectives}. arXiv (2024{\natexlab{a}}), \doi{10.48550/arXiv.2406.00415}

\bibitem[{Wu et~al.(2024{\natexlab{b}})Wu, Wu, Wu, Feng, and Tan}]{wu2024evolutionary}
Wu, X., Wu, S.h., Wu, J., Feng, L., Tan, K.C.: {Evolutionary Computation in the Era of Large Language Model: Survey and Roadmap}. arXiv (2024{\natexlab{b}}), \doi{10.48550/arXiv.2401.10034}

\bibitem[{Wu et~al.(2024{\natexlab{c}})Wu, Zhong, Wu, Jiang, and Tan}]{10.24963/ijcai.2024/579}
Wu, X., Zhong, Y., Wu, J., Jiang, B., Tan, K.C.: {Large Language Model-Enhanced Algorithm Selection: Towards Comprehensive Algorithm Representation}. In: Proceedings of the Thirty-Third International Joint Conference on Artificial Intelligence, IJCAI '24, International Joint Conferences on Artificial Intelligence, Jeju Island, South Korea (2024{\natexlab{c}}), \doi{10.24963/ijcai.2024/579}

\bibitem[{Xiao et~al.(2024)Xiao, Zhang, Wu, Xu, Wang, Han, Fu, Zhong, Zeng, Song, and Chen}]{xiao2024chainofexperts}
Xiao, Z., Zhang, D., Wu, Y., Xu, L., Wang, Y.J., Han, X., Fu, X., Zhong, T., Zeng, J., Song, M., Chen, G.: {Chain-of-Experts: When LLMs Meet Complex Operations Research Problems}. In: The Twelfth International Conference on Learning Representations (ICLR 2024), OpenReview, Virtual Conference (2024), \urlprefix\url{https://openreview.net/forum?id=HobyL1B9CZ}

\bibitem[{Yang et~al.(2024{\natexlab{a}})Yang, Wang, Lu, Liu, Le, Zhou, and Chen}]{yang2024large}
Yang, C., Wang, X., Lu, Y., Liu, H., Le, Q.V., Zhou, D., Chen, X.: {Large Language Models as Optimizers}. arXiv (2024{\natexlab{a}}), \doi{10.48550/arXiv.2309.03409}

\bibitem[{Yang et~al.(2024{\natexlab{b}})Yang, Wang, Huang, Guo, Shi, Han, Feng, Song, Liang, and Tang}]{yang2024optibenchmeetsresocraticmeasure}
Yang, Z., Wang, Y., Huang, Y., Guo, Z., Shi, W., Han, X., Feng, L., Song, L., Liang, X., Tang, J.: {OptiBench Meets ReSocratic: Measure and Improve LLMs for Optimization Modeling} (2024{\natexlab{b}}), \doi{10.48550/arXiv.2407.09887}

\bibitem[{Yao et~al.(2024)Yao, Liu, Lin, Lu, Wang, and Zhang}]{yao2024multiobjectiveevolutionheuristicusing}
Yao, S., Liu, F., Lin, X., Lu, Z., Wang, Z., Zhang, Q.: {Multi-objective Evolution of Heuristic Using Large Language Model}. arXiv (2024), \doi{10.48550/arXiv.2409.16867}

\bibitem[{Yatong et~al.(2024)Yatong, Yuchen, and Yuqi}]{yatong2024tseohedgeservertask}
Yatong, W., Yuchen, P., Yuqi, Z.: {TS-EoH: An Edge Server Task Scheduling Algorithm Based on Evolution of Heuristic}. arXiv (2024), \doi{10.48550/arXiv.2409.09063}

\bibitem[{Ye et~al.(2024)Ye, Wang, Cao, Berto, Hua, Kim, Park, and Song}]{ye2024large}
Ye, H., Wang, J., Cao, Z., Berto, F., Hua, C., Kim, H., Park, J., Song, G.: {ReEvo: Large Language Models as Hyper-Heuristics with Reflective Evolution}. In: Proceedings of the Thirty-Eighth Annual Conference on Neural Information Processing Systems (NeurIPS 2024), pp. 1--32, Neural Information Processing Systems Foundation, Vancouver, Canada (2024), \urlprefix\url{https://openreview.net/forum?id=483IPG0HWL}

\bibitem[{You et~al.(2023)You, Ye, Zhou, Zhu, and Du}]{buildings13071772}
You, H., Ye, Y., Zhou, T., Zhu, Q., Du, J.: {Robot-Enabled Construction Assembly with Automated Sequence Planning Based on ChatGPT: RoboGPT}. Buildings \textbf{13}(7) (2023), ISSN 2075-5309, \doi{10.3390/buildings13071772}

\bibitem[{Yu and Liu(2024{\natexlab{a}})}]{yu2024autornetautomaticallyoptimizingheuristics}
Yu, H., Liu, J.: {AutoRNet: Automatically Optimizing Heuristics for Robust Network Design via Large Language Models}. arXiv (2024{\natexlab{a}}), \doi{10.48550/arXiv.2410.17656}

\bibitem[{Yu and Liu(2024{\natexlab{b}})}]{yu2024deepinsightsautomatedoptimization}
Yu, H., Liu, J.: {Deep Insights into Automated Optimization with Large Language Models and Evolutionary Algorithms}. arXiv (2024{\natexlab{b}}), \doi{10.48550/arXiv.2410.20848}

\bibitem[{Zanazzo et~al.(2024)Zanazzo, Ceschia, and Schaerf}]{10.1007/978-3-031-62912-9_15}
Zanazzo, E., Ceschia, S., Schaerf, A.: {Solving the Integrated Patient-to-Room and Nurse-to-Patient Assignment by Simulated Annealing}. In: Metaheuristics: 15th International Conference, MIC 2024, Lorient, France, June 4–7, 2024, Proceedings, Part I, p. 158–163, Springer-Verlag, Berlin, Heidelberg (2024), ISBN 978-3-031-62911-2, \doi{10.1007/978-3-031-62912-9_15}

\bibitem[{Zhang et~al.(2024{\natexlab{a}})Zhang, Wang, Guo, Wang, Lin, Yang, and Yin}]{zhang-etal-2024-solving}
Zhang, J., Wang, W., Guo, S., Wang, L., Lin, F., Yang, C., Yin, W.: {Solving General Natural-Language-Description Optimization Problems with Large Language Models}. In: Proceedings of the 2024 Conference of the North American Chapter of the ACL: Human Language Technologies (Volume 6: Industry Track), pp. 483--490, ACL, Mexico City, Mexico (Jun 2024{\natexlab{a}}), \doi{10.18653/v1/2024.naacl-industry.42}

\bibitem[{Zhang et~al.(2023)Zhang, Yin, Wang, Shen, Xiang, Wu, Zhao, Pan, Jiang, and Huang}]{zhang2023mindopttunerboostperformance}
Zhang, M., Yin, W., Wang, M., Shen, Y., Xiang, P., Wu, Y., Zhao, L., Pan, J., Jiang, H., Huang, K.: {MindOpt Tuner: Boost the Performance of Numerical Software by Automatic Parameter Tuning} (2023), \doi{10.48550/arXiv.2307.08085}

\bibitem[{Zhang et~al.(2024{\natexlab{b}})Zhang, Liu, Lin, Wang, Lu, and Zhang}]{10.1007/978-3-031-70068-2_12}
Zhang, R., Liu, F., Lin, X., Wang, Z., Lu, Z., Zhang, Q.: {Understanding the Importance of Evolutionary Search in Automated Heuristic Design with Large Language Models}. In: Parallel Problem Solving from Nature -- PPSN XVIII, pp. 185--202, Springer Nature Switzerland, Cham (2024{\natexlab{b}}), \doi{10.1007/978-3-031-70068-2_12}

\bibitem[{Zhang et~al.(2020)Zhang, Kishore, Wu, Weinberger, and Artzi}]{zhang2020bertscoreevaluatingtextgeneration}
Zhang, T., Kishore, V., Wu, F., Weinberger, K.Q., Artzi, Y.: {BERTScore: Evaluating Text Generation with BERT}. arXiv (2020), \doi{10.48550/arXiv.1904.09675}

\bibitem[{Zhang et~al.(2024{\natexlab{c}})Zhang, Wang, Zhang, Li, Qin, and Zhu}]{zhang2024llm4dyglargelanguagemodels}
Zhang, Z., Wang, X., Zhang, Z., Li, H., Qin, Y., Zhu, W.: {LLM4DyG: Can Large Language Models Solve Spatial-Temporal Problems on Dynamic Graphs?} arXiv (2024{\natexlab{c}}), \doi{10.48550/arXiv.2310.17110}

\bibitem[{Zhao et~al.(2024)Zhao, Cheng, Ding, Zhou, Zhang, Xu, and Zhao}]{zhao2024survey}
Zhao, Z., Cheng, S., Ding, Y., Zhou, Z., Zhang, S., Xu, D., Zhao, Y.: {A Survey of Optimization-Based Task and Motion Planning: From Classical to Learning Approaches}. IEEE/ASME Transactions on Mechatronics \textbf{29}(5), 1--27 (2024), \doi{10.1109/TMECH.2024.3452509}

\bibitem[{Çalık et~al.(2024)Çalık, Wauters, and Vanden~Berghe}]{CALIK2024106438}
Çalık, H., Wauters, T., Vanden~Berghe, G.: {The exam location problem: Mathematical formulations and variants}. Computers \& Operations Research \textbf{161}, 106438 (2024), ISSN 0305-0548, \doi{10.1016/j.cor.2023.106438}

\end{thebibliography}

\clearpage
\appendix

\section{PRISMA Checklists}
\label{sec:prisma-checklist}

This appendix provides the ``PRISMA 2020 for Abstracts Checklist'' and the ``PRISMA 2020 Checklist'' \cite{Pagen71, Pagen160}, \new{which were} used to guide our systematic review \new{on} the applications of \glspl{llm} in the field of \acrlong{co} (\Cref{sec:methodology}).

\begin{minipage}{\textwidth}
  

## A   PRISMA Checklists

This appendix provides the "PRISMA 2020 for Abstracts Checklist" and the "PRISMA 2020 Checklist" [163, 164], which were used to guide our systematic review on the applications of LLMs in the field of Combinatorial Optimization (Section 5).

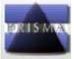 **PRISMA 2020 for Abstracts Checklist**

| Section and Topic | Item # | Checklist item | Reported (Yes/No) |
|---|---|---|---|
| **TITLE** | | | |
| Title | 1 | Identify the report as a systematic review. | Yes |
| **BACKGROUND** | | | |
| Objectives | 2 | Provide an explicit statement of the main objective(s) or question(s) the review addresses. | Yes |
| **METHODS** | | | |
| Eligibility criteria | 3 | Specify the inclusion and exclusion criteria for the review. | Yes |
| Information sources | 4 | Specify the information sources (e.g. databases, registers) used to identify studies and the date when each was last searched. | Yes |
| Risk of bias | 5 | Specify the methods used to assess risk of bias in the included studies. | No |
| Synthesis of results | 6 | Specify the methods used to present and synthesise results. | Yes |
| **RESULTS** | | | |
| Included studies | 7 | Give the total number of included studies and participants and summarise relevant characteristics of studies. | Yes |
| Synthesis of results | 8 | Present results for main outcomes, preferably indicating the number of included studies and participants for each. If meta-analysis was done, report the summary estimate and confidence/credible interval. If comparing groups, indicate the direction of the effect (i.e. which group is favoured). | Yes |
| **DISCUSSION** | | | |
| Limitations of evidence | 9 | Provide a brief summary of the limitations of the evidence included in the review (e.g. study risk of bias, inconsistency and imprecision). | No |
| Interpretation | 10 | Provide a general interpretation of the results and important implications. | Yes |
| **OTHER** | | | |
| Funding | 11 | Specify the primary source of funding for the review. | Within paper body |
| Registration | 12 | Provide the register name and registration number. | Not relevant |

*From:*  Page MJ, McKenzie JE, Bossuyt PM, Boutron I, Hoffmann TC, Mulrow CD, et al. The PRISMA 2020 statement: an updated guideline for reporting systematic reviews. BMJ 2021;372:n71. doi: 10.1136/bmj.n71

For more information, visit: http://www.prisma-statement.org/

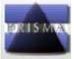 **PRISMA 2020 Checklist**

| Section and Topic | Item # | Checklist item | Location where item is reported |
|---|---|---|---|
| **TITLE** | | | |
| Title | 1 | Identify the report as a systematic review. | Title |
| **ABSTRACT** | | | |
| Abstract | 2 | See the PRISMA 2020 for Abstracts checklist. | Abstract, see PRISMA for Abstracts checklist |
| **INTRODUCTION** | | | |
| Rationale | 3 | Describe the rationale for the review in the context of existing knowledge. | Section 1 |
| Objectives | 4 | Provide an explicit statement of the objective(s) or question(s) the review addresses. | Section 2 |
| **METHODS** | | | |
| Eligibility criteria | 5 | Specify the inclusion and exclusion criteria for the review and how studies were grouped for the syntheses. | Section 5.3.3 |
| Information sources | 6 | Specify all databases, registers, websites, organisations, reference lists and other sources searched or consulted to identify studies. Specify the date when each source was last searched or consulted. | Section 5.3.2 |
| Search strategy | 7 | Present the full search strategies for all databases, registers and websites, including any filters and limits used. | Section 5.3.2 |
| Selection process | 8 | Specify the methods used to decide whether a study met the inclusion criteria of the review, including how many reviewers screened each record and each report retrieved, whether they worked independently, and if applicable, details of automation tools used in the process. | Section 5.3.3 |
| Data collection process | 9 | Specify the methods used to collect data from reports, including how many reviewers collected data from each report, whether they worked independently, any processes for obtaining or confirming data from study investigators, and if applicable, details of automation tools used in the process. | Section 5.3.4 |
| Data items | 10a | List and define all outcomes for which data were sought. Specify whether all results that were compatible with each outcome domain in each study were sought (e.g. for all measures, time points, analyses), and if not, the methods used to decide which results to collect. | Section 6 and Appendix B |
| | 10b | List and define all other variables for which data were sought (e.g. participant and intervention characteristics, funding sources). Describe any assumptions made about any missing or unclear information. | Section 6 and Appendix B |
| Study risk of bias assessment | 11 | Specify the methods used to assess risk of bias in the included studies, including details of the tool(s) used, how many reviewers assessed each study and whether they worked independently, and if applicable, details of automation tools used in the process. | Section 5.3 |
| Effect measures | 12 | Specify for each outcome the effect measure(s) (e.g. risk ratio, mean difference) used in the synthesis or presentation of results. | Not relevant |
| Synthesis methods | 13a | Describe the processes used to decide which studies were eligible for each synthesis (e.g. tabulating the study intervention characteristics and comparing against the planned groups for each synthesis (item #5)). | Section 5.3.3 |
| | 13b | Describe any methods required to prepare the data for presentation or synthesis, such as handling of missing summary statistics, or data conversions. | Not relevant |
| | 13c | Describe any methods used to tabulate or visually display results of individual studies and syntheses. | Section 6 |
| | 13d | Describe any methods used to synthesize results and provide a rationale for the choice(s). If meta-analysis was performed, describe the model(s), method(s) to identify the presence and extent of statistical heterogeneity, and software package(s) used. | Section 6 |
| | 13e | Describe any methods used to explore possible causes of heterogeneity among study results (e.g. subgroup analysis, meta-regression). | Not relevant |
| | 13f | Describe any sensitivity analyses conducted to assess robustness of the synthesized results. | Not relevant |
| Reporting bias assessment | 14 | Describe any methods used to assess risk of bias due to missing results in a synthesis (arising from reporting biases). | Section 8 |
| Certainty assessment | 15 | Describe any methods used to assess certainty (or confidence) in the body of evidence for an outcome. | Not relevant |
| **RESULTS** | | | |
| Study selection | 16a | Describe the results of the search and selection process, from the number of records identified in the search to the number of studies included in the review, ideally using a flow diagram. | Figure 2 |

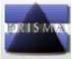 **PRISMA 2020 Checklist**

| Section and Topic | Item # | Checklist item | Location where item is reported |
|---|---|---|---|
| | 16b | Cite studies that might appear to meet the inclusion criteria, but which were excluded, and explain why they were excluded. | Section 5.3.4 |
| Study characteristics | 17 | Cite each included study and present its characteristics. | Appendix B |
| Risk of bias in studies | 18 | Present assessments of risk of bias for each included study. | Not relevant |
| Results of individual studies | 19 | For all outcomes, present, for each study: (a) summary statistics for each group (where appropriate) and (b) an effect estimate and its precision (e.g. confidence/credible interval), ideally using structured tables or plots. | Not relevant |
| Results of syntheses | 20a | For each synthesis, briefly summarise the characteristics and risk of bias among contributing studies. | Section 6 |
| | 20b | Present results of all statistical syntheses conducted. If meta-analysis was done, present for each the summary estimate and its precision (e.g. confidence/credible interval) and measures of statistical heterogeneity. If comparing groups, describe the direction of the effect. | Not relevant |
| | 20c | Present results of all investigations of possible causes of heterogeneity among study results. | Section 6 |
| | 20d | Present results of all sensitivity analyses conducted to assess the robustness of the synthesized results. | Not relevant |
| Reporting biases | 21 | Present assessments of risk of bias due to missing results (arising from reporting biases) for each synthesis assessed. | Section 8 |
| Certainty of evidence | 22 | Present assessments of certainty (or confidence) in the body of evidence for each outcome assessed. | Not relevant |
| **DISCUSSION** | | | |
| Discussion | 23a | Provide a general interpretation of the results in the context of other evidence. | Section 6 |
| | 23b | Discuss any limitations of the evidence included in the review. | Section 8 |
| | 23c | Discuss any limitations of the review processes used. | Section 8 |
| | 23d | Discuss implications of the results for practice, policy, and future research. | Section 7 and Section 9 |
| **OTHER INFORMATION** | | | |
| Registration and protocol | 24a | Provide registration information for the review, including register name and registration number, or state that the review was not registered. | Not relevant |
| | 24b | Indicate where the review protocol can be accessed, or state that a protocol was not prepared. | Not relevant |
| | 24c | Describe and explain any amendments to information provided at registration or in the protocol. | Not relevant |
| Support | 25 | Describe sources of financial or non-financial support for the review, and the role of the funders or sponsors in the review. | End of the paper, before references |
| Competing interests | 26 | Declare any competing interests of review authors. | End of the paper, before references |
| Availability of data, code and other materials | 27 | Report which of the following are publicly available and where they can be found: template data collection forms; data extracted from included studies; data used for all analyses; analytic code; any other materials used in the review. | The paper is self contained and all the material is included either in the paper or in the appendix. |

*From:* Page MJ, McKenzie JE, Bossuyt PM, Boutron I, Hoffmann TC, Mulrow CD, et al. The PRISMA 2020 statement: an updated guideline for reporting systematic reviews. BMJ 2021;372:n71. doi: 10.1136/bmj.n71

For more information, visit: http://www.prisma-statement.org/



\section{Description of Included Studies}
\label{sec:description-of-selected-studies}

This appendix provides the full list of the \numstudiesincluded studies found in the literature using the methodology described in \Cref{sec:methodology} and analyzed in \Cref{sec:analysis}. For each study, we report its type (\ref{sec:methodology_subsec:literature-collection_subsec:screening_criteria:inclusion-type-of-paper}-\ref{sec:methodology_subsec:literature-collection_subsec:screening_criteria:exclusion-type-of-paper}) and provide a brief description of the application of \glspl{llm} in the field of \gls{co}. When applicable, we also mention the specific \gls{cop} involved.

\begin{itemize}[label={--}, leftmargin=2em]

    \item \citet{abdullin-etal-2023-synthetic} published a research study introducing a goal-oriented conversational agent to assist users in constructing accurate \gls{lp} models. The \gls{llm} is used in the dialogue generation phase to automate the interaction between a conversational agent and a simulated user, effectively extracting necessary information to formulate \gls{lp} models.

    \item \citet{ahmaditeshnizi2024optimus} published a research study investigating the usage of \gls{llm} to formulate and solve \gls{lp} and \gls{milp} problems from \gls{nl} descriptions. The \gls{llm} is employed to develop an agent that recognizes entities and translates \gls{nl} descriptions into code. This code is subsequently used by a solver to find solutions and potentially debug them. \new{The work has been extended by \citet{ahmaditeshnizi2024optimus03usinglargelanguage}.}

    \item \citet{ahmed2024lm4opt} published a research study to investigate the usage of \glspl{llm} for recognizing entities and translating \gls{nl} descriptions in optimization problem formulations. A \gls{llm} is used for this task in both zero-shot and one-shot settings.
    
    \item \new{\citet{ALIPOURVAEZI2024110574} proposed a research study on using \glspl{llm} to extract domain knowledge in the context of the motion picture industry for portfolio optimization.}

    \item \citet{almonacid2023automatic} published a research study investigating \gls{llm} capabilities of generating optimization code. The \gls{llm} is used to generate the model and its code using MiniZinc, eventually debugging it.

    \item \citet{amarasinghe2023aicopilot} published a research study introducing a framework based on Copilot, designed to automate the generation of code from \gls{nl} descriptions. The \gls{llm} is used to automate the generation of Python code employing the CPMpy library starting from business descriptions.

    \item \new{\citet{bohnet2024exploringbenchmarkingplanningcapabilities} published a research study on using \glspl{llm} in generating solutions for planning problems.}

    \item \new{\citet{borazjanizadeh2024navigatinglabyrinthevaluatingenhancing} published a research study on solving search problems using \glspl{llm}. The framework involves generating solution code and producing the solution in a specific format. The results were evaluated in terms of feasibility, correctness, and optimality. The study also introduces a new benchmark called SearchBench.}

    \item \new{\citet{10.1145/3638530.3664086} published a position paper on the possible advantages \glspl{llm} can bring to evolutionary computation, including evolutionary-based optimization.}

    \item \citet{sartori2024large} published a research study integrating \glspl{llm} into a web-based tool for visualizing the behavior of optimization algorithms. The \gls{llm} generates prompts for the tool, enabling users to comprehend how multiple algorithms behave when applied to specific instances of a \gls{cop}.

    \item \new{\citet{chen2024uberuncertaintybasedevolutionlarge} published a research study related to FunSearch \cite{Romera-Paredes2024}; specifically, it introduces the concept of uncertainty to maintain diversity in the population of codes while ensuring a balance between exploration and intensification during the search. The results are compared to FunSearch and Evolution of Heuristic \cite{10.5555/3692070.3693374}.}

    \item \citet{chin2024learning} published a research study addressing the resolution of a particular \gls{co}, the \gls{fmcvrp}. The \gls{llm} is used to train a model in a supervised manner on computationally inexpensive, sub-optimal solutions obtained algorithmically.

    \item \new{\citet{Da2024} published a research study on integrating \glspl{llm} into traffic and routing contexts. Specifically, the framework enables user--\gls{llm} conversations, generates solutions based on domain knowledge, and allows for solution visualization and validation.}

    \item \new{\citet{delarosa2024trippaltravelplanningguarantees} published a research study on using \glspl{llm} for travel planning. The \gls{llm} retrieves information about a given city and generates a route to visit specified landmarks.}

    \item \citet{ai5010006} published a research study addressing the resolution of problems in a specific \gls{co} domain (financial and budget optimization). The \gls{llm} is used to provide domain-specific advice to the user.

    \item \new{\citet{10711695} published a research study on integrating human preferences into task scheduling for human-robot teams using \gls{cp} and \glspl{llm}.}

    \item \citet{doan2022vtccnlp} published a research study on the topic of recognizing optimization entities. The \gls{llm} is used to recognize such entities from problem descriptions expressed in \gls{nl}.

    \item \new{\citet{make6030093} proposed a research study on solving the \gls{tsp} using \glspl{llm}. Differently from other studies, this one uses the visual capabilities of \glspl{llm}.}

    \item \citet{fan2024artificial} published a literature review exploring the integration of \gls{ai} within the \gls{or} process, focusing on enhancing various stages such as parameter generation and model formulation.

    \item \citet{Freuder_2024} published a position paper discussing the role of \glspl{llm} in constraint satisfaction problems, where traditionally a crucial component is the dialogue between an optimization expert and a domain expert. The \gls{llm} is used to replace the optimization expert.

    \item \citet{gangwar2023highlighting} published a research study describing a method to translate \gls{nl} descriptions into \gls{lp} problem formulations.

    \item \new{\citet{a17120582} proposed a research study on using \glspl{llm} to integrate stakeholders' preferences in decision-making processes. The methodology is applied to a last-mile delivery problem, and the \gls{llm} is used in two ways: to act as a construction heuristic and to set the weights of the optimizer (i.e., parameter tuning).}

    \item \new{\citet{guo2024optimizinglargelanguagemodels} published a research study on using \glspl{llm} to generate solutions. Specifically, the \gls{llm} is asked to mimic the behavior of well-known algorithms, such as \gls{hc}.}

    \item \new{\citet{hao2025planningrigorgeneralpurposezeroshot} published a research study on using \glspl{llm} in planning. Specifically, \glspl{llm} are used to recognize entities, formulate the model, generate code, and catch errors.}

    \item \citet{chen2023diagnosing} published a research study on the usage of \gls{llm} in detecting model infeasibility. The \gls{llm} is used to provide \gls{nl} descriptions of the optimization model itself, identify potential sources of infeasibility, and offer suggestions to make the model feasible.

    \item \citet{he2022linear} published a research study describing a method to generate \gls{lp} problem formulations from \gls{nl} descriptions. The \gls{llm} automates the transformation of text-based \gls{lp} problem descriptions into structured formats, including entity recognition tasks.

    \item \new{\citet{hu2024scalableaccurategraphreasoning} published a research study on finding solutions for graph-related problems. Additionally, an explainability layer is included to outline the reasoning process.}

    \item \new{\citet{huang2024exploringtruepotentialevaluating} published a research study on using \glspl{llm} to generate solutions for \glspl{cop}. The experiments are conducted on the \gls{tsp}.}

    \item \new{\citet{huang2025orlmcustomizableframeworktraining} published a research study proposing a tool called OR-Instruct, used to create synthetic data tailored to optimization modeling. They test the approach by developing the IndustryOR dataset.}

    \item \citet{huang2024large} published a literature review exploring the integration of \gls{llm} and optimization, considering both the perspective of \gls{llm} for optimization and optimization for \gls{llm}. The \gls{llm} is used as a black-box optimization search model or to generate optimization algorithms.

    \item \citet{huang2024multimodal} published a research study proposing an optimization framework based on \glspl{llm} in the context of \gls{cvpr}. The \gls{llm} is used to generate solutions using textual and visual prompts.

    \item \citet{huang2024words} published a research study addressing a specific class of \glspl{cop}, the \glspl{vrp}. The \gls{llm} is used to generate Python code from \gls{nl} descriptions.

    \item \citet{jang2022tag} published a research study introducing a method to generate \gls{lp} problem formulations. The \gls{llm} is used to translate \gls{nl} description into \gls{lp} formulation with entity recognition.

    \item \new{\citet{jiang2025llmopt} published a research study on using \glspl{llm} in an end-to-end approach: the \gls{llm} generates solution code directly from \gls{nl} problem descriptions.}

    \item \new{\citet{jiang2024largelanguagemodelscombinatorial} published a research study on using \glspl{llm} as optimizers, i.e., generating solutions in the context of planning.}

    \item \new{\citet{10738100} proposed a research study on using \glspl{llm} for problem formulation and code generation in the context of energy management.}

    \item \new{\citet{10.1145/3664646.3665084} published a research study on using \glspl{llm} to create feasible schedules for conferences. The \gls{llm} either groups presentation titles or generates solutions.}

    \item \new{\citet{ju-etal-2024-globe} published a research study on using \glspl{llm} for travel planning. \glspl{llm} recognize entities, create a corresponding \gls{milp} model, and solve the code.}

    \item \new{\citet{Khan_2024} published a research study investigating \glspl{llm}'s capabilities in solving \glspl{cop}. Specifically, \glspl{llm} either generate code or produce solutions. The study considers the \gls{tsp}, the assignment problem, the transportation problem, and the shortest path problem.}

    \item \citet{10.1007/978-981-97-2259-4_3} published a research study introducing a framework designed to explain the decision-making process for a certain class of problems. The \gls{llm} is used to explain the influence of each route edge and integrate counterfactual explanations. The \gls{cop} addressed is the \gls{vrp}.

    \item \new{\citet{han-2024-syrvey} reviewed the literature on \glspl{llm} regarding their mathematical capabilities and briefly discussed their applications in \gls{co}, along with math word problems, geometry problems, and theorem proving.}

    \item \new{\citet{lawless2024llmscoldstartcuttingplane} published a research study on using \glspl{llm} to decide which cutting-plane strategy to use, thus assisting the \gls{milp} solver in choosing the appropriate parameters.}

    \item \citet{lawless2024i} published a research study introducing a hybrid framework that integrates \glspl{llm} with \gls{cp} to enable interactive decision support. The \gls{llm} is used in the preference elicitation phase to translate user input into structured constraint functions that the \gls{cp} model can understand. The \gls{cop} addressed is the Meeting Scheduling Problem, a specific \gls{csp}.

    \item \citet{li2023large} published a research study investigating the usage of \gls{llm} for reasoning about supply chain optimization. The \gls{llm} translates human queries into code, which is then utilized by an optimization solver. The final answer is returned via the \gls{llm}.

    \item \new{\citet{10704489} published a research study on using \glspl{llm} to generate solutions for a travel planning problem. The solutions are evaluated and validated with user feedback.}

    \item \citet{li2023synthesizing} published a research study describing a method for synthesizing \gls{milp} models directly from \gls{nl} descriptions. The \gls{llm} is used in the initial modeling phase to identify and classify decision variables and constraints.

    \item \new{\citet{li2024foundationmodelsmixedinteger} published a research study on generating \gls{milp} codes (and instances) with \glspl{llm} to train learning-based methods. The pipeline is called MILP-Evolve.}

    \item \new{\citet{li2025graphteamfacilitatinglargelanguage} published a research study on reasoning with graphs and networks. \glspl{llm} are used for knowledge extraction, entity recognition, and code generation in Python.}

    \item \citet{liu2024large} published a research study on automating the design of search operators within multi-objective evolutionary procedures. The \gls{llm} is used to generate new individuals for each subproblem within the decomposition approach.

    \item \citet{liu2024evolution} published a research study on the usage of \gls{llm} in automatic heuristic design. The \gls{llm} is used to generate and evolve heuristic algorithm components, tested on \glspl{tsp}, \gls{pfsp}, and online \gls{bpp}.

    \item \citet{liu2023algorithm} published a research study on automatically generating optimization algorithms through an evolutionary framework. The \gls{llm} creates the initial solutions—each individual represents an algorithm—and applies mutation and crossover. The \gls{cop} addressed is the \gls{tsp}.

    \item \new{\citet{liu2024systematicsurveylargelanguage} published a literature review on using \glspl{llm} for algorithm design, including optimization algorithms.}

    \item \new{\citet{liu2024llm4adplatformalgorithmdesign} published a research study on algorithm design (heuristic code generation in Python), comparing results with other approaches such as FunSearch \cite{Romera-Paredes2024}.}

    \item \citet{liu2024large1} published a research study investigating \glspl{llm} as evolutionary combinatorial optimizers. A \gls{llm} selects parent solutions from a given population, performing crossover and mutation to generate offspring. The \gls{cop} addressed is the \gls{tsp}.

    \item \citet{LIU2023100520} published a research study investigating the integration of \gls{llm} in a delivery route optimization problem (a variation of the \gls{tsp}). The \gls{llm} is used to gather knowledge regarding delivery patterns.

    \item \new{\citet{10829820} published a literature review on Network Operations and Performance Optimization, discussing potential roles for \glspl{llm}.}

    \item \citet{mao2024identify} published a research study investigating the usage of \gls{llm} to enhance evolutionary procedures for identifying critical nodes in a graph. The \gls{llm} crosses and mutates given functions to generate new code snippets.

    \item \new{\citet{DBLP:conf/esann/MartinekLG24} proposed a research study in which \glspl{llm} are used to set \glspl{mh} parameters, tested on the \gls{tsp} and the Graph Coloring Problem.}

    \item \new{\citet{michailidis_et_al:LIPIcs.CP.2024.20} proposed a research study examining \glspl{llm} from entity recognition to solution generation. In the context of entity recognition, the authors used Ner4Opt \cite{10.1007/978-3-031-33271-5_20}.}

    \item \new{\citet{10675146} published a research study on generating solutions using \glspl{llm} in the context of planning.}

    \item \new{\citet{mostajabdaveh2024evaluatingllmreasoningoperations} published a research study analyzing the capabilities of \glspl{llm} in \gls{co} compared to human experts. A benchmark dataset called ORQUA is introduced, evaluating \glspl{llm} by asking them to answer questions based on \gls{nl} problem descriptions.}

    \item \new{\citet{Mostajabdaveh04112024} published a research study on using \glspl{llm} in an end-to-end optimization process, where the \gls{llm} generates solution code from \gls{nl} problem descriptions.}

    \item \citet{teukam2024integrating} published a research study introducing a framework for optimizing enzymes. The \gls{llm} learns relationships between amino acid residues linked to structure and function, which are then used as input for an evolutionary search procedure aimed at improved catalytic performance.

    \item \citet{ning2023novel} published a research study on recognizing optimization entities and providing \gls{lp} problem formulations. The \gls{llm} is used to identify these entities from \gls{nl} problem descriptions and to generate the \gls{lp} model.

    \item \new{\citet{10803039} proposed a research study on integrating \gls{lp} and Dependency Graph approaches with \glspl{llm} for robot planning tasks.}

    \item \new{\citet{10.1609/icaps.v34i1.31503} proposed a review on the applications of \glspl{llm} in planning, including language translation, plan generation, model construction, multi-agent planning, interactive planning, heuristic optimization, tool integration, and brain-inspired planning.}

    \item \citet{ramamonjison-etal-2022-augmenting} published a research study to recognize entities and generate \gls{lp} formulations from \gls{nl} descriptions of \gls{lp} problems.

    \item \citet{ramamonjison2023nl4opt} published a technical report on the \gls{nl4opt} competition, describing tasks, datasets, metrics for evaluation, and competition statistics.

    \item \new{\citet{regin_et_al:LIPIcs.CP.2024.25} proposed a research study introducing GenCP, a framework combining \gls{cp} with \glspl{llm} to handle structural constraints, including the semantic meaning in text generation.}

    \item \new{\citet{reinhart-2024} proposed a research study on using \glspl{llm} to generate solutions for macromolecule design. The \gls{llm} was also used to explain or justify the proposed solution.}

    \item \citet{RePEc:spr:snopef:v:4:y:2023:i:4:d:10.1007_s43069-023-00277-6} published a research study evaluating \gls{llm} stock-picking capability. The \gls{llm} is used to generate financial assets (solutions) for which a Mean-Variance Cardinality-Constrained Portfolio Optimization Model is computed.

    \item \citet{Romera-Paredes2024} published a research study proposing a new evolutionary procedure. The \gls{llm} is integrated to find new solutions and heuristics for \glspl{cop} by evolving the code of an initial program skeleton at each step. The \glspl{cop} addressed are the Cap Set problem and online bin packing.

    \item \citet{SAKA2024100300} published a literature review investigating the usage of \gls{gpt} models in the \gls{aec} industry. The review identifies opportunities, evaluates limitations, and validates a \gls{llm} use case for solution generation in \gls{aec}.

    \item \new{\citet{10818476} published a research study proposing a tool called OptiPattern. The \gls{llm} extracts knowledge from instances and integrates it into the \gls{mh} (\gls{ga}) process. The methodology is applied to a network problem.}

    \item \new{\citet{srivastava2024casedevelopingfoundationmodel} published a position paper on the application of \glspl{llm} to planning and scheduling.}

    \item \new{\citet{sui-2024} published a literature review addressing deep learning methods for the \gls{tsp}, which also mentions the role of \glspl{llm}.}

    \item \new{\citet{sun2024autosatautomaticallyoptimizesat} presented a research study on leveraging \glspl{llm} to solve SAT problems through a framework called AutoSAT. The framework automatically selects and optimizes heuristics within a predefined search space, building on conflict-driven clause learning solvers.}

    \item \new{\citet{10628050} published a research study on using \glspl{llm} in protein design. The \gls{llm} generates feasible solutions from incomplete ones, acting similarly to a repair operator in local neighborhood search (LNS).}

    \item \citet{tsouros2023holy} published a research study introducing a framework for translating \gls{nl} descriptions into \gls{cp}. The \gls{llm} is used to produce executable code that can be run and debugged.

    \item \new{\citet{smartcities7050094} published a research study on applying \glspl{llm} to domain knowledge and entity recognition. Specifically, the \gls{llm} builds knowledge graphs on the problem at hand (e.g., transportation networks).}

    \item \new{\citet{10720437} published a research study on using \glspl{llm} for parameter tuning. Their methodology leverages BERT-based embeddings to predict solver parameters for \gls{milp} problems, aiming to automate and improve efficiency in Gurobi.}

    \item \new{\citet{vanstein2024intheloophyperparameteroptimizationllmbased} published a research study on integrating \gls{llm}-generated code with parameter tuning. The code is generated via LLaMEA \cite{10752628}, and the tuning is performed using SMAC \cite{JMLR:v23:21-0888}. While LLaMEA was originally designed for black-box optimization, this study also addresses \gls{co} (\gls{tsp} and online bin packing).}

    \item \new{\citet{voboril2025generatingstreamliningconstraintslarge} proposed a research study introducing StreamLLM. The \gls{llm} enriches existing \gls{cp} code in Python with additional constraints, aiming to reduce the search space.}

    \item \citet{wang2023opdnl4opt} published a research study on the topic of recognizing optimization entities. The \gls{llm} identifies such entities from \gls{nl} problem descriptions.

    \item \new{\citet{Wang_2024} published a research study on using \glspl{llm} as optimizers, i.e., generating solutions for drone placement.}

    \item \citet{wasserkrug2024large} published a position paper advocating for introducing \glspl{llm} into \gls{co} to democratize optimization practices, helping non-experts across different sectors.

    \item \new{\citet{wu2024neuralcombinatorialoptimizationalgorithms} published a literature review on vehicle routing, noting \glspl{llm} among other deep learning methods for \gls{tsp}.}

    \item \citet{wu2024evolutionary} published a literature review on the intersection between \gls{llm} and evolutionary computation, offering insights on possible research directions.

    \item \new{\citet{10.24963/ijcai.2024/579} published a research study on using \glspl{llm} for algorithm selection. More specifically, the \gls{llm} extracts features related to the underlying optimization algorithms.}

    \item \citet{xiao2024chainofexperts} published a research study introducing a multi-agent framework that automates the translation of problem descriptions into mathematical formulations and executable code. The \gls{llm} formulates models from \gls{nl} descriptions and generates runnable code.

    \item \citet{yang2024large} published a research study on \glspl{llm} as optimizers. The \gls{llm} generates candidate solutions based on the problem description and previously evaluated solutions in the meta-prompt. The \gls{cop} addressed is the \gls{tsp}.

    \item \new{\citet{yang2024optibenchmeetsresocraticmeasure} proposed a research study on using \glspl{llm} for entity recognition and Python code generation. The final \gls{llm} output also includes the solution, though not directly computed by the \gls{llm}. Two datasets, Optibench and ReSocratic-29k, are introduced.}

    \item \new{\citet{yao2024multiobjectiveevolutionheuristicusing} published a research study on generating code that balances two objectives: efficiency of the code and the quality of solutions generated by it, tested on \gls{tsp} and online bin packing.}

    \item \new{\citet{yatong2024tseohedgeservertask} published a research study on generating heuristic code for task scheduling in edge servers.}

    \item \citet{ye2024large} published a research study on integrating \gls{llm} into hyper-heuristics generation. The \gls{llm} is used to generate heuristic algorithms in Python, tested on well-known \glspl{cop} like \gls{tsp}, \gls{cvpr}, \gls{op}, \gls{dpp}, \gls{mkp}, and \gls{bpp}.

    \item \citet{buildings13071772} published a research study on a domain-specific \gls{cop} (robot task sequencing). The \gls{llm} is used as a black-box optimizer.

    \item \new{\citet{yu2024autornetautomaticallyoptimizingheuristics} published a research study on the usage of \glspl{llm} for robust network design. The \gls{llm} generates heuristic code.}

    \item \new{\citet{yu2024deepinsightsautomatedoptimization} published a position paper on the evolution of optimization toward automation, mentioning the integration of \glspl{llm}.}

    \item \new{\citet{zhang-etal-2024-solving} published a research study proposing OptLLM, a system that accepts user queries in natural language, converts them into mathematical formulations and programming code, and calls solvers for decision-making. OptLLM supports multi-round dialogues to iteratively refine modeling and solving.}

    \item \new{\citet{10.1007/978-3-031-70068-2_12} published a research study on automated heuristic design (code generation). Results are compared with ReEvo \cite{ye2024large} and FunSearch \cite{Romera-Paredes2024}.}

    \item \citet{zhao2024survey} published a literature review on optimization-based task and motion planning within a domain-specific \gls{co}. The review discusses how \glspl{llm} might generate domain knowledge and goal descriptions for planning methods.

\end{itemize}

\clearpage

\section{Classification of Studies by Optimization Process Step}
\label{app:analysis-co}

\new{This appendix provides a breakdown of the identified studies with respect to the optimization process (\Cref{tab:analysis-co})}. Columns display the general tasks (e.g., problem modeling), further divided into activities (e.g., domain knowledge). Furthermore,  we report the optimization paradigm (e.g., \gls{cp}) and technical details on the implementation (i.e., programming languages and commercial solvers). Each row groups together studies that share the same characteristics. The checkmark symbol (\checkmark) identifies whether a study performs/is related to the column item. 

\begin{longtable}{
    p{0.9cm}| 
    p{0.15cm} p{0.15cm} p{0.15cm}| 
    p{0.15cm} p{0.15cm} p{0.15cm} p{0.15cm}| 
    p{0.15cm} p{0.15cm}| 
    p{0.15cm} p{0.15cm}| 
    p{0.12cm} p{0.12cm} p{0.12cm} p{0.12cm} p{0.12cm} p{0.12cm} p{0.12cm} p{0.12cm} p{0.12cm}p{0.12cm} p{0.12cm} p{0.12cm} p{0.12cm}| 
    p{0.93cm}| 
    p{1.2cm} 
}
    \caption{Classification of the studies by Combinatorial Optimization task within the optimization process.}\label{tab:analysis-co} \\
    
    \hline

    \footnotesize\textbf{Studies} &
    \multicolumn{3}{c|}{\footnotesize\textbf{Prob. Mod.}} &
    \multicolumn{4}{c|}{\footnotesize\textbf{Sol. Meth.}} &
    \multicolumn{2}{c|}{\footnotesize\textbf{Ben.}} &
    \multicolumn{2}{c|}{\footnotesize\textbf{Valid.}} &
    \multicolumn{13}{c|}{\footnotesize\textbf{Models/Algorithms Types}} &
    \footnotesize\textbf{Prog. Lang.} &
    \footnotesize\textbf{Lib./ Solv.} \\

    \cline{2-4} \cline{5-8} \cline{9-10}  \cline{11-12} \cline{13-25}

     & \rotatebox{90}{\footnotesize Domain Know.} & 
     \rotatebox{90}{\footnotesize Entity Rec.} & 
     \rotatebox{90}{\footnotesize Model Creation} & 
     \rotatebox{90}{\footnotesize Code Gen.} & 
     \rotatebox{90}{\footnotesize Solution Gen.} & 
     \rotatebox{90}{\footnotesize Param. Tuning} & 
     \rotatebox{90}{\footnotesize Alg. Selection} & 
     \rotatebox{90}{\footnotesize Explain.} & 
     \rotatebox{90}{\footnotesize Visual Analysis} & 
     \rotatebox{90}{\footnotesize Sol. Valid.} & 
     \rotatebox{90}{\footnotesize Model Val.} & 
     \rotatebox{90}{\footnotesize CP} & 
     \rotatebox{90}{\footnotesize LP} & 
     \rotatebox{90}{\footnotesize ILP} & 
     \rotatebox{90}{\footnotesize MILP} & 
     \rotatebox{90}{\footnotesize Heuristic} & 
     \rotatebox{90}{\footnotesize HH} & 
     \rotatebox{90}{\footnotesize EA} & 
     \rotatebox{90}{\footnotesize GA} & 
     \rotatebox{90}{\footnotesize QAP} & 
     \rotatebox{90}{\footnotesize MOO} & 
     \rotatebox{90}{\footnotesize Non Linear} & 
     \rotatebox{90}{\footnotesize MH (general)} & 
     \rotatebox{90}{\footnotesize SAT} & &  \\
    \hline
     \endfirsthead
     
    \caption{Classification of the studies by Combinatorial Optimization task within the optimization process (continued).}\\
     \hline

     \footnotesize\textbf{Studies} &
     \multicolumn{3}{c|}{\footnotesize\textbf{Prob. Mod.}} &
     \multicolumn{4}{c|}{\footnotesize\textbf{Sol. Meth.}} &
     \multicolumn{2}{c|}{\footnotesize\textbf{Ben.}} &
     \multicolumn{2}{c|}{\footnotesize\textbf{Valid.}} &
     \multicolumn{13}{c|}{\footnotesize\textbf{Models/Algorithms Types}} &
     \footnotesize\textbf{Prog. Lang.} &
     \footnotesize\textbf{Lib./ Solv.} \\
 
     \cline{2-4} \cline{5-8} \cline{9-10}  \cline{11-12} \cline{13-25}
 
    & \rotatebox{90}{\footnotesize Domain Know.} & 
     \rotatebox{90}{\footnotesize Entity Rec.} & 
     \rotatebox{90}{\footnotesize Model Creation} & 
     \rotatebox{90}{\footnotesize Code Gen.} & 
     \rotatebox{90}{\footnotesize Solution Gen.} & 
     \rotatebox{90}{\footnotesize Param. Tuning} & 
     \rotatebox{90}{\footnotesize Alg. Selection} & 
     \rotatebox{90}{\footnotesize Explain.} & 
     \rotatebox{90}{\footnotesize Visual Analysis} & 
     \rotatebox{90}{\footnotesize Sol. Valid.} & 
     \rotatebox{90}{\footnotesize Model Val.} & 
     \rotatebox{90}{\footnotesize CP} & 
     \rotatebox{90}{\footnotesize LP} & 
     \rotatebox{90}{\footnotesize ILP} & 
     \rotatebox{90}{\footnotesize MILP} & 
     \rotatebox{90}{\footnotesize Heuristic} & 
     \rotatebox{90}{\footnotesize HH} & 
     \rotatebox{90}{\footnotesize EA} & 
     \rotatebox{90}{\footnotesize GA} & 
     \rotatebox{90}{\footnotesize QAP} & 
     \rotatebox{90}{\footnotesize MOO} & 
     \rotatebox{90}{\footnotesize Non Linear} & 
     \rotatebox{90}{\footnotesize MH (general)} & 
     \rotatebox{90}{\footnotesize SAT} & &  \\
     \hline
    \endhead
    
    \hline
    \multicolumn{26}{r}{\textit{\footnotesize Continued on next page}} \\
    \endfoot

    \hline
    \endlastfoot
    
    \hline
    \endlastfoot

    \footnotesize{\cite{LIU2023100520,ai5010006,zhao2024survey}} &\footnotesize \checkmark&  &  &  &  &  &  &  &  &  &  &  &  &  &  &  &  &  &  &  &  &  &  &  &  &  \\
    
    \footnotesize{\cite{10818476}} &\footnotesize \checkmark&  &  &  &  &  &  &  &  &  &  &  &  &  &  &  &  &  &\footnotesize \checkmark&  &  &  &  &  &  &  \\
    
    \footnotesize{\cite{delarosa2024trippaltravelplanningguarantees}} &\footnotesize \checkmark&  &  &  &\footnotesize \checkmark&  &  &  &  &  &  &  &  &  &  &  &  &  &  &  &  &  &  &  &  &  \\
    
    \footnotesize{\cite{10.1145/3664646.3665084}} &\footnotesize \checkmark&  &  &  &\footnotesize \checkmark&  &  &  &  &  &  &  &  &\footnotesize \checkmark&  &  &  &  &  &  &  &  &  &  &  &  \\
    
    \footnotesize{\cite{ALIPOURVAEZI2024110574,smartcities7050094}} &\footnotesize \checkmark&\footnotesize \checkmark&  &  &  &  &  &  &  &  &  &  &  &  &  &  &  &  &  &  &  &  &  &  &  &  \\
    
    \footnotesize{\cite{Da2024}} &\footnotesize \checkmark&\footnotesize \checkmark&  &  &\footnotesize \checkmark&  &  &  &\footnotesize \checkmark&\footnotesize \checkmark&&  &  &  &  &  &  &  &  &  &  &  &  &  &  &  \\
    
    \footnotesize{\cite{10711695}} &\footnotesize \checkmark&\footnotesize \checkmark&\footnotesize \checkmark&  &  &  &  &  &  &  &  &\footnotesize \checkmark&  &  &  &  &  &  &  &  &  &  &  &  & \footnotesize CP-SAT &  \\
    
    \footnotesize{\cite{li2025graphteamfacilitatinglargelanguage}} &\footnotesize \checkmark&\footnotesize \checkmark&  &\footnotesize \checkmark&  &  &  &  &  &  &  &  &  &  &  &  &  &  &  &  &  &  &  &  & \footnotesize Python &  \\
    
    \footnotesize{\cite{wang2023opdnl4opt,doan2022vtccnlp}} &  &\footnotesize \checkmark&  &  &  &  &  &  &  &  &  &  &\footnotesize \checkmark&  &  &  &  &  &  &  &  &  &  &  &  &  \\
    
    \footnotesize{\cite{huang2024words}} &  &\footnotesize \checkmark&  &\footnotesize \checkmark&  &  &  &  &  &\footnotesize \checkmark&  &  &  &  &  &  &  &  &  &  &  &  &  &  & \footnotesize Python &  \\
    
    \footnotesize{\cite{li2023synthesizing}} &  &\footnotesize \checkmark&\footnotesize \checkmark&  &  &  &  &  &  &  &  &  &  &  &\footnotesize \checkmark&  &  &  &  &  &  &  &  &  &  &  \\
    
    \footnotesize{\cite{ning2023novel,he2022linear,jang2022tag,ramamonjison-etal-2022-augmenting,ahmed2024lm4opt}} &  &\footnotesize \checkmark&\footnotesize \checkmark&  &  &  &  &  &  &  &  &  &\footnotesize \checkmark&  &  &  &  &  &  &  &  &  &  &  &  &  \\
    
    \footnotesize{\cite{10803039}} &  &\footnotesize \checkmark&\footnotesize \checkmark&  &\footnotesize \checkmark&  &  &  &  &  &  &  &\footnotesize \checkmark&  &  &  &  &  &  &  &  &  &  &  &  &  \\
    
    \footnotesize{\cite{tsouros2023holy}} &  &\footnotesize \checkmark&\footnotesize \checkmark&\footnotesize \checkmark&  &  &  &  &  &  &  &\footnotesize \checkmark&  &  &  &  &  &  &  &  &  &  &  &  & \footnotesize Python & \footnotesize CMPy \cite{guns2019increasing} \\
    
    \footnotesize{\cite{ju-etal-2024-globe}} &  &\footnotesize \checkmark&\footnotesize \checkmark&\footnotesize \checkmark&  &  &  &  &  &  &  &  &  &  &\footnotesize \checkmark&  &  &  &  &  &  &  &  &  &  &  \\
    
    \footnotesize{\cite{Mostajabdaveh04112024}} &  &\footnotesize \checkmark&\footnotesize \checkmark&\footnotesize \checkmark&  &  &  &  &  &  &\footnotesize \checkmark&  &\footnotesize \checkmark&\footnotesize \checkmark&  &  &  &  &  &\footnotesize \checkmark&  &  &  &  &  & \footnotesize Zimpl \\
    
    \footnotesize{\cite{hao2025planningrigorgeneralpurposezeroshot}} &  &\footnotesize \checkmark&\footnotesize \checkmark&\footnotesize \checkmark&  &  &  &  &  &  &\footnotesize \checkmark&  &  &  &\footnotesize \checkmark&  &  &  &  &  &  &  &  &\footnotesize \checkmark& \footnotesize Python & \footnotesize MST, Gurobi \\
    
    \footnotesize{\cite{jiang2025llmopt,yang2024optibenchmeetsresocraticmeasure,ahmaditeshnizi2024optimus03usinglargelanguage,ahmaditeshnizi2024optimus}} &  &\footnotesize \checkmark&\footnotesize \checkmark&\footnotesize \checkmark&  &  &  &  &  &  &  &  &\footnotesize \checkmark&  &\footnotesize \checkmark&  &  &  &  &  &  &  &  &  & \footnotesize Python & \footnotesize Pyomo~\cite{Hart2011}, pyscipopt, Gurobi~\cite{gurobi} (gurobipy) \\
    
    \footnotesize{\cite{zhang-etal-2024-solving}} &  &\footnotesize \checkmark&\footnotesize \checkmark&\footnotesize \checkmark&  &  &  &  &  &\footnotesize \checkmark&\footnotesize\checkmark  &  &\footnotesize \checkmark&  &\footnotesize \checkmark&  &  &  &  &  &  &  &\footnotesize \checkmark&  & \footnotesize MAPL code & \footnotesize MindOpt \cite{zhang2023mindopttunerboostperformance} \\
    
    \footnotesize{\cite{michailidis_et_al:LIPIcs.CP.2024.20}} &  &\footnotesize \checkmark&\footnotesize \checkmark&\footnotesize \checkmark&\footnotesize \checkmark&  &  &  &  &  &  &\footnotesize \checkmark&  &  &  &  &  &  &  &  &  &  &  &  &  \footnotesize Python & \footnotesize CMPy \cite{guns2019increasing} \\
    
    \footnotesize{\cite{regin_et_al:LIPIcs.CP.2024.25}} &  &  &\footnotesize \checkmark&  &  &  &  &  &  &  &  &\footnotesize \checkmark&  &  &  &  &  &  &  &  &  &  &  &  &  &  \\
    
    \footnotesize{\cite{huang2025orlmcustomizableframeworktraining}} &  &  &\footnotesize \checkmark&\footnotesize \checkmark&  &  &  &  &  &  &  &\footnotesize \checkmark&\footnotesize \checkmark&  &\footnotesize \checkmark&  &  &  &  &  &\footnotesize \checkmark&\footnotesize \checkmark&  &  & \footnotesize Python & \footnotesize coptpy \\
    
    \footnotesize{\cite{10738100}} &  &  &\footnotesize \checkmark&\footnotesize \checkmark&  &  &  &  &  &  &  &  &  &  &\footnotesize \checkmark&  &  &  &  &  &  &  &  &  & \footnotesize Python & \footnotesize CVXPY~\cite{10.5555/2946645.3007036} \\
    
    \footnotesize{\cite{Khan_2024}} &  &  &  &\footnotesize \checkmark&\footnotesize \checkmark&  &  &  &  &  &  &  &  &  &  &  &  &  &  &  &  &  &  &  & \footnotesize Python &  \\
    
    \footnotesize{\cite{a17120582}} &  &  &  &  &\footnotesize \checkmark&\footnotesize \checkmark&  &\footnotesize \checkmark&  &  &  &  &  &  &&  &  &  &  &  &\footnotesize \checkmark&  &  &  &  &  \\
    
    \footnotesize{\cite{reinhart-2024,hu2024scalableaccurategraphreasoning}} &  &  &  &  &\footnotesize \checkmark&  &  &\footnotesize \checkmark&  &  &  &  &  &  &  &  &  &  &  &  &  &  &  &  &  &  \\
    
    \footnotesize{\cite{make6030093}} &  &  &  &  &\footnotesize \checkmark&  &  &  &\footnotesize \checkmark&\footnotesize \checkmark&\footnotesize \checkmark&  &  &  &  &  &  &  &  &  &  &  &  &  &  &  \\
    
    \footnotesize{\cite{10720437}} &  &  &  &  &  &\footnotesize \checkmark&  &  &  &  &  &  &  &  &\footnotesize \checkmark&  &  &  &  &  &  &  &  &  &  &  \\
    
    \footnotesize{\cite{10704489}} &  &  &  &  &\footnotesize \checkmark&  &  &  &  &\footnotesize \checkmark&  &  &  &  &  &  &  &  &  &  &  &  &  &  &  &  \\
    
    \footnotesize{\cite{10628050,Wang_2024,jiang2024largelanguagemodelscombinatorial,bohnet2024exploringbenchmarkingplanningcapabilities,10675146,yang2024large,huang2024multimodal,SAKA2024100300,chin2024learning,buildings13071772,huang2024exploringtruepotentialevaluating,guo2024optimizinglargelanguagemodels}} &  &  &  &  &\footnotesize \checkmark&  &  &  &  &  &  &  &  &  &  &  &  &  &  &  &  &  &  &  &  &  \\
    
    \footnotesize{\cite{DBLP:conf/esann/MartinekLG24}} &  &  &  &  &  &\footnotesize \checkmark&  &  &  &  &  &  &  &  &  &  &  &  &\footnotesize \checkmark&  &  &  &\footnotesize \checkmark &  & \footnotesize Python & \footnotesize  Mealpy \cite{van2023mealpy} \\
    
    \footnotesize{\cite{lawless2024llmscoldstartcuttingplane}} &  &  &  &  &  &\footnotesize \checkmark&  &  &  &  &  &  &  &  &\footnotesize \checkmark&  &  &  &  &  &  &  &  &  &  &  \\
    
    \footnotesize{\cite{10.1007/978-3-031-70068-2_12,yatong2024tseohedgeservertask,chen2024uberuncertaintybasedevolutionlarge,liu2024llm4adplatformalgorithmdesign,yu2024autornetautomaticallyoptimizingheuristics,yao2024multiobjectiveevolutionheuristicusing}} &  &  &  &\footnotesize \checkmark&  &  &  &  &  &  &  &  &  &  &  &\footnotesize \checkmark&  &  &  &  &  &  &  &  & \footnotesize Python &  \\
    
    \footnotesize{\cite{sun2024autosatautomaticallyoptimizesat}} &  &  &  &\footnotesize \checkmark&  &  &  &  &  &  &  &  &  &  &  &\footnotesize \checkmark&  &  &  &  &  &  &  &  & \footnotesize Cpp &  \\
    
    \footnotesize{\cite{li2024foundationmodelsmixedinteger}} &  &  &  &\footnotesize \checkmark&  &  &  &  &  &  &  &  &  &  &\footnotesize \checkmark&  &  &  &  &  &  &  &  &  & \footnotesize Python &  \\
    
    \footnotesize{\cite{vanstein2024intheloophyperparameteroptimizationllmbased}} &  &  &  &\footnotesize \checkmark&  &  &  &  &  &  &  &  &  &  &  &  &  &\footnotesize \checkmark&  &  &  &  &  &  & \footnotesize Python &  \\
    
    \footnotesize{\cite{borazjanizadeh2024navigatinglabyrinthevaluatingenhancing}} &  &  &  &\footnotesize \checkmark&\footnotesize \checkmark&  &  &  &  &&  &  &  &  &  &  &  &  &  &  &  &  &  &  & \footnotesize Python &  \\
    
    \footnotesize{\cite{10.24963/ijcai.2024/579}} &  &  &  &  &  &  &\footnotesize \checkmark&  &  &  &  &  &  &  &  &  &  &  &  &  &  &  &  &  &  &  \\
    
    \footnotesize{\cite{gangwar2023highlighting,abdullin-etal-2023-synthetic}} &  &  &\footnotesize \checkmark&  &  &  &  &  &  &  &  &  &\footnotesize \checkmark&  &  &  &  &  &  &  &  &  &  &  &  &  \\
    
    \footnotesize{\cite{xiao2024chainofexperts}} &  &  &\footnotesize \checkmark&\footnotesize \checkmark&  &  &  &  &  &  &\footnotesize \checkmark&  &\footnotesize \checkmark&  &\footnotesize \checkmark&  &  &  &  &  &  &  &  &  & \footnotesize Python & \footnotesize Gurobi~\cite{gurobi} (gurobipy) \\
    
    \footnotesize{\cite{lawless2024i}} &  &  &\footnotesize \checkmark&\footnotesize \checkmark&  &  &  &  &  &  &  &  &\footnotesize \checkmark&  &  &  &  &  &  &  &  &  &  &  & \footnotesize Python &  \\
    
    \footnotesize{\cite{Romera-Paredes2024,liu2023algorithm}} &  &  &  &\footnotesize \checkmark&  &  &  &  &  &  &  &  &  &  &  &\footnotesize \checkmark&  &  &  &  &  &  &  &  & \footnotesize Python &  \\
    
    \footnotesize{\cite{ye2024large}} &  &  &  &\footnotesize \checkmark&  &  &  &  &  &  &  &  &  &  &  &  &\footnotesize \checkmark&  &  &  &  &  &  &  & \footnotesize Python &  \\
    
    \footnotesize{\cite{li2023large}} &  &  &  &\footnotesize \checkmark&  &  &  &  &  &  &  &  &  &  &\footnotesize \checkmark&  &  &  &  &  &  &  &  &  & \footnotesize Python & \footnotesize Gurobi~\cite{gurobi} (gurobipy) \\
    
    \footnotesize{\cite{mao2024identify}} &  &  &  &\footnotesize \checkmark&  &  &  &  &  &  &  &  &  &  &  &  &  &\footnotesize \checkmark&  &  &  &  &  &  & \footnotesize Python &  \\
    
    \footnotesize{\cite{almonacid2023automatic}} &  &  &  &\footnotesize \checkmark&  &  &  &  &  &  &  &  &  &  &\footnotesize \checkmark&  &  &  &  &  &  &  &  &  & \footnotesize MiniZinc &  \\
    
    \footnotesize{\cite{liu2024evolution}} &  &  &  &\footnotesize \checkmark&  &  &  &  &  &  &  &  &  &  &  &  &\footnotesize \checkmark&  &  &  &  &  &  &  & \footnotesize Python &  \\
    
    \footnotesize{\cite{amarasinghe2023aicopilot}} &  &  &  &\footnotesize \checkmark&  &  &  &  &  &  &  &\footnotesize \checkmark&  &  &  &  &  &  &  &  &  &  &  &  & \footnotesize Python & \footnotesize CMPy \cite{guns2019increasing} \\
    
    \footnotesize{\cite{voboril2025generatingstreamliningconstraintslarge}} &  &  &  &\footnotesize \checkmark&  &  &  &  &  &  &  &\footnotesize \checkmark&  &  &  &  &  &  &  &  &  &  &  &  & \footnotesize Python & \footnotesize MiniZinc \\
    
    \footnotesize{\cite{RePEc:spr:snopef:v:4:y:2023:i:4:d:10.1007_s43069-023-00277-6}} &  &  &  &  &\footnotesize \checkmark&  &  &  &  &  &  &  &  &  &\footnotesize \checkmark&  &  &  &  &  &  &  &  &  & \footnotesize Python & \footnotesize CPLEX~\cite{IBMsched2017} (CVXPY~\cite{10.5555/2946645.3007036}) \\
    
    \footnotesize{\cite{liu2024large1}} &  &  &  &  &\footnotesize \checkmark&  &  &  &  &  &  &  &  &  &  &  &\footnotesize \checkmark&  &  &  &  &  &  &  &  &  \\
    
    \footnotesize{\cite{liu2024large}} &  &  &  &  &\footnotesize \checkmark&  &  &  &  &  &  &  &  &  &  &  &  &\footnotesize \checkmark&  &  &\footnotesize \checkmark&  &  &  & \footnotesize Python & \footnotesize PyMoo~\cite{9078759} \\
    
    \footnotesize{\cite{teukam2024integrating}} &  &  &  &  &\footnotesize \checkmark&  &  &  &  &  &  &  &  &  &  &  &  &  &\footnotesize \checkmark&  &  &  &  &  & \footnotesize Python & \footnotesize GT4SD~\cite{Manica2023} \\
    
    \footnotesize{\cite{10.1007/978-981-97-2259-4_3}} &  &  &  &  &  &  &  &\footnotesize \checkmark&  &  &  &  &  &  &  &  &  &  &  &  &  &  &  &  &  &  \\
    
    \footnotesize{\cite{sartori2024large}} &  &  &  &  &  &  &  &  &\footnotesize \checkmark&  &  &  &  &  &  &  &  &  &  &  &  &  &  &  & \footnotesize R/Web Int. &  \\
    
    \footnotesize{\cite{chen2023diagnosing}} &  &  &  &  &  &  &  &  &  &  &\footnotesize \checkmark&  &  &  &\footnotesize \checkmark&  &  &  &  &  &  &  &  &  & \footnotesize Python & \footnotesize Gurobi~\cite{gurobi} (Pyomo~\cite{Hart2011})  \\
\end{longtable}

\clearpage

\section{Classification of Studies by LLM Architecture}
\label{app:llms}

\new{This appendix provides the analysis of the retrieved studies in terms of \glspl{llm} (\Cref{tab:llms}). We categorize each \gls{llm} by its architecture, the tasks researchers employed it for during optimization, its release date, evaluation metrics, and corresponding studies. The table also highlights the access type (\texttt{F/P}, indicating free or paid) and source availability (\texttt{O/C}, indicating open or closed). Overlapping naming conventions often create confusion, as the same term can refer to both a general architecture and specific models. Additionally, it is sometimes unclear whether researchers fine-tuned a model directly or used a version adapted for conversational purposes. When researchers did not specify the exact model in their studies, we refer to the base architecture and denote it with the keyword \texttt{Family}, as is commonly done with \texttt{GPT-4}.}

\begin{longtable}{>{\arraybackslash}m{2cm}>{\arraybackslash}m{3.2cm}>{\arraybackslash}m{0.8cm}>{\arraybackslash}m{2.9cm}>{\centering\arraybackslash}m{0.3cm}>{\centering\arraybackslash}m{0.3cm}>{\arraybackslash}m{3.1cm}>{\arraybackslash}m{1cm}}
    \caption{\new{Classification of studies by the role of \numllm \glspl{llm} in combinatorial optimization tasks. The \texttt{F/P} column indicates free or paid access type, while \texttt{O/C} denotes open or closed source availability.}} \label{tab:llms} \\
    \toprule
    \footnotesize{\textbf{Architecture}} & \footnotesize{\textbf{LLM}} & \footnotesize{\textbf{Date}} & \footnotesize{\textbf{Task}} & \footnotesize{\textbf{F/P}} & \footnotesize{\textbf{O/C}} & \footnotesize{\textbf{Metrics}} & \footnotesize{\textbf{Studies}} \\
    \midrule
    \endfirsthead
    \caption[]{\new{Classification of studies by the role of \numllm \glspl{llm} in combinatorial optimization tasks. The \texttt{F/P} column indicates free or paid access type, while \texttt{O/C} denotes open or closed source availability} (continued).} \\
    \toprule
    \footnotesize{\textbf{Architecture}} & \footnotesize{\textbf{LLM}} & \footnotesize{\textbf{Date}} & \footnotesize{\textbf{Task}} & \footnotesize{\textbf{F/P}} & \footnotesize{\textbf{O/C}} & \footnotesize{\textbf{Metrics}} & \footnotesize{\textbf{Studies}} \\
    \midrule
    \endhead
    \midrule
    \multicolumn{7}{r}{\textit{\footnotesize Continued on next page}} \\
    \endfoot
    \bottomrule
    \endlastfoot
    
    \multirow{3}{*}{\footnotesize\llm{T5} \footnotesize{\cite{raffel2023exploringlimitstransferlearning}}} 
    & \footnotesize\llm{T5-Base} & \footnotesize 10/2019 & \footnotesize{Problem Modeling, Solution Method} & \footnotesize F & \footnotesize O & \footnotesize / & \footnotesize \cite{chin2024learning, he2022linear} \\
    \cmidrule{2-8}
    & \footnotesize \llm{CodeT5-fine\-tuned\_CodeRL} \footnotesize{\new{\cite{le2022coderlmasteringcodegeneration}}} & \footnotesize \new{07/2022} & \footnotesize{Problem Modeling, Solution Method} & \footnotesize F & \footnotesize O & \footnotesize Training Loss, Success Rate, Correctness & \footnotesize \cite{amarasinghe2023aicopilot} \\
    \midrule

    \multirow{2}{*}{\footnotesize\llm{BART} \footnotesize{\new{\cite{lewis-etal-2020-bart}}}} 
    & \footnotesize \llm{BART-Base} & \footnotesize \new{10/2019} & \footnotesize{Problem Modeling} & \footnotesize F & \footnotesize O & \footnotesize Accuracy & \footnotesize \cite{gangwar2023highlighting, ramamonjison-etal-2022-augmenting} \\
    \cmidrule{2-8}
    & \footnotesize \llm{BART-Large} & \footnotesize \new{10/2019} & \footnotesize{Problem Modeling} & \footnotesize F & \footnotesize O & \footnotesize Accuracy & \footnotesize \cite{gangwar2023highlighting, jang2022tag} \\
    \midrule

    \multirow{1}{*}{\footnotesize \new{\llm{UnixCoder}} \footnotesize{\cite{guo2022unixcoder}}} 
    & \footnotesize \new{\llm{UnixCoder}} & \footnotesize \new{03/2022} & \footnotesize \new{Solution Method} & \footnotesize \new{F} & \footnotesize \new{O} & \footnotesize \new{PAR10} & \footnotesize \new{\cite{10.24963/ijcai.2024/579}} \\
    \midrule

    \multirow{14}{*}{\footnotesize\llm{GPT-3.5} \footnotesize{\cite{NEURIPS2020_1457c0d6}}} 
    & \footnotesize \llm{Text-Davinci-Edit-001} & \footnotesize \new{07/2022} & \footnotesize Solution Method & \footnotesize P & \footnotesize C & \footnotesize / & \footnotesize \cite{almonacid2023automatic} \\
    \cmidrule{2-8}
    & \footnotesize \llm{Text-Davinci-003} & \footnotesize \new{11/2022} & \footnotesize Problem Modeling, Solution Method & \footnotesize P & \footnotesize C & \footnotesize Accuracy & \footnotesize \cite{li2023large, lawless2024i, almonacid2023automatic} \\
    \cmidrule{2-8}
    & \footnotesize \llm{GPT-3.5 (Family)} & \footnotesize 11/2022 & \footnotesize Problem Modeling, Solution Method\new{, Benchmarking, Validation} & \footnotesize P & \footnotesize C & \footnotesize \new{\# API Call Rate, API Mismatching Rate, Error Raise Rate, Throughput, Average Travel Time, GAP}
 & \footnotesize \new{\cite{ahmaditeshnizi2024optimus, Da2024, zhang-etal-2024-solving, 10.1007/978-3-031-70068-2_12, borazjanizadeh2024navigatinglabyrinthevaluatingenhancing}} \\
    \cmidrule{2-8}
    & \footnotesize \llm{ChatGPT 3.5} & \footnotesize 11/2022 & \footnotesize Problem Modeling & \footnotesize F & \footnotesize C & \footnotesize Accuracy & \footnotesize \cite{li2023synthesizing} \\
    \cmidrule{2-8}
    & \footnotesize \llm{ChatGPT 3.5-Turbo} & \footnotesize 11/2022 & \footnotesize Problem Modeling, Solution Method & \footnotesize P & \footnotesize C & \footnotesize / & \footnotesize \cite{yang2024large, liu2023algorithm, liu2024large} \\
    \cmidrule{2-8}
    & \footnotesize \llm{GPT-3.5-turbo} & \footnotesize 11/2022 & \footnotesize Problem Modeling, Solution Method, \new{Validation} & \footnotesize P & \footnotesize C & \footnotesize Accuracy, Hypervolume, Inverted Generational Distance, Compile Error Rate, Runtime Error Rate & \footnotesize \new{\cite{xiao2024chainofexperts, liu2024evolution, ye2024large, yang2024optibenchmeetsresocraticmeasure, yatong2024tseohedgeservertask, liu2024llm4adplatformalgorithmdesign, yao2024multiobjectiveevolutionheuristicusing}} \\
    \cmidrule{2-8}
    & \footnotesize \llm{GPT-3.5-turbo-0613} & \footnotesize 06/2023 & \footnotesize Problem Modeling, Solution Method & \footnotesize P & \footnotesize C & \footnotesize F1-Score, ANC, Rank\new{, Uncertainty, Policy Metric, Goal Metric} & \footnotesize \new{\cite{liu2024large1, ahmed2024lm4opt, mao2024identify, guo2024optimizinglargelanguagemodels}} \\
    \cmidrule{2-8}
    & \footnotesize \new{\llm{GPT-3.5-Turbo-1106}} & \footnotesize \new{11/2023} & \footnotesize \new{Solution Method} & \footnotesize \new{P} & \footnotesize \new{C} & \footnotesize \new{Correctness, Output Format Consistency} & \new{\footnotesize \cite{huang2024exploringtruepotentialevaluating}} \\

    \multirow{24}{*}{\footnotesize \llm{GPT-4} \footnotesize\cite{openai2024gpt4}} 
    & \footnotesize \llm{GPT-4 (Family)} & \footnotesize 03/2023 & \footnotesize Problem Modeling, Solution Method, \new{Benchmarking, Validation} & \footnotesize P & \footnotesize C & \footnotesize Accuracy, Feasibility, Optimality, Efficiency, ROUGE \cite{lin-2004-rouge}, BERTScore \cite{zhang2020bertscoreevaluatingtextgeneration}, \new{Completeness Score, Homogeneity Score, No API Call Rate, API Mismatching Rate, Error Raise Rate, Throughput, Average Travel Time, Code Compilation Success, Debugging Success Rate, and Code Generation Efficiency} & \footnotesize \new{\cite{yang2024large, li2023large, lawless2024i, huang2024words,  abdullin-etal-2023-synthetic, chen2023diagnosing, ahmaditeshnizi2024optimus, smartcities7050094, Da2024, 10.1145/3664646.3665084, 10738100, yang2024optibenchmeetsresocraticmeasure, zhang-etal-2024-solving, delarosa2024trippaltravelplanningguarantees, 10.1007/978-3-031-70068-2_12, borazjanizadeh2024navigatinglabyrinthevaluatingenhancing}} \\
    \cmidrule{2-8}
    & \footnotesize \llm{ChatGPT 4} & \footnotesize 03/2023 & \footnotesize \new{Problem Modeling}, Solution Method, Benchmarking & \footnotesize P & \footnotesize C & \footnotesize Macro-F1 Score, Time & \footnotesize \new{\cite{RePEc:spr:snopef:v:4:y:2023:i:4:d:10.1007_s43069-023-00277-6, 10.1007/978-981-97-2259-4_3, buildings13071772, ALIPOURVAEZI2024110574}} \\
    \cmidrule{2-8}
    & \footnotesize \llm{GPT-4-0613} & \footnotesize 06/2023 & \footnotesize Problem Modeling\new{, Solution Method} & \footnotesize P & \footnotesize C & \footnotesize F1-Score\new{, Correctness, Output Format Consistency} & \footnotesize \cite{ahmed2024lm4opt, ai5010006, huang2024exploringtruepotentialevaluating} \\
    \cmidrule{2-8}
    & \footnotesize \llm{GPT-4-Turbo} & \footnotesize 11/2023 & \footnotesize \new{Problem Modeling, }Solution Method, Benchmarking & \footnotesize P & \footnotesize C & \footnotesize Score & \new{\footnotesize \cite{sartori2024large, liu2023algorithm, ye2024large, 10.1145/3638530.3664181, Wang_2024, Khan_2024, yu2024autornetautomaticallyoptimizingheuristics}} \\
    \cmidrule{2-8}
    & \footnotesize \llm{GPT-4-Vision-Preview} & \footnotesize 12/2023 & \footnotesize Solution Method & \footnotesize P & \footnotesize C & / & \footnotesize \cite{huang2024multimodal} \\
    \cmidrule{2-8}
    & \footnotesize \new{\llm{GPT-4o}} & \footnotesize \new{05/2024} & \footnotesize \new{Problem Modeling, Solution Method, Validation} & \footnotesize \new{P} & \footnotesize \new{C} & \footnotesize \new{Execution Rate, Solving Accuracy, Average Solving Times} & \footnotesize \cite{10675146, jiang2025llmopt, Khan_2024, voboril2025generatingstreamliningconstraintslarge, 10818476, hao2025planningrigorgeneralpurposezeroshot, li2024foundationmodelsmixedinteger, ahmaditeshnizi2024optimus03usinglargelanguage} \\
    \cmidrule{2-8}
    & \footnotesize \new{\llm{ChatGPT-4o}} & \footnotesize \new{05/2024} & \footnotesize \new{Problem Modeling, Solution Method, Benchmarking, Validation} & \footnotesize \new{P} & \footnotesize \new{C} & \footnotesize \new{Accuracy, Consistency, Stability, Solution Consistency, Constraints Robustness, Runtime Efficiency} & \footnotesize \new{\cite{a17120582, make6030093}} \\
    \cmidrule{2-8}
    & \footnotesize \new{\llm{GPT-4o-2024-05-13}} & \footnotesize \new{05/2024} & \footnotesize \new{Solution Method} & \footnotesize \new{P} & \footnotesize \new{C} & \footnotesize \new{/} & \footnotesize \new{\cite{vanstein2024intheloophyperparameteroptimizationllmbased}} \\
    \cmidrule{2-8}
    & \footnotesize \new{\llm{ChatGPT-4-Turbo}} & \footnotesize \new{07/2024} & \footnotesize \new{Problem Modeling, Solution Method} & \footnotesize \new{P} & \footnotesize \new{C} & \footnotesize \new{/} & \footnotesize \new{\cite{hu2024scalableaccurategraphreasoning}} \\
    \pagebreak
    & \footnotesize \new{\llm{GPT-4o-mini}} & \footnotesize \new{07/2024} & \footnotesize \new{Solution Method} & \footnotesize \new{P} & \footnotesize \new{C} & \footnotesize \new{/} & \footnotesize \new{\cite{Wang_2024, hu2024scalableaccurategraphreasoning, li2025graphteamfacilitatinglargelanguage, liu2024llm4adplatformalgorithmdesign}} \\
    \cmidrule{2-8}
    & \footnotesize \new{\llm{ChatGPT-4o-mini}} & \footnotesize \new{07/2024} & \footnotesize \new{Problem Modeling, Solution Method} & \footnotesize \new{P} & \footnotesize \new{C} & \footnotesize \new{/} & \footnotesize \new{\cite{hu2024scalableaccurategraphreasoning}} \\
    \cmidrule{2-8}
    & \footnotesize \new{\llm{GPT-4o-2024-08-06}} & \footnotesize \new{08/2024} & \footnotesize \new{Solution Method} & \footnotesize \new{P} & \footnotesize \new{C} & \footnotesize \new{Solved Instances, Penalized Average Runtime} & \footnotesize \new{\cite{sun2024autosatautomaticallyoptimizesat}} \\
    \midrule

    \multirow{6}{*}{\footnotesize \llm{PaLM 2} \footnotesize{\cite{anil2023palm}}} 
    & \footnotesize \llm{Bard} & \footnotesize \new{05/2023} & \footnotesize{Problem Modeling} & \footnotesize F & \footnotesize C & \footnotesize Accuracy & \footnotesize \cite{li2023synthesizing} \\
    \cmidrule{2-8}
    & \footnotesize \llm{Codey} & \footnotesize \new{05/2023} & \footnotesize{Solution Method} & \footnotesize P & \footnotesize C & \footnotesize / & \footnotesize \cite{Romera-Paredes2024} \\
    \cmidrule{2-8}
    & \footnotesize \llm{PaLM 2-L} & \footnotesize \new{05/2023} & \footnotesize{Solution Method} & \footnotesize P & \footnotesize C & \footnotesize Accuracy & \footnotesize \cite{yang2024large} \\
    \cmidrule{2-8}
    & \footnotesize \llm{PaLM 2-L-IT} & \footnotesize \new{05/2023} & \footnotesize{Solution Method} & \footnotesize P & \footnotesize C & \footnotesize Accuracy & \footnotesize \cite{yang2024large} \\
    \cmidrule{2-8}
    & \footnotesize \llm{Text-Bison} & \footnotesize \new{05/2023} & \footnotesize{Solution Method} & \footnotesize P & \footnotesize C & \footnotesize Accuracy & \footnotesize \cite{yang2024large} \\
    \midrule

    \multirow{17}{*}{\footnotesize\llm{LLaMa 2} \footnotesize{\cite{touvron2023llama2}}} 
    & \footnotesize \llm{LLaMa 2 (Family)} & \footnotesize 07/2023 & \footnotesize{Solution Method} & \footnotesize F & \footnotesize O & \footnotesize Accuracy & \footnotesize \cite{chin2024learning} \\
    \cmidrule{2-8}
    & \footnotesize \llm{LLaMa 2-7b} & \footnotesize 07/2023 & \footnotesize Problem Modeling\new{, Solution Method, Benchmarking, Validation} & \footnotesize F & \footnotesize O & \footnotesize F1-Score\new{, Throughput, Average Travel Time, No API Call Rate, API Mismatching Rate, Error Raise Rate} & \footnotesize \new{\cite{ahmed2024lm4opt, Da2024}} \\
    \cmidrule{2-8}
    & \footnotesize \new{\llm{LLaMa 2-13b}} & \footnotesize \new{07/2023} & \footnotesize \new{Problem Modeling, Solution Method, Benchmarking, Validation} & \footnotesize \new{F} & \footnotesize \new{O} & \footnotesize \new{Throughput, Average Travel Time, No API Call Rate, API Mismatching Rate, Error Raise Rate} & \footnotesize \new{\cite{Da2024}} \\
    \cmidrule{2-8}
    & \footnotesize \new{\llm{LLaMa 2-13b-Chat}} & \footnotesize \new{07/2023} & \footnotesize \new{Solution Method} & \footnotesize \new{F} & \footnotesize \new{O} & \footnotesize \new{Correctness, Output Format Consistency} & \footnotesize \new{\cite{sun2024autosatautomaticallyoptimizesat}} \\
    \cmidrule{2-8}
    & \footnotesize \llm{CodeLlama} \footnotesize{\cite{roziere2024codellamaopenfoundation}} & \footnotesize \new{08/2023} & \footnotesize{Solution Method} & \footnotesize F & \footnotesize O & \footnotesize / & \footnotesize \cite{liu2024evolution, 10.1007/978-3-031-70068-2_12, borazjanizadeh2024navigatinglabyrinthevaluatingenhancing} \\
    \cmidrule{2-8}
    & \footnotesize \new{\llm{CodeLlama-Instruct}} \new{\footnotesize{\cite{roziere2024codellamaopenfoundation}}} & \footnotesize \new{08/2023} & \footnotesize \new{Problem Modeling, Solution Method, Validation} & \footnotesize \new{F} & \footnotesize \new{O} & \footnotesize \new{Accuracy, Feasibility, Efficiency} & \footnotesize \new{\cite{Mostajabdaveh04112024, borazjanizadeh2024navigatinglabyrinthevaluatingenhancing}} \\
    \cmidrule{2-8}
    & \footnotesize \llm{Tulu-v2-dpo-7b} \footnotesize{\cite{ivison2023camelschangingclimateenhancing}} & \footnotesize \new{11/2023} & \footnotesize{Benchmarking} & \footnotesize F & \footnotesize O & \footnotesize Score & \footnotesize \cite{sartori2024large} \\
    \midrule

    \multirow{5}{*}{\footnotesize \new{\llm{Mistral}} \footnotesize{\cite{jiang2023mistral}}} 
    & \footnotesize \new{\llm{Mistral-7B}} & \footnotesize \new{09/2023} & \footnotesize \new{Problem Modeling, Solution Method} & \footnotesize \new{F} & \footnotesize \new{O} & \footnotesize \new{Accuracy, Pass@K} & \footnotesize \new{\cite{huang2025orlmcustomizableframeworktraining}} \\
    \cmidrule{2-8}
    & \footnotesize \new{\llm{Zephyr-7B-beta}} \new{\footnotesize{\cite{tunstall2023zephyrdirectdistillationlm}}} & \footnotesize \new{10/2023} & \footnotesize \new{Problem Modeling, Solution Method, Validation} & \footnotesize \new{F} & \footnotesize \new{O} & \footnotesize / & \footnotesize \new{\cite{Mostajabdaveh04112024}} \\
    \pagebreak
    & \footnotesize \new{\llm{Le Chat}} & \footnotesize \new{07/2024} & \footnotesize \new{Problem Modeling} & \footnotesize \new{P} & \footnotesize \new{C} & \footnotesize / & \footnotesize \cite{DBLP:conf/esann/MartinekLG24} \\

    \midrule

    \multirow{4}{*}{\footnotesize \new{\llm{Qwen}} \new{\footnotesize \cite{bai2023qwentechnicalreport}}} 
    & \footnotesize \new{\llm{Qwen1.5-14B}} & \footnotesize \new{10/2023} & \footnotesize \new{Problem Modeling, Solution Method} & \footnotesize \new{F} & \footnotesize \new{O} & \footnotesize \new{Execution Rate, Solving Accuracy, Average Solving Times} & \footnotesize \new{\cite{jiang2025llmopt}} \\
    \cmidrule{2-8}
    & \footnotesize \new{\llm{Qwen-Turbo}} & \footnotesize \new{08/2024} & \footnotesize \new{Solution Method}  & \footnotesize \new{F} & \footnotesize \new{O} & \footnotesize \new{/} & \footnotesize \new{\cite{liu2024llm4adplatformalgorithmdesign}} \\
    \cmidrule{2-8}
    & \footnotesize \new{\llm{Qwen (LoRA Fine-Tuned)}} & \footnotesize \new{08/2024} & \footnotesize \new{Problem Modeling, Solution Method} & \footnotesize \new{P} & \footnotesize \new{C} & \footnotesize \new{/} & \footnotesize \new{\cite{zhang-etal-2024-solving}} \\
    \midrule

    \multirow{6}{*}{\footnotesize \new{\llm{Gemini}} \cite{geminiteam2024geminifamilyhighlycapable}}  
    & \footnotesize \llm{Gemini 1.0 Pro} & \footnotesize 12/2023 & \footnotesize{Problem Modeling, Solution Method, \new{Validation}} & \footnotesize P & \footnotesize C & \footnotesize Feasibility, Optimality, Efficiency, F1-Score & \footnotesize \cite{huang2024words, liu2024evolution} \\
    \cmidrule{2-8}
    & \footnotesize \new{\llm{Gemini 1.5 Pro}} & \footnotesize \new{02/2024} & \footnotesize \new{Problem Modeling} & \footnotesize \new{P} & \footnotesize \new{C} & \footnotesize / & \footnotesize \new{\cite{bohnet2024exploringbenchmarkingplanningcapabilities}} \\
    \cmidrule{2-8}
    & \footnotesize \new{\llm{Gemini 1.5 Flash}} & \footnotesize \new{02/2024} & \footnotesize \new{Problem Modeling} & \footnotesize \new{P} & \footnotesize \new{C} & \footnotesize / & \footnotesize \new{\cite{bohnet2024exploringbenchmarkingplanningcapabilities}}\\
    \cmidrule{2-8}
    & \footnotesize \new{\llm{Gemini 2.0 Flash}} & \footnotesize \new{08/2024} & \footnotesize \new{Problem Modeling} & \footnotesize \new{P} & \footnotesize \new{C} & \footnotesize / & \new{\footnotesize \cite{DBLP:conf/esann/MartinekLG24}} \\
    \cmidrule{2-8}
    & \footnotesize \new{\llm{Gemini 2.0 Pro}} & \footnotesize \new{08/2024} & \footnotesize \new{Solution Method} & \footnotesize \new{P} & \footnotesize \new{C} & \footnotesize / & \new{\footnotesize \cite{huang2024exploringtruepotentialevaluating}} \\
    \midrule

    \footnotesize \multirow{3}{*}{\llm{\new{Mixtral} \footnotesize{\cite{jiang2024mixtral}}}}  
    & \footnotesize \llm{Mixtral-8x7b-instruct-v0.1} & \footnotesize \new{01/2024} & \footnotesize{Benchmarking} & \footnotesize P & \footnotesize C & \footnotesize Score & \footnotesize \cite{sartori2024large} \\
    \cmidrule{2-8}
    & \footnotesize \llm{\new{Mixtral-8x22B}} & \footnotesize \new{07/2024} & \footnotesize \new{Problem Modeling, Solution Method} & \footnotesize \new{P} & \footnotesize \new{C} & \footnotesize / & \footnotesize \new{\cite{10818476}} \\
    \midrule

    \multirow{8}{*}{\footnotesize \new{\llm{DeepSeek}} \footnotesize{\cite{deepseekai2024deepseekllmscalingopensource}}}
    & \footnotesize \llm{DeepSeek-LLM-7B-Base} & \footnotesize \new{01/2024} & \footnotesize{Solution Method} & \footnotesize P & \footnotesize C & \footnotesize / & \footnotesize \cite{liu2024evolution} \\
    \cmidrule{2-8}
    & \footnotesize \llm{\new{DeepSeek-Math-7B}} \cite{shao2024deepseekmathpushinglimitsmathematical} & \footnotesize \new{06/2024} & \footnotesize \new{Problem Modeling, Solution Method} & \footnotesize \new{P} & \footnotesize \new{C} & \footnotesize \new{Accuracy, Pass@K} & \footnotesize \new{\cite{huang2025orlmcustomizableframeworktraining}} \\
    \cmidrule{2-8}
    & \footnotesize \new{\llm{DeepSeek-Coder-33B}} & \footnotesize \new{07/2024} & \footnotesize \new{Solution Method} & \footnotesize \new{P} & \footnotesize \new{C} & \footnotesize \new{/} & \footnotesize \new{\cite{10.1007/978-3-031-70068-2_12, chen2024uberuncertaintybasedevolutionlarge}} \\
    \cmidrule{2-8}
    & \footnotesize \new{\llm{DeepSeek-V2} \cite{deepseekai2024deepseekv2strongeconomicalefficient}} & \footnotesize \new{08/2024} & \footnotesize \new{Problem Modeling, Solution Method} & \footnotesize \new{P} & \footnotesize \new{C} & \footnotesize / & \footnotesize \new{\cite{yang2024optibenchmeetsresocraticmeasure}} \\
    \cmidrule{2-8}
    & \footnotesize \new{\llm{DeepSeek-Coder-V2} \cite{deepseekai2024deepseekcoderv2breakingbarrierclosedsource}} & \footnotesize \new{08/2024} & \footnotesize \new{Solution Method} & \footnotesize \new{P} & \footnotesize \new{C} & \footnotesize / & \footnotesize \new{\cite{yatong2024tseohedgeservertask}} \\
    \midrule

    \multirow{1}{*}{\footnotesize \llm{\new{Claude 3.5}} \footnotesize{\new{\cite{anthropic2024claude35}}}}  
    & \footnotesize \llm{\new{Claude 3.5 Sonnet}} & \footnotesize \new{06/2024} & \footnotesize \new{Problem Modeling, Solution Method, Benchmarking, Validation} & \footnotesize \new{P} & \footnotesize \new{C} & \footnotesize \new{Accuracy, Solution Quality, \# Good Solutions, Convergence Rate, Instruction Adherence, Consistency, Stability, Prompt Sensitivity, Stochastic Variability, Constraints Robustness, Runtime Efficiency} & \footnotesize \new{\cite{reinhart-2024, voboril2025generatingstreamliningconstraintslarge, hao2025planningrigorgeneralpurposezeroshot}} \\
    \pagebreak
    & \footnotesize \llm{\new{Claude 3.5 Opus}} & \footnotesize \new{06/2024} & \footnotesize \new{Problem Modeling, Solution Method} & \footnotesize \new{P} & \footnotesize \new{C} & \footnotesize \new{/} & \footnotesize \new{\cite{10818476, 10.1007/978-3-031-70068-2_12}} \\
    \cmidrule{2-8}
    & \footnotesize \llm{\new{Claude 3.5 Haiku}} & \footnotesize \new{06/2024} & \footnotesize \new{Solution Method} & \footnotesize \new{P} & \footnotesize \new{C} & \footnotesize \new{/} & \footnotesize \new{\cite{liu2024llm4adplatformalgorithmdesign, liu2024llm4adplatformalgorithmdesign}}\\
    \cmidrule{2-8}
    & \footnotesize \llm{\new{Claude-3.5-Sonnet-20241022}} & \footnotesize \new{10/2024} & \footnotesize \new{Solution Method} & \footnotesize \new{P} & \footnotesize \new{C} & \footnotesize \new{/} & \footnotesize \new{\cite{jiang2024largelanguagemodelscombinatorial}} \\
    \midrule

    \multirow{1}{*}{\footnotesize \new{\llm{Cohere} \cite{cohere2024commandrplus}}} 
    & \footnotesize \new{\llm{Command-R+}} & \footnotesize \new{06/2024} & \footnotesize \new{Problem Modeling, Solution Method} & \footnotesize \new{P} & \footnotesize \new{C} & \footnotesize \new{/} & \footnotesize \new{\cite{10818476}} \\
    \midrule

    \multirow{6}{*}{\footnotesize \llm{LLaMa 3} \footnotesize{\cite{dubey2024llama3herdmodels}}}
    & \footnotesize \llm{LLaMa 3-70B} & \footnotesize \new{07}/2024 & \footnotesize \new{Problem Modeling,} Solution Method & \footnotesize F & \footnotesize O & / & \footnotesize \new{\cite{ye2024large, yang2024optibenchmeetsresocraticmeasure, ju-etal-2024-globe, ahmaditeshnizi2024optimus03usinglargelanguage}} \\
    \cmidrule{2-8}
    & \footnotesize \new{\llm{LLaMa 3-8B}} & \footnotesize \new{08/2024} & \footnotesize \new{Problem Modeling, Solution Method} & \footnotesize \new{F} & \footnotesize \new{O} & \footnotesize \new{Accuracy, Compilation Error Rate, Runtime Error Rate}  & \footnotesize \cite{huang2025orlmcustomizableframeworktraining} \\
    \cmidrule{2-8}
    & \footnotesize \new{\llm{LLaMa 3-70B-Instruct}} & \footnotesize \new{08/2024} & \footnotesize \new{Solution Method} & \footnotesize \new{F} & \footnotesize \new{O} & \footnotesize \new{Accuracy, Pass@K} & \footnotesize \cite{yatong2024tseohedgeservertask} \\
    \cmidrule{2-8}
    & \footnotesize \new{\llm{LLaMa 3.1-8B}} & \footnotesize \new{08/2024} & \footnotesize \new{Solution Method} & \footnotesize \new{F} & \footnotesize \new{O} & \footnotesize / & \footnotesize \cite{liu2024llm4adplatformalgorithmdesign} \\
    \midrule

    \footnotesize \multirow{1}{*}{\footnotesize \new{\llm{Qwen2}} \new{\footnotesize \cite{yang2024qwen2technicalreport}}} 
    & \footnotesize \new{\llm{Qwen2.5-7B}} & \footnotesize \new{07/2024} & \footnotesize \new{Problem Modeling, Solution Method} & \footnotesize \new{F} & \footnotesize \new{O} & \footnotesize \new{Accuracy, Pass@K} & \footnotesize \new{\cite{huang2025orlmcustomizableframeworktraining}} \\
    \midrule

    \footnotesize \llm{\new{StarCoder}} \footnotesize{\cite{li2023starcoder}} 
    & \footnotesize \new{\llm{StarCoder2} \cite{lozhkov2024starcoder2stackv2}} & \footnotesize \new{07/2024} & \footnotesize \new{Solution Method}  & \footnotesize \new{P} & \footnotesize \new{C} & \footnotesize \new{/} & \footnotesize \new{\cite{10.1007/978-3-031-70068-2_12}} \\
    \midrule

    \footnotesize \new{\llm{GLM} \footnotesize{\cite{du2022glm}}} 
    & \footnotesize \new{\llm{GLM-3-Turbo}} & \footnotesize \new{07/2024} & \footnotesize \new{Solution Method}  & \footnotesize \new{P} & \footnotesize \new{C} & \footnotesize \new{/} & \footnotesize \new{\cite{liu2024llm4adplatformalgorithmdesign, yatong2024tseohedgeservertask}} \\
    \midrule

    \footnotesize \new{\llm{OpenCoder} \footnotesize{\cite{huang2024opencoder}}} 
    & \footnotesize \new{\llm{OpenCoder-8B-Instruct}} & \footnotesize \new{07/2024} & \footnotesize \new{Solution Method} & \footnotesize \new{P} & \footnotesize \new{C} & \footnotesize \new{/} & \footnotesize \new{\cite{chen2024uberuncertaintybasedevolutionlarge}} \\
    \midrule

    \footnotesize \new{\llm{Yi}} \footnotesize{\cite{yi2024open}} 
    & \footnotesize \new{\llm{Yi-34B-Chat}} & \footnotesize \new{07/2024} & \footnotesize \new{Solution Method} & \footnotesize \new{P} & \footnotesize \new{C} & \footnotesize \new{/} & \footnotesize \new{\cite{liu2024llm4adplatformalgorithmdesign}} \\
    \midrule

    \footnotesize \new{\llm{Gemma} \footnotesize{\cite{gemmateam2024gemma2improvingopen}}}
    & \footnotesize \new{\llm{Gemma 2 27B}} & \footnotesize \new{07/2024} & \footnotesize \new{Problem Modeling} & \footnotesize \new{P} & \footnotesize \new{O} & \footnotesize / & \footnotesize \new{\cite{bohnet2024exploringbenchmarkingplanningcapabilities}} \\
    \midrule

    \footnotesize \new{\llm{InternLM2} \footnotesize{\cite{cai2024internlm2technicalreport}}}
    & \footnotesize \new{\llm{InternLM2-20B-Chat}} & \footnotesize \new{08/2024} & \footnotesize \new{Solution Method} & \footnotesize \new{P} & \footnotesize \new{C} & \footnotesize \new{Correctness, Output Format Consistency} & \footnotesize \new{\cite{huang2024exploringtruepotentialevaluating}} \\
    
\end{longtable}

\clearpage

\section{Classification of Studies by Benchmark Dataset}
\label{app:benchmark-datasets-analysis}

\new{This appendix provides the tabular analysis related to the classification of studies by dataset (\Cref{tab:benchmark-datasets-analysis}). Please refer to \Cref{sec:datasets} for further context.}

\begin{table}[htbp]
    \centering
    \caption{Classification of studies by benchmark dataset.}
    \label{tab:benchmark-datasets-analysis}
    \begin{tabular}{p{3.3cm}
    p{3.5cm}
    p{6.8cm}
    C{0.8cm}}
    \toprule
    \footnotesize \textbf{Name} & \footnotesize \textbf{Source} & \footnotesize \textbf{Studies}  & \footnotesize \textbf{\#}\\
    \midrule
    \footnotesize LPWP or NL4Opt & \footnotesize \citet{ramamonjison-etal-2022-augmenting} & \footnotesize \cite{ahmed2024lm4opt,li2023synthesizing,ahmaditeshnizi2024optimus,xiao2024chainofexperts,wang2023opdnl4opt,doan2022vtccnlp,ning2023novel,gangwar2023highlighting,he2022linear,abdullin-etal-2023-synthetic,jang2022tag,ramamonjison-etal-2022-augmenting,michailidis_et_al:LIPIcs.CP.2024.20,jiang2025llmopt,zhang-etal-2024-solving,huang2025orlmcustomizableframeworktraining} & \footnotesize \new{16}\\
    
    \footnotesize ComplexOR & \footnotesize\citet{xiao2024chainofexperts}  & \footnotesize\cite{xiao2024chainofexperts,ahmaditeshnizi2024optimus,jiang2025llmopt}   & \footnotesize \new{3}\\
    \footnotesize NLP4LP & \footnotesize\citet{ahmaditeshnizi2024optimus,ahmaditeshnizi2024optimus03usinglargelanguage} & \footnotesize\cite{ahmaditeshnizi2024optimus,jiang2025llmopt,ahmaditeshnizi2024optimus03usinglargelanguage} & \footnotesize \new{3} \\
    \footnotesize \new{IndustryOR} & \footnotesize \new{\citet{huang2025orlmcustomizableframeworktraining}} & \footnotesize \new{\cite{huang2025orlmcustomizableframeworktraining,jiang2025llmopt}} & \footnotesize \new{2} \\

    \footnotesize \new{Mamo} & \footnotesize \new{\citet{huang2025llmsmathematicalmodelingbridging}} & \footnotesize \new{\cite{jiang2025llmopt,huang2025orlmcustomizableframeworktraining}} & \footnotesize \new{2} \\

    \footnotesize \new{GraphInstruct} & \footnotesize \citet{luo2024graphinstructempoweringlargelanguage} & \footnotesize \cite{hu2024scalableaccurategraphreasoning,li2025graphteamfacilitatinglargelanguage} & \footnotesize \new{2} \\

    \footnotesize AI-copilot-data & \footnotesize\citet{amarasinghe2023aicopilot} & \footnotesize\cite{amarasinghe2023aicopilot}   & \footnotesize 1 \\

    \footnotesize Almonacid & \footnotesize\citet{almonacid2023automatic}  & \footnotesize\cite{almonacid2023automatic} & \footnotesize 1 \\

    \footnotesize Safeguard, Code Generation & \footnotesize\citet{lawless2024i}  & \footnotesize\cite{lawless2024i}  & \footnotesize 1 \\

    \footnotesize Huang et al. & \footnotesize\citet{huang2024words}  & \footnotesize\cite{huang2024words}  & \footnotesize 1 \\
    
    \footnotesize OptiChat & \footnotesize\citet{chen2023diagnosing} &  \footnotesize\cite{chen2023diagnosing}  & \footnotesize 1 \\
    
    \footnotesize \new{Michailidis et al.} & \new{\footnotesize \citet{michailidis_et_al:LIPIcs.CP.2024.20}} & \footnotesize\new{\cite{michailidis_et_al:LIPIcs.CP.2024.20}} & \footnotesize\new{1}\\

    \footnotesize \new{Optibench, ReScratic-29k} & \footnotesize\new{\citet{yang2024optibenchmeetsresocraticmeasure}} & \footnotesize\new{\cite{yang2024optibenchmeetsresocraticmeasure}} & \footnotesize\new{1}\\

    \footnotesize \new{Mostajabdaveh et al.} & \footnotesize \new{\citet{Mostajabdaveh04112024}} & \footnotesize \new{\cite{Mostajabdaveh04112024}} & \footnotesize \new{1} \\

    \footnotesize \new{Zhang et al.} & \footnotesize \citet{zhang-etal-2024-solving} & \footnotesize \cite{zhang-etal-2024-solving} & \footnotesize \new{1} \\

    \footnotesize \new{SearchBench} & \footnotesize \citet{borazjanizadeh2024navigatinglabyrinthevaluatingenhancing} & \footnotesize \cite{borazjanizadeh2024navigatinglabyrinthevaluatingenhancing} & \footnotesize \new{1}\\

    \footnotesize \new{Ju et al.} & \footnotesize \citet{ju-etal-2024-globe} & \footnotesize \cite{ju-etal-2024-globe} & \footnotesize \new{1} \\
    \footnotesize \new{Talk Like A Graph} & \footnotesize \citet{fatemi2023talklikegraphencoding} & \footnotesize\cite{li2025graphteamfacilitatinglargelanguage} & \footnotesize\new{1} \\
    \footnotesize \new{LLM4DyG} & \footnotesize \citet{zhang2024llm4dyglargelanguagemodels} & \footnotesize\cite{li2025graphteamfacilitatinglargelanguage} & \footnotesize\new{1} \\
    \footnotesize \new{GraphViz} & \footnotesize \citet{chen2024graphwizinstructionfollowinglanguagemodel} & \footnotesize\cite{li2025graphteamfacilitatinglargelanguage} & \footnotesize\new{1} \\
    \footnotesize \new{NLGraph} & \footnotesize \citet{wang2024languagemodelssolvegraph} & \footnotesize\cite{li2025graphteamfacilitatinglargelanguage} & \footnotesize\new{1} \\
    \footnotesize \new{GNN-AutoGL} & \footnotesize \citet{li2025graphteamfacilitatinglargelanguage} & \footnotesize\cite{li2025graphteamfacilitatinglargelanguage} & \footnotesize\new{1} \\
    \footnotesize \new{ORQUA} & \footnotesize \citet{mostajabdaveh2024evaluatingllmreasoningoperations} & \footnotesize \cite{mostajabdaveh2024evaluatingllmreasoningoperations} & \footnotesize \new{1}\\

    \bottomrule
    \end{tabular}
\end{table}

\clearpage

\section{Classification of Studies by Application Domain}
\label{app:application-analysis}

\new{This appendix provides the tabular analysis related to the classification of studies by application domain (\Cref{tab:application-analysis}). Please refer to \Cref{sec:application} for further context.}

\begin{table}[htbp]
    \centering
    \caption{Classification of studies by application domain.}
    \label{tab:application-analysis}
    \begin{tabular}{
    p{3cm}
    p{5.5cm}
    p{4.5cm}
    C{0.5cm}}
    \toprule
    \footnotesize \textbf{Domain} & \footnotesize \textbf{\gls{cop}/Detail} & \footnotesize \textbf{Studies} & \footnotesize \textbf{\#}\\
    \midrule
    \multirow{4}{*}{\footnotesize Routing} & \footnotesize \raisebox{-0.6\height}{Traveling Salesperson} & \footnotesize \cite{yang2024large,liu2024large,LIU2023100520,liu2023algorithm,liu2024evolution,ye2024large,make6030093,Khan_2024,huang2024exploringtruepotentialevaluating,yao2024multiobjectiveevolutionheuristicusing,vanstein2024intheloophyperparameteroptimizationllmbased,sui-2024,chen2024uberuncertaintybasedevolutionlarge,liu2024llm4adplatformalgorithmdesign,DBLP:conf/esann/MartinekLG24,10.1007/978-3-031-70068-2_12} & \multirow{4}{*}{\footnotesize \new{26}}\\
    
    & \footnotesize Vehicle Routing & \footnotesize \cite{10.1007/978-981-97-2259-4_3,huang2024words,huang2024multimodal,chin2024learning,ye2024large,a17120582, Da2024,Khan_2024,liu2024llm4adplatformalgorithmdesign,wu2024neuralcombinatorialoptimizationalgorithms} \\
    
    & \footnotesize Orienteering & \footnotesize \cite{ye2024large} \\

    & \footnotesize \new{Travel} & \footnotesize \cite{ju-etal-2024-globe,delarosa2024trippaltravelplanningguarantees,10704489} \\
        
    \midrule

    \multirow{4}{*}{\footnotesize Scheduling and Planning} & \footnotesize Permutation Flowshop Scheduling & \footnotesize \cite{amarasinghe2023aicopilot,liu2024evolution} & \multirow{4}{*}{\footnotesize \new{14}} \\
    
    & \footnotesize Meeting and Conferece Scheduling & \footnotesize \cite{lawless2024i,10.1145/3664646.3665084} \\

    & \footnotesize \new{Server Scheduling} & \footnotesize\cite{yatong2024tseohedgeservertask} \\

     & \footnotesize Planning & \footnotesize \cite{buildings13071772,10.1609/icaps.v34i1.31503,10803039,10675146,10711695,jiang2024largelanguagemodelscombinatorial,bohnet2024exploringbenchmarkingplanningcapabilities,hao2025planningrigorgeneralpurposezeroshot,Wang_2024} \\

     \midrule

    \multirow{6}{*}{\footnotesize Network and Graphs} & \footnotesize \new{Network Design} & \footnotesize \cite{yu2024autornetautomaticallyoptimizingheuristics,10829820,hu2024scalableaccurategraphreasoning,li2025graphteamfacilitatinglargelanguage,smartcities7050094} & \multirow{6}{*}{\footnotesize \new{10}}\\ 

    & \footnotesize Critical Node Identification & \footnotesize \cite{mao2024identify} & \\
    
     & \footnotesize \new{Coloring} & \footnotesize\cite{DBLP:conf/esann/MartinekLG24} \\

    & \footnotesize \new{Social Network} & \footnotesize \cite{10818476} \\

    & \footnotesize \new{Shortest Path, Assignement} & \footnotesize \cite{Khan_2024} \\

    & \footnotesize \new{Path Finding} & \footnotesize \cite{borazjanizadeh2024navigatinglabyrinthevaluatingenhancing} \\
    
    \midrule

    \multirow{2}{*}{\footnotesize Packing} & \footnotesize Bin Packing & \footnotesize \cite{ye2024large,Romera-Paredes2024,liu2024evolution,yao2024multiobjectiveevolutionheuristicusing,vanstein2024intheloophyperparameteroptimizationllmbased,chen2024uberuncertaintybasedevolutionlarge,liu2024llm4adplatformalgorithmdesign,10.1007/978-3-031-70068-2_12} & \multirow{2}{*}{\footnotesize \new{8}} \\
    
    & \footnotesize Multiple Knapsack & \footnotesize \cite{ye2024large} \\ 
    
    \midrule
    
    \multirow{4}{*}{\footnotesize Combinatorics} & \footnotesize  Cap Set & \footnotesize \cite{Romera-Paredes2024} & \multirow{4}{*}{\footnotesize \new{4}} \\

    & \footnotesize \new{Admissible Set} & \footnotesize \cite{10.1007/978-3-031-70068-2_12}\\

    & \footnotesize \new{Puzzles} & \footnotesize \cite{borazjanizadeh2024navigatinglabyrinthevaluatingenhancing,michailidis_et_al:LIPIcs.CP.2024.20} \\

    & \footnotesize \new{Subset Sum, Sorting, Under-determined Systems} & \footnotesize \cite{borazjanizadeh2024navigatinglabyrinthevaluatingenhancing} \\
 
    \midrule
    
    \multirow{3}{*}{\footnotesize Engineering} & \footnotesize Construction & \footnotesize \cite{SAKA2024100300} &  \multirow{3}{*}{\footnotesize \new{3}} \\

    & \footnotesize Circuit Design & \footnotesize \cite{ye2024large} \\    

    & \footnotesize \new{Energy Management} & \footnotesize \cite{10738100}\\

    \midrule
 
    \footnotesize Finance & \footnotesize Portfolio Optimization & \footnotesize \cite{RePEc:spr:snopef:v:4:y:2023:i:4:d:10.1007_s43069-023-00277-6,ai5010006,ALIPOURVAEZI2024110574} & \footnotesize \new{3} \\ 
 
    \midrule
    
    \footnotesize Bioinformatics & \footnotesize Enzyme Design & \footnotesize \cite{teukam2024integrating,reinhart-2024,10628050} & \footnotesize \new{3} \\

    \midrule

    \footnotesize Supply Chain & \footnotesize Warehousing & \footnotesize \cite{li2023large} & \footnotesize 1\\

    \midrule
    
    \footnotesize \new{Strings} & \footnotesize \new{Text Generation} & \footnotesize \cite{regin_et_al:LIPIcs.CP.2024.25} & \footnotesize \new{1} \\
    
    \bottomrule     
    \end{tabular}
\end{table}

\end{document}